\documentclass[10pt,twocolumn,letterpaper]{article}

\usepackage{iccv}
\usepackage{times}
\usepackage{epsfig}
\usepackage{graphicx}
\usepackage{amsmath}
\usepackage{amssymb}

\usepackage{algorithm}
\usepackage{algorithmic}
\usepackage{bm}
\usepackage{booktabs}
\usepackage{multirow}
\usepackage[pagebackref=true,breaklinks=true,letterpaper=true,colorlinks,bookmarks=false]{hyperref}

\iccvfinalcopy % *** Uncomment this line for the final submission

\def\iccvPaperID{****} % *** Enter the ICCV Paper ID here

\ificcvfinal\fi

\begin{document}

%%%%%%%%% TITLE
\title{Towards High-Resolution Salient Object Detection}
\author{Yi Zeng$^1$, Pingping Zhang$^1$, Jianming Zhang$^2$, Zhe Lin$^2$, Huchuan Lu$^1$\thanks{Corresponding author.}\\
$^1$ Dalian University of Technology, China\\
$^2$ Adobe Research, USA\\
{\tt\small \{dllgzy, jssxzhpp\}@mail.dlut.edu.cn, \{jianmzha, zlin\}@adobe.com, lhchuan@dlut.edu.cn}\\
%\and
%{\tt\small @adobe.com, }\\
}

\maketitle
% Remove page # from the first page of camera-ready.
\ificcvfinal\thispagestyle{empty}\fi

%%%%%%%%% ABSTRACT
\begin{abstract}
Deep neural network based methods have made a significant breakthrough in salient object detection.
However, they are typically limited to input images with low resolutions ($400\times400$ pixels or less).
Little effort has been made to train deep neural networks to directly handle salient object detection in very high-resolution images.
This paper pushes forward high-resolution saliency detection, and contributes a new dataset, named High-Resolution Salient Object Detection (HRSOD).
To our best knowledge, HRSOD is the first high-resolution saliency detection dataset to date.
As another contribution, we also propose a novel approach, which incorporates both global semantic information and local high-resolution details, to address this challenging task.
More specifically, our approach consists of a Global Semantic Network (GSN), a Local Refinement Network (LRN) and a Global-Local Fusion Network (GLFN).
GSN extracts the global semantic information based on down-sampled entire image.
Guided by the results of GSN, LRN focuses on some local regions and progressively produces high-resolution predictions.
GLFN is further proposed to enforce spatial consistency and boost performance.
Experiments illustrate that our method outperforms existing state-of-the-art methods on high-resolution saliency datasets by a large margin, and achieves comparable or even better performance than them on widely-used saliency benchmarks.
The HRSOD dataset is available at \textcolor{red}{https://github.com/yi94code/HRSOD}.
\end{abstract}

%%%%%%%%% BODY TEXT
\section{Introduction}
Salient object detection, aiming at accurately detecting and segmenting the most distinctive object regions in a scene, has drawn increasing attention in recent years~\cite{fan2018salient,zhang2019salient,zhang2018hyperfusion,zhang2018agile,zhang2018non}. It is regarded as a very important task that can facilitate a wide range of applications, such as image understanding~\cite{lai2016saliency,zhu2015unsupervised,zhang2015saliency}, object segmentation~\cite{joon2017exploiting}, image captioning~\cite{fang2015captions,das2017human,xu2015show} and light field 3D display~\cite{wang2018salience}.

\begin{figure}
\centering
\includegraphics[width=0.45\textwidth,height=0.3\textheight]{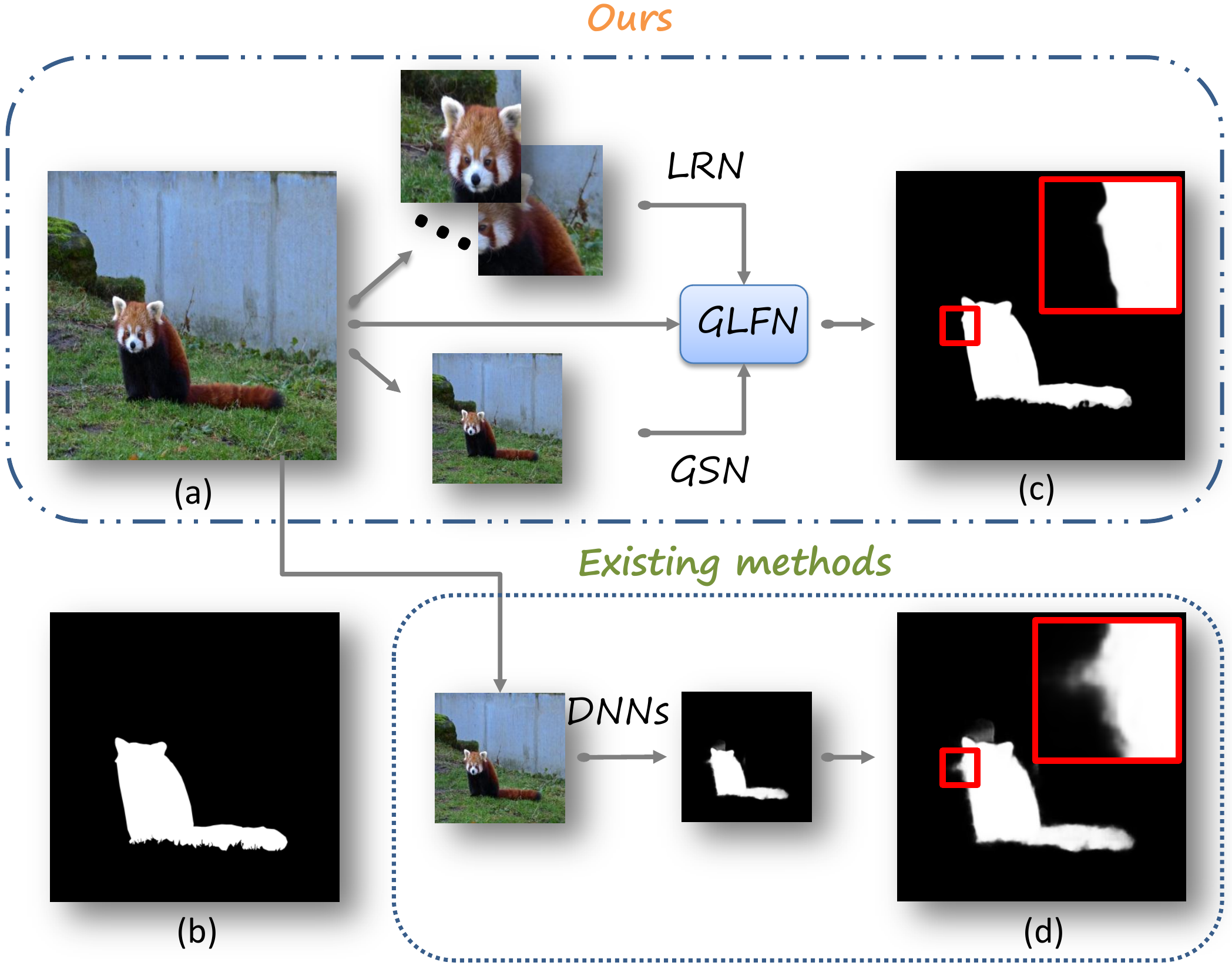}\\
\caption{Pipeline comparison with state-of-the-art methods. (a) Input image. (b) Ground truth mask.  (c) Our method. (d) Amulet \cite{zhang2017amulet}. Best viewed by zooming in.}
\label{fig:intro}
\vspace{-5mm}
\end{figure}

Deep Neural Networks (DNNs), \emph{e.g.}, VGG~\cite{simonyan2014very}, ResNet~\cite{he2016deep}, have achieved remarkable success in computer vision tasks using the typical input size such as $224\times224$, $384\times384$, etc. For most applications, such as image classification, object detection and visual tracking, the typical input size is enough to obtain satisfied results. For dense prediction tasks, \emph{e.g.}, image segmentation and saliency detection, deep learning based approaches also show impressive performance. But the inherited defect is very apparent, \emph{i.e.}, blurry boundary. Many research efforts have been made to remedy this problem. For example, Zhang \emph{et al.}~\cite{zhang2017amulet} employ deep recursive supervision and integrate multi-level features for accurate boundary prediction.
However, the improvement is not significant, as illustrated in Figure \ref{fig:intro} (d).

Furthermore, the resolution of the images taken by electronic products (\emph{e.g.}, smartphones) becomes very high, \emph{e.g.}, 720p, 1080p and 4K. When processing high-resolution images, the above defect becomes more severe. The state-of-the-art saliency detection methods generally down-scale the inputs to extract semantic information. In this process, many details are inevitably lost. Thus, they are not suitable for high-resolution saliency detection task.
Meanwhile, there is little research effort to train neural networks to directly handle salient object segmentation in very high-resolution images.

However, this line of work is very important since it can inspire or enable many practical tasks such as image editing~\cite{tsai2016sky,xiao2019auto,lischinski2006interactive}, medical image analysis~\cite{chen2019learning}, etc. Specifically, when served as a pre-processing step of background replacement and
depth-of-field, high-resolution salient object detection should be as accurate as possible to provide users with realistic composite images~\cite{shen2016automatic}. If the predicted boundaries are not accurate, there may be artifacts which certainly affect users' experience. Thus, this paper pushes forward the task of high-resolution salient object detection.

To our best knowledge, our approach is the first work for high-resolution salient object detection. Since there is no high-resolution training and test dataset for saliency detection, we contribute a new dataset, High-Resolution Salient Object Detection (HRSOD). More details about our HRSOD will be presented in Section \ref{dataset}.

As for developing high-resolution saliency detection methods, there are three intuitive methods. The first is simply increasing the input size to maintain a relative high resolution and object details after a series of pooling operations.
However, the large input size results in significant increases in memory usage. Moreover, it remains a question that if we can effectively extract details from lower-level layers in such a deep network through back propagation. The second method is partitioning inputs into patches and making predictions patch-by-patch. However, this type of method is time-consuming and can easily be affected by background noise. The third one includes some post-processing methods such as CRF~\cite{krahenbuhl2011efficient} or graph cuts~\cite{rother2004grabcut}, which can address this issue to a certain degree. But very few works attempted to solve it directly within the neural network training process. As a result, the problem of applying DNNs for high-resolution salient object detection is fairly unsolved.

To address above issues, we propose a novel deep learning approach for high-resolution salient object detection without any post-processing.
It has a Global Semantic Network (GSN) for extracting global semantic information, and a Local Refinement Network (LRN) for optimizing local object details.
A global semantic guidance is introduced from GSN to LRN in order to ensure global consistency. Besides, an Attended Patch Sampling (APS) scheme is proposed to enforce LRN to focus on uncertain regions, and this scheme provides a good trade-off between performance and efficiency. Finally, a Global-Local Fusion Network (GLFN) is proposed to enforce spatial consistency and further boost performance at high resolution.

To summarize, our contributions are as follows:
\begin{itemize}
\item
We introduce the first high-resolution salient object detection dataset (HRSOD) with rich boundary details and accurate pixel-wise annotations.
\item
We provide a new paradigm for high-resolution salient object detection which first uses GSN for extracting semantic information, and a guided LRN for optimizing local details, and finally GLFN for prediction fusion.
\item
We perform extensive experiments to demonstrate that our method outperforms other state-of-the-art methods on high-resolution saliency datasets by a large margin, and achieves comparable performance on some widely used saliency benchmarks.
\end{itemize}
%-------------------------------------------------------------------------
\section{Related Work}
In the past few decades, lots of approaches have been proposed to solve the saliency detection problem. Early researches are mainly based on low-level features, such as image contrast~\cite{itti1998model,cheng2015global}, texture~\cite{yan2013hierarchical,yang2013saliency} and background prior~\cite{li2013saliency,wei2012geodesic}. These models are efficient and effective in simple scenarios, but they are not always robust in handling challenging cases. A detailed survey of these methods can be found in~\cite{borji2015salient}.

More recently, learning based saliency detection methods have achieved expressive performance, and they can coarsely be divided into two categories, \emph{i.e.}, patch-based saliency and FCN-based saliency.
\subsection{Patch-based Saliency}
Existing patch-based methods make saliency prediction for each image patch. For example, Wang \emph{et al.}~\cite{wang2015deep} present a saliency detection algorithm by integrating both local estimation and global search. Then, Li \emph{et al.}~\cite{li2015visual} propose to utilize multi-scale features in multiple generic CNNs to predict the saliency degree of each superpixel. With the same purpose of predicting the saliency degree of each superpixel, Zhao \emph{et al.}~\cite{zhao2015saliency} use a multi-context deep CNN to predict saliency maps taking global and local context into account. The above methods include several fully connected layers to make predictions in superpixel-level, resulting in expensive computational cost and the loss of spatial information. What's more, all of them make very coarse predictions and lack low-level details.
\subsection{FCN-based Saliency}
Liu \emph{et al.}~\cite{liu2016dhsnet} design a deep hierarchical saliency network and progressively recover image details via integrating local context information. Zhang \emph{et al.}~\cite{zhang2017amulet} propose a generic framework to integrate multi-level features into different resolutions for finer saliency maps. In order to better integrate features from different levels, Zhang \emph{et al.}~\cite{zhang2018bi} propose a bi-directional message passing module with a gate function to integrate multi-level features. Wang \emph{et al.}~\cite{wang2018detect} use a boundary refinement network to learn propagation coefficients for each spatial position.

%\vspace{4mm}
Lots of research efforts have been made to recover image details in final predictions. However, for high-resolution images, all existing FCN-based methods down-sample the inputs, thus lose high-resolution details and fail to predict fine-grained saliency maps. %Prior works

Several researchers attempt to remedy this problem by using post-processing techniques for finer predictions. However, traditional CRF~\cite{krahenbuhl2011efficient} and guided filtering are very time-consuming and their improvement is very limited. Wu \emph{et al.} \cite{wu2018fast} propose a more efficient guided filtering layer. However, their performance is just comparable with the CRF. To reduce this gap, we propose a method to combine the advantages of patch-based methods (maintaining details and saving memory) and FCN-based methods (having rich contextual information).
\section{High-Resolution Saliency Detection Dataset}\label{dataset}

\begin{figure}[!t]
    \centering
    \hspace{-2mm}
        \begin{tabular}{c@{\hspace{0.5mm}} c@{\hspace{0.5mm}}}
            \includegraphics[width=0.23\textwidth,height=0.15\textheight]{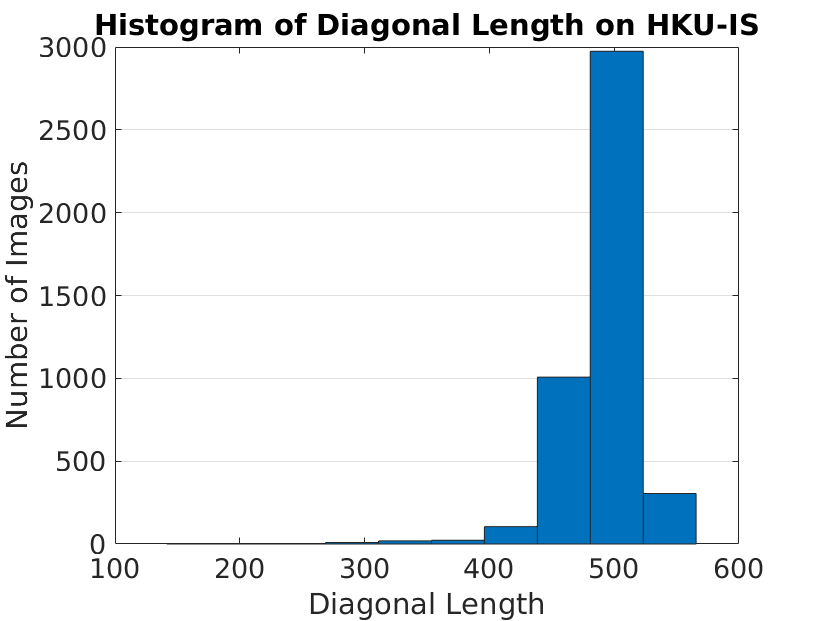}&
            \includegraphics[width=0.23\textwidth,height=0.15\textheight]{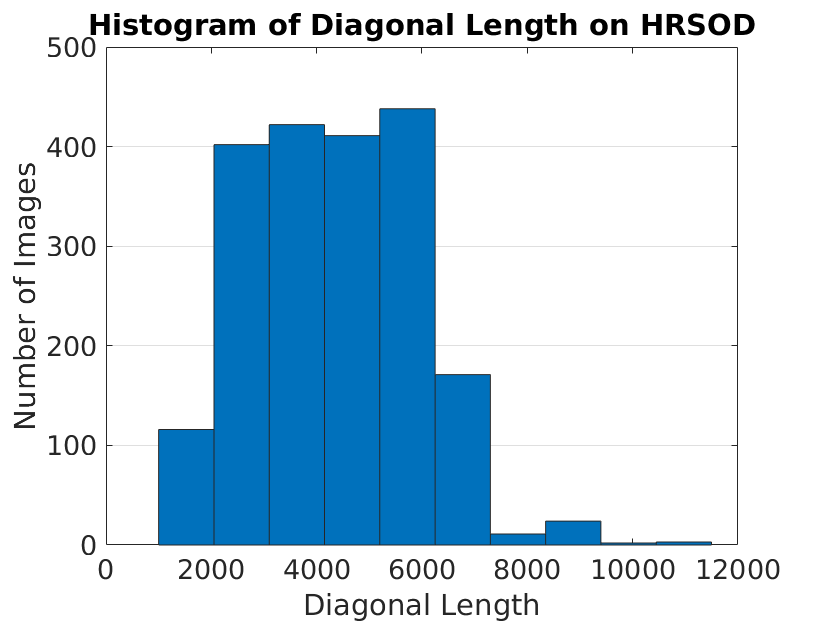}\\
            (a)&(b)\\
            \includegraphics[width=0.20\textwidth,height=0.12\textheight]{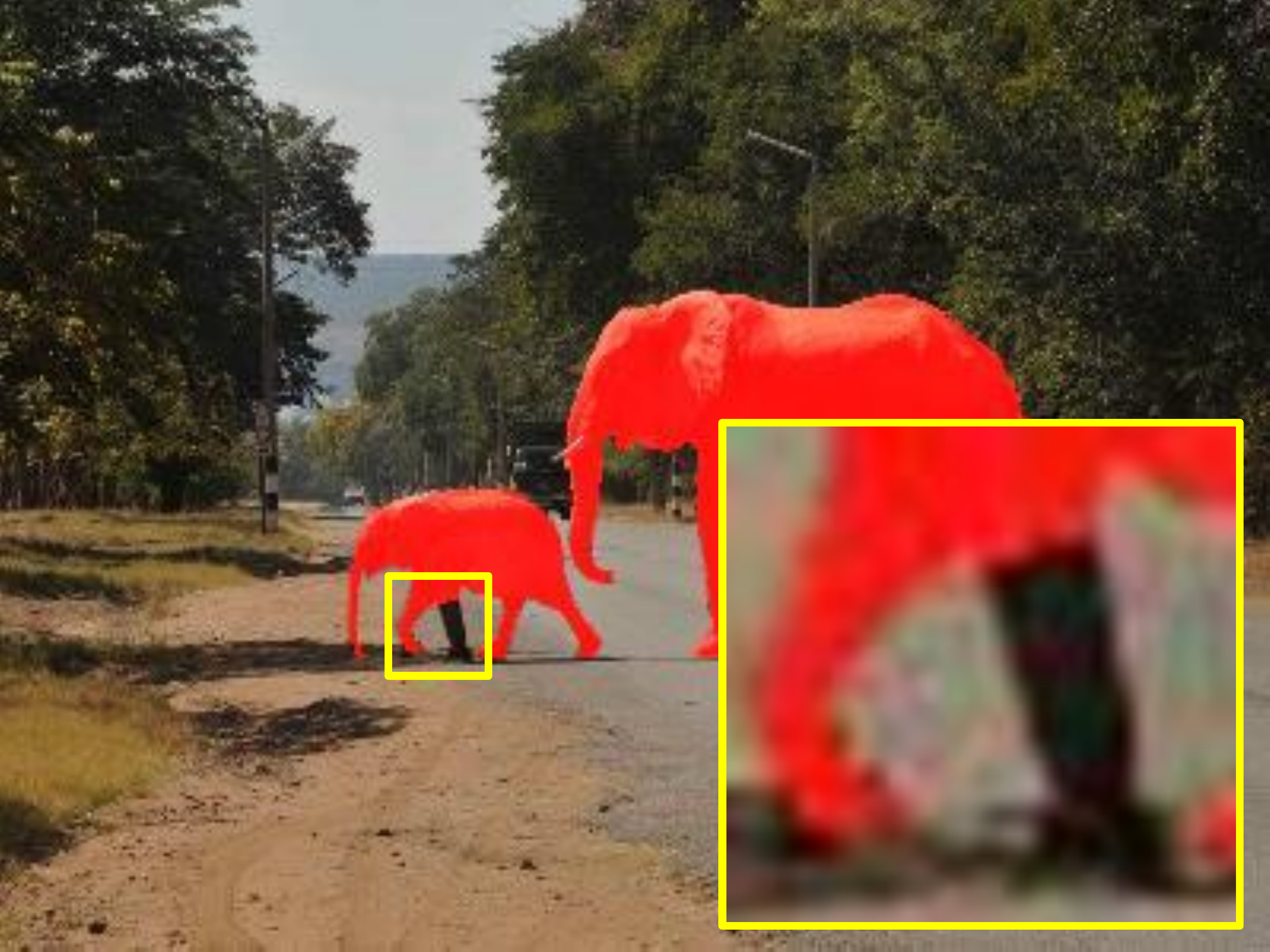}&
            \includegraphics[width=0.20\textwidth,height=0.12\textheight]{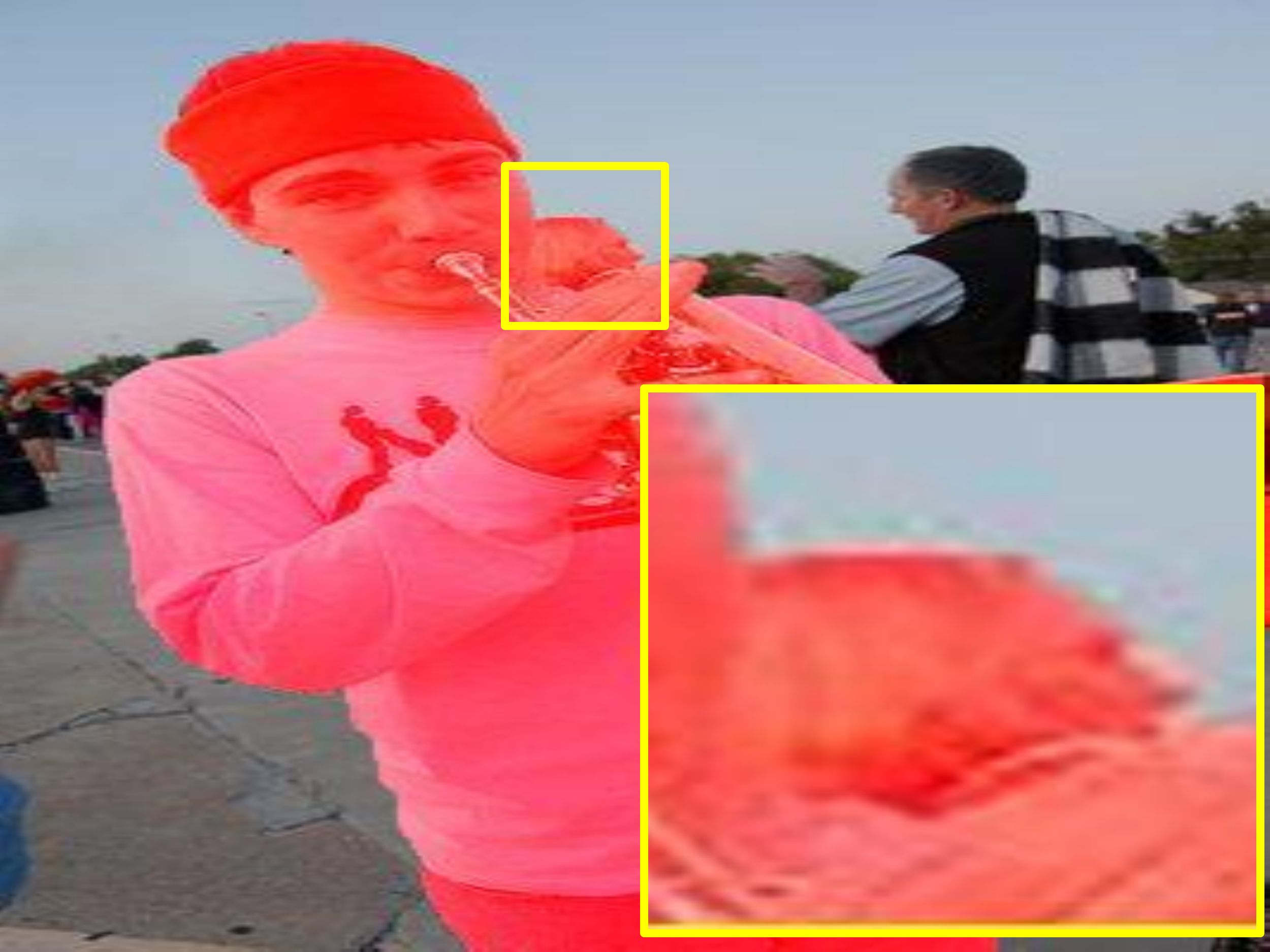}\\
            (c)&(d)\\
            \includegraphics[width=0.20\textwidth,height=0.12\textheight]{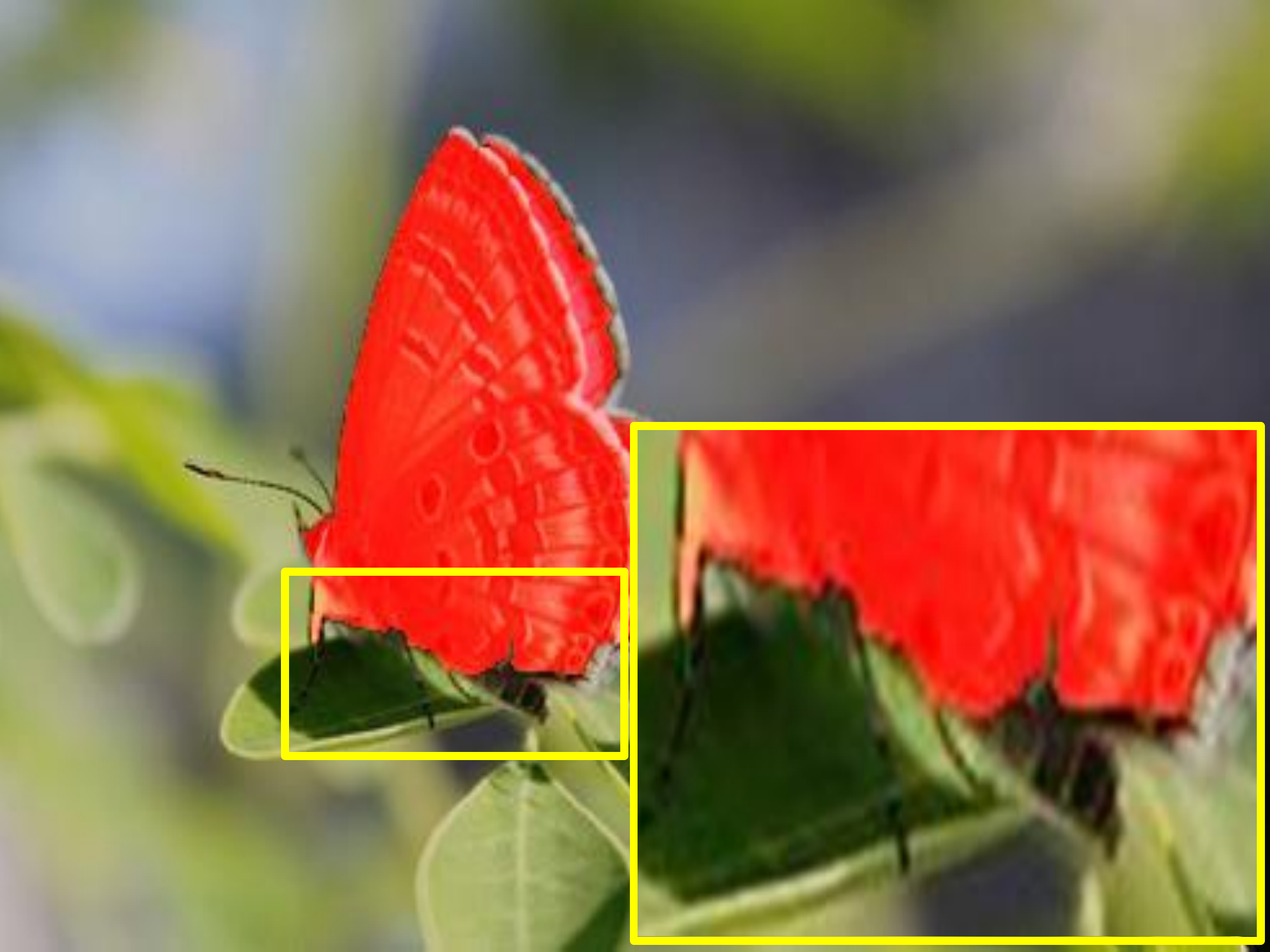}&
            \includegraphics[width=0.20\textwidth,height=0.12\textheight]{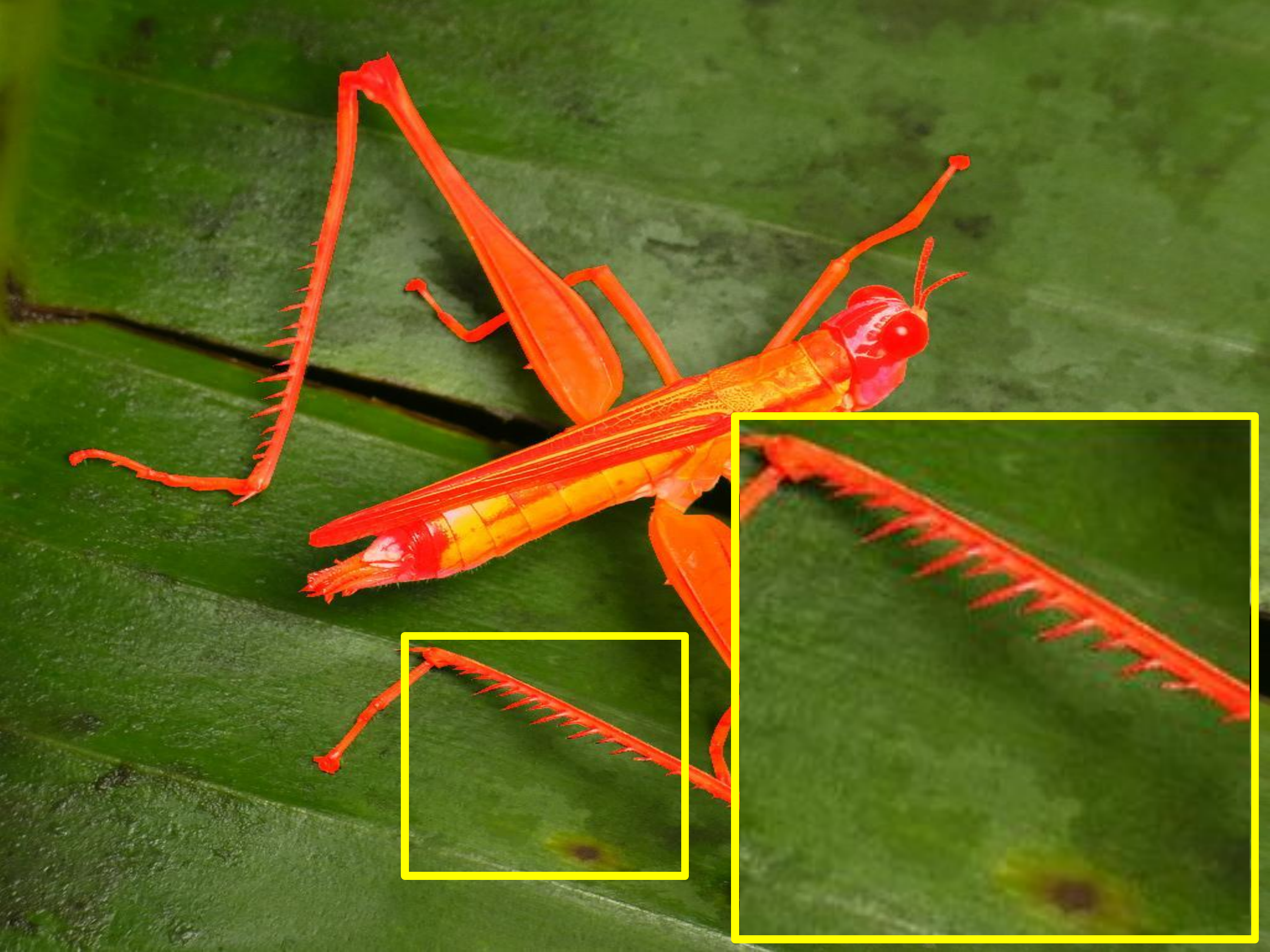}\\
            (e)&(f)\\
        \end{tabular}
    \caption{(a) The histogram of diagonal length on HKU-IS \cite{li2015visual} (The maximum is less than 600.). (b) The histogram of diagonal length on our HRSOD (The minimum is over 1000.). (c)-(f) Sample images from various dataset, with ground truth masks overlayed. Concretely, (c) is from HKU-IS \cite{li2015visual}. (d) is from DUTS-Test \cite{wang2017}. (e) is from THUR \cite{cheng2014salientshape}. And (f) is an example of our HRSOD. Best viewed by zooming in.}
    \label{fig:datset}
    \vspace{-5mm}
\end{figure}

There exist several datasets for saliency detection, but none of them is specifically designed for high-resolution salient object detection. Three main drawbacks are apparent. First, all images in current datasets have extremely limited resolutions. Concretely, the longest edge of each image is less than 500 pixels. These low-resolution images are not representative
for today's image processing applications.
Second, to relieve the burden of users, it is essential to output masks with extremely high accuracy in boundary regions. But images in existing saliency detection datasets are inadequate in \emph{providing rich object boundary details} for training DNNs. In addition, widely used saliency datasets also have some problems in annotation quality, such as failing to cover all saliency regions (Figure \ref{fig:datset} (c)), including background disturbance into foreground annotation (Figure \ref{fig:datset} (d)), or low contour accuracy (Figure \ref{fig:datset} (e)).

To address the above urgent issues, we contribute a $\bm{H}$igh-$\bm{R}$esolution $\bm{S}$alient $\bm{O}$bject $\bm{D}$etection (HRSOD) dataset, containing 1610 training images and 400 test images. The total 2010 images are collected from the website of Flickr\footnote{https://www.flickr.com} with the license of all creative commons. Pixel-level ground truths are manually annotated by 40 subjects. The shortest edge of each image in our HRSOD is more than 1200 pixels. Figure \ref{fig:datset} presents the image size comparison between our HRSOD and existing saliency detection datasets. For existing datasets, we only show the results on HKU-IS dataset~\cite{li2015visual}, and the results hold the same on other datasets. Besides, we provide an analysis of shape complexity in supplementary material. Compared with existing saliency datasets, our HRSOD avoids low-level mistakes via careful check by over 5 subjects (an example shown in Figure \ref{fig:datset} (f)). To our best knowledge, HRSOD is currently the first high-resolution dataset for salient
object detection. It is specifically designed for training and evaluating DNNs aiming at high-resolution salient object detection. The whole dataset is publicly available\footnote{https://github.com/yi94code/HRSOD}.
%-------------------------------------------------------------------------
%-------------------------------------------------------------------------
\section{Our Method}
\begin{figure}[!t]
  \centering
  % Requires \usepackage{graphicx}
  \includegraphics[width=1 \linewidth]{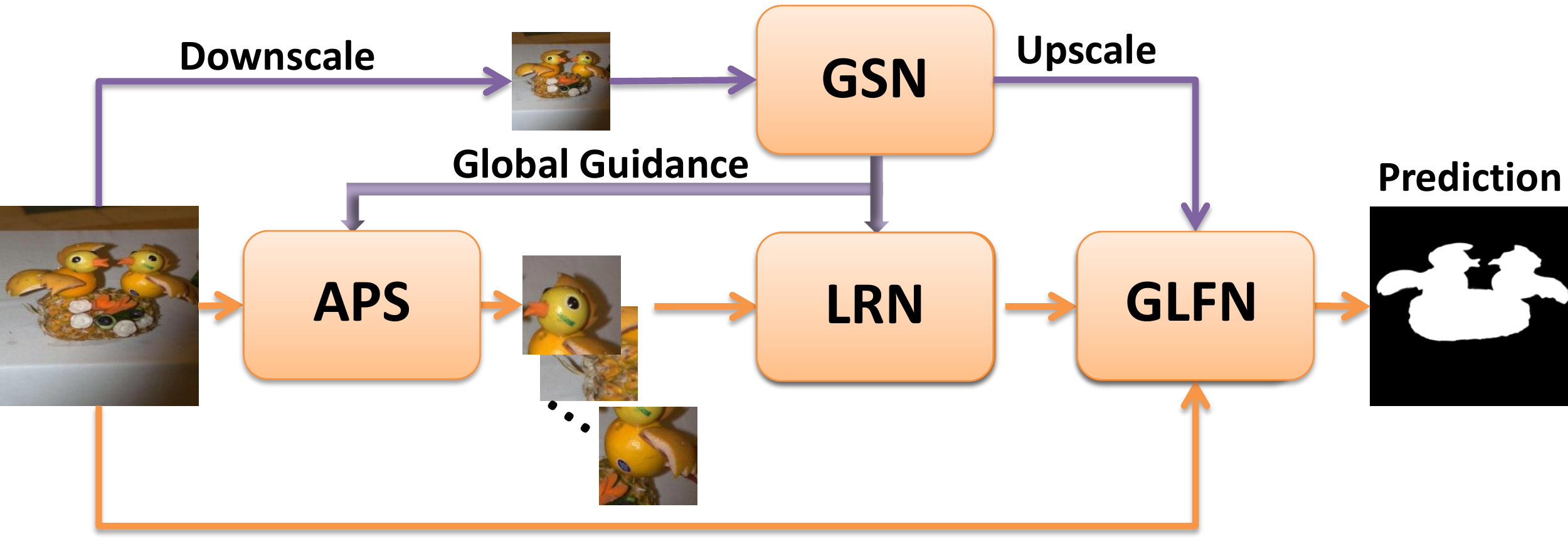}\\
  \caption{Overview of the network architecture. GSN and LRN takes downscaled entire images and attended sub-images as input, respectively. The guidance from GSN provides some semantic knowledge and ensures that our APS and LRN are attended to uncertain regions. A GLFN is appended to directly leverage high-resolution information to fuse the predictions from GSN and LRN.}\label{overview}%In the first stage, we train the GSN for overall coarse predictions. In the second stage, the LRN is trained for finer saliency predictions under the guidance from GSN.
  \vspace{-4mm}
\end{figure}

In this paper, we propose a novel method for detecting salient objects in high-resolution images with limited GPU memory. Our framework includes three branches, \emph{i.e.,} Global Semantic Network (GSN), Local Refinement Network (LRN) and Global-Local Fusion Network (GLFN). Figure~\ref{overview} shows an overall illustration of the proposed approach. GSN aims at extracting semantic knowledge in a global view. Guided by GSN, LRN is designed to refine uncertain sub-regions. Finally, GLFN takes high-resolution images as inputs and further enforces spatial consistency of the fused predictions from GSN and LRN.

To be specific, let $\{\bm{X}_i = (\bm{I}_i, \bm{L}_i)\}_{i=1}^N$ be the training set, containing both the training image $\bm{I}_i$ and its pixel-wise saliency label $\bm{L}_i$. The input image $\bm{I}_i$ is first fed forward through GSN to obtain a coarse saliency map $\bm{F}_i$, denoted as:
\begin{equation}
\bm{F}_i = UP(GSN(DS(\bm{I}_i), \bm{\theta}))
\end{equation}
where $DS(\cdot)$ denotes down-sampling images to $384\times384$ while $UP(\cdot)$ denotes up-sampling predictions to original size. $\bm{\theta}$ denotes all parameters in GSN. Then %$\bm{I_i}$, $\bm{L_i}$\}
$\bm{I}_i$ is put into our proposed Attended Patch Sampling (APS) scheme (Algorithm \ref{alg:crop}) to generate sub-images $\{\bm{P}_m^{\bm{I}_i}\}_{m=1}^M$, which are attended to uncertain regions ($M$ is the total number of sub-images for each input $\bm{I}_i$). Subsequently, each $\bm{P}_m^{\bm{I}_i}$ is fed forward through LRN to get a refined saliency map $\bm{R}_m^{\bm{I}_i}$. Semantic guidance is introduced from GSN to LRN (Section \ref{42} ). Finally, the outputs of GSN and LRN are fused and fed forward through GLFN for final prediction $\bm{S}_i$. These two stages can be formulated as:
\begin{equation}
\{\bm{R}_m^{\bm{I}_i}\}_{m=1}^M = LRN(\{\bm{P}_m^{\bm{I}_i}\}_{m=1}^M, \bm{\phi})
\end{equation}
\begin{equation}
\bm{S}_i = GLFN(\bm{I}_i, Fuse(\{\bm{R}_m^{\bm{I}_i}\}_{m=1}^M, \bm{F}_i), \bm{\psi})
\end{equation}
where $\bm{\phi}$ and $\bm{\psi}$ denote the parameters of LRN and GLFN, respectively. $Fuse(\cdot)$ denotes fusion operation (more details can be seen in Section \ref{44}).

\subsection{Network Architecture for GSN and LRN}
  \vspace{-2mm}
\begin{figure}[h]
  \centering
  % Requires \usepackage{graphicx}
  \includegraphics[width=1 \linewidth]{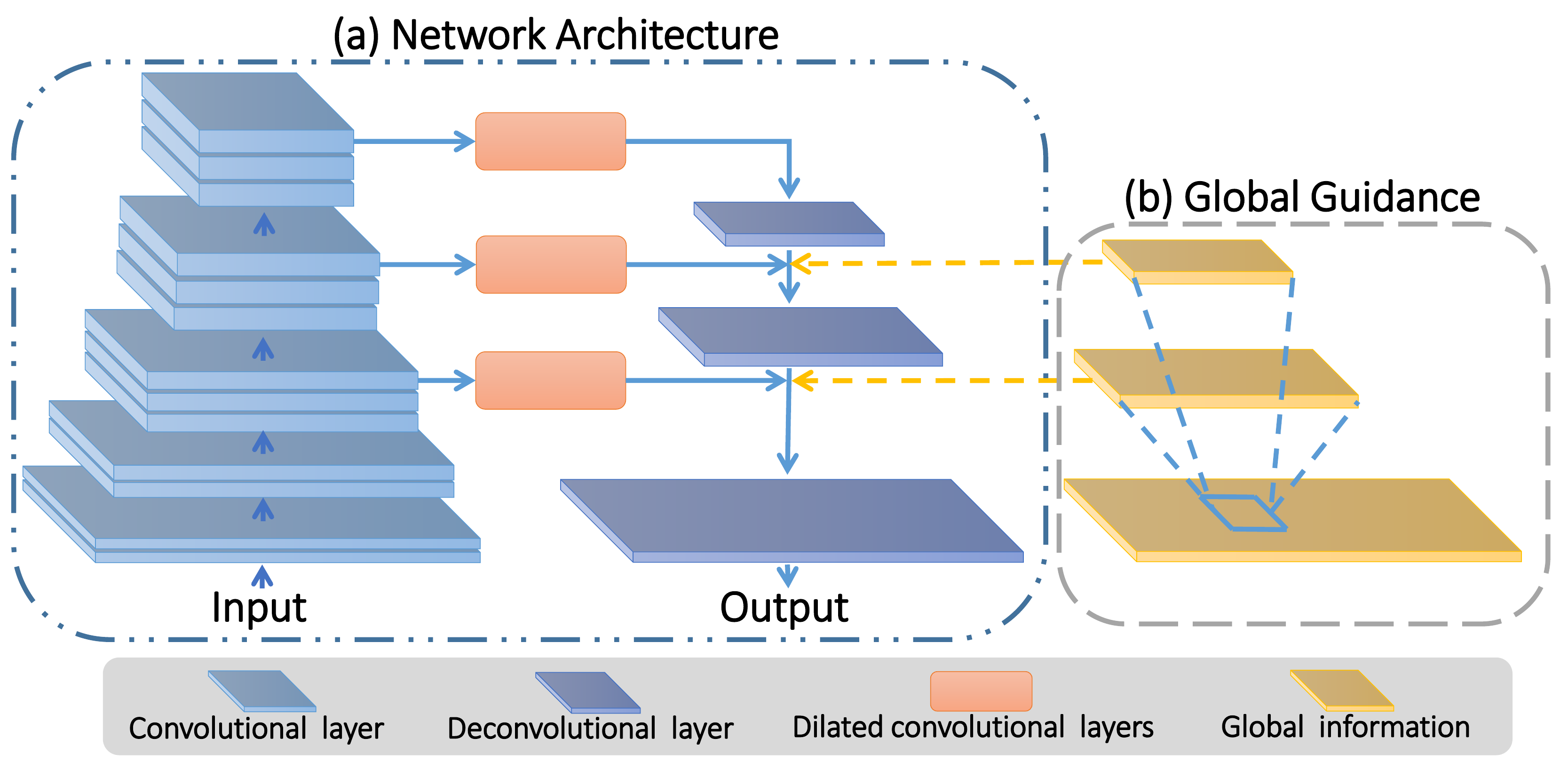}\\
  \caption{(a) Network architecture for both GSN and LRN. (b) Incorporate global guidance only for LRN.}
  \label{gsnandlrn}
  \vspace{-2mm}
\end{figure}
We adopt the same backbone for GSN and LRN.
Our model is simply built on the FCN architecture with the pre-trained 16-layer VGG network~\cite{simonyan2014very}. The original VGG-16 network~\cite{simonyan2014very} is trained for image classification task while our model is trained for saliency detection, a pixel-wise prediction task. Therefore, we simply abandon all layers after conv5\_3 to maintain a higher resolution.

In order to enlarge receptive field, we employ dilated convolutional layers~\cite{yu2015multi} to capture contextual information. Dilated convolution, also known as atrous convolution, has a superior ability to enlarge the field of view without increasing the number of parameters. As shown in Figure~\ref{gsnandlrn} (a), we add four dilated convolutional layers on the top of conv3-3, conv4-3 and conv5-3 in our revised VGG-16. All the dilated convolutional layers have the same kernel size and output channels, \emph{i.e.}, $k=3$ and $c=32$. The rates of the four dilated convolutional layers in the same block are set with $dilation=1, 3, 5, 7$ respectively.

To improve the output resolution, we first generate three saliency score maps through the last three blocks. Secondly, we add three additional deconvolutional layers, the first two of which have 2{$\times$} upsampling factors and the last of which has a 4{$\times$} upsampling factor. Thirdly, inspired by~\cite{long2015fully}, we build two skip connections from the saliency score maps generated by block 3 and block 4 to combine high-level features with meaningful semantic information and low-level features with large amount of details (See Figure~\ref{gsnandlrn} (a)). More details are provided in the supplementary material.

\subsection{Semantic Guidance from GSN to LRN} \label{42}

The saliency maps generated by GSN are based on the full image and embedded with rich contextual information. Nevertheless, due to its small input size of 384{$\times$}384, lots of low-level details are lost, especially when the original images have very high resolutions (\emph{e.g.}, 1920{$\times$}1080). That is to say, it barely learns to capture saliency properties at a coarse scale. As a result, GSN is competent in giving a rough saliency prediction but insufficient to precisely localize salient objects. In contrary, LRN takes sub-images as input, avoiding down-sampling which results in the loss of details. However, since sub-images are too local to indicate which area is more salient, LRN may be confused about which region should be highlighted. Also, LRN alone may have false alarms in some locally salient regions. Therefore, we propose to introduce the semantic guidance from GSN to LRN, in order to enhance global contextual knowledge while maintain high-resolution details.

Specifically, we incorporate global semantic guidance
in the decoder part. As illustrated in Figure \ref{gsnandlrn} (b), given the coarse result $\bm{F}_i$ of GSN, a patch $\bm{P}_m^{\bm{F}_i}$ is first cropped according to the location of
patch $\bm{P}_m^{\bm{I}_i}$ in LRN. Then we concatenate $\bm{P}_m^{\bm{F}_i}$ with the corresponding feature maps in LRN.
\begin{figure*}[!t]
  \vspace{-3mm}
  \centering
  % Requires \usepackage{graphicx}
  \includegraphics[width=1\textwidth,height=0.25\textheight]{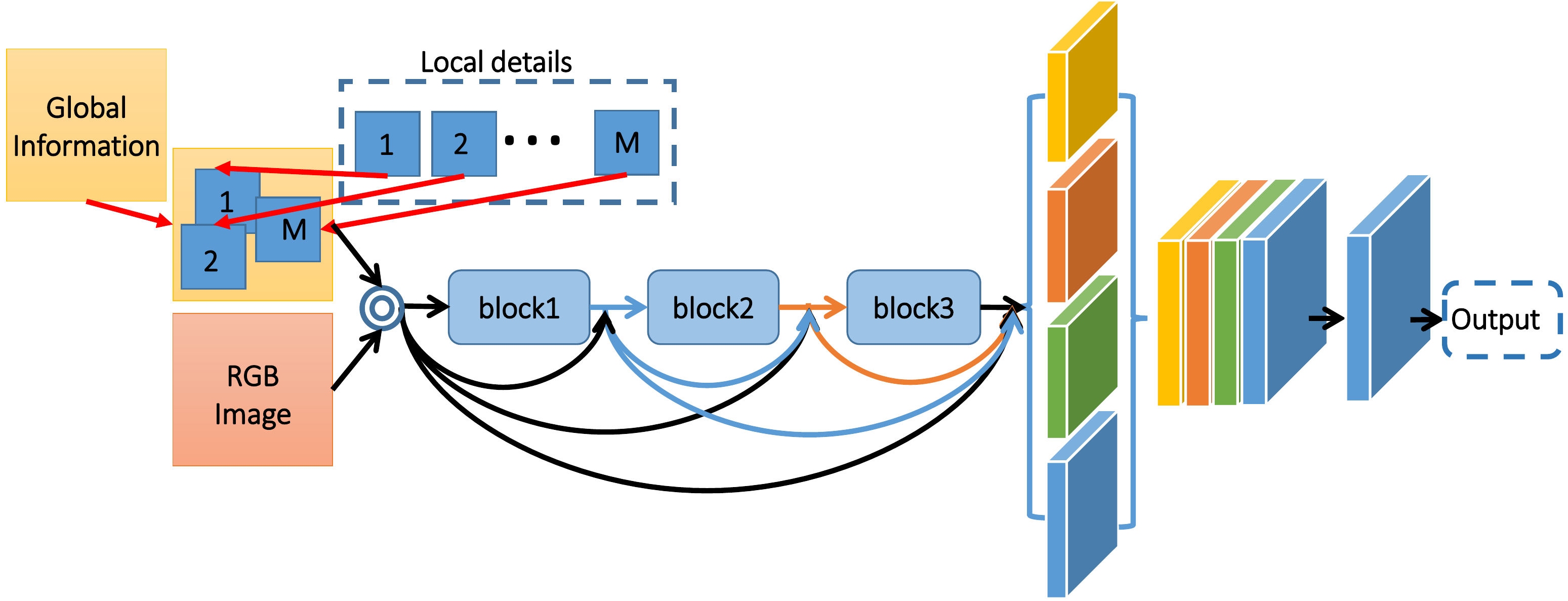}\\
  \caption{Global-Local Fusion Network.}\label{glfn}
  \vspace{-4mm}
\end{figure*}
\subsection{Focus on Uncertain Regions}
\begin{figure}[h]
    \centering
    \tabcolsep0.3mm \renewcommand{\arraystretch}{0.5}
        \begin{tabular}{ccc}
            \includegraphics[width=0.14\textwidth,height=0.09\textheight]{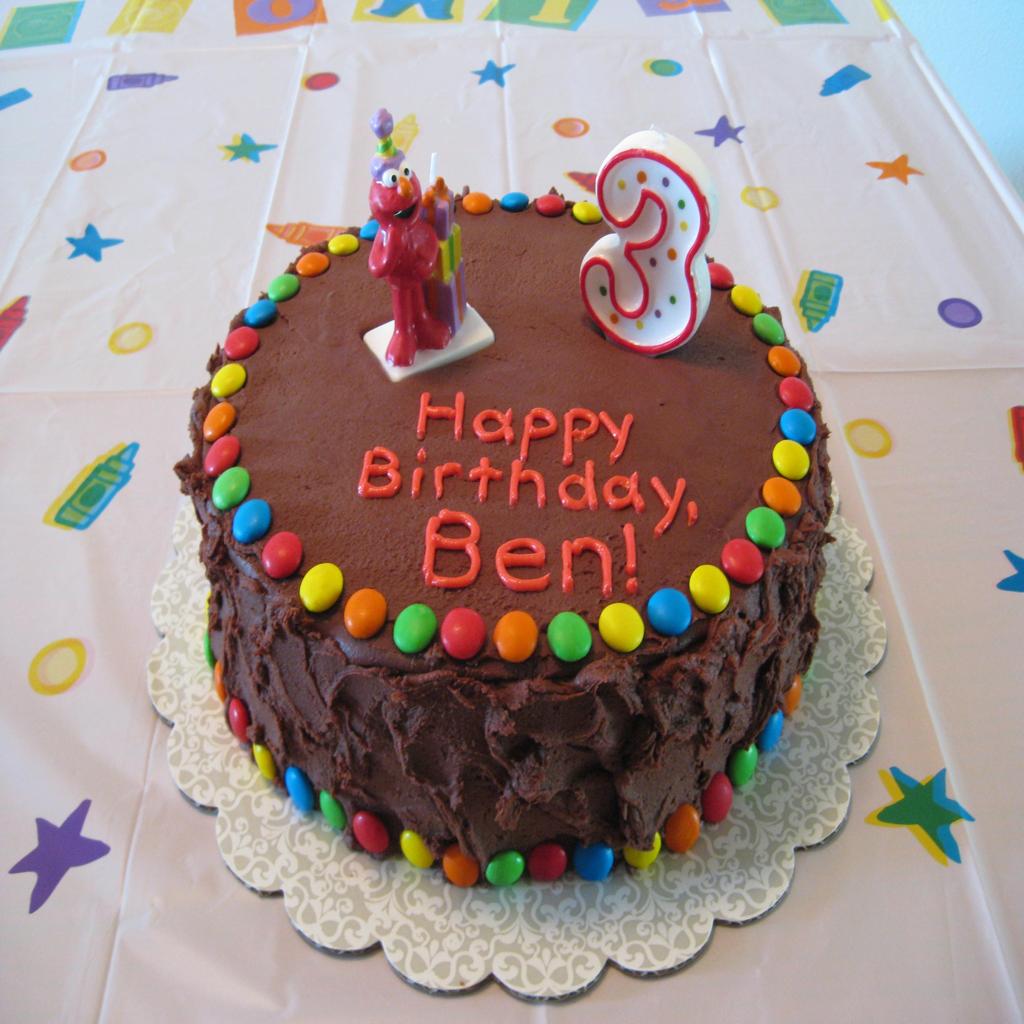}&
            \includegraphics[width=0.14\textwidth,height=0.09\textheight]{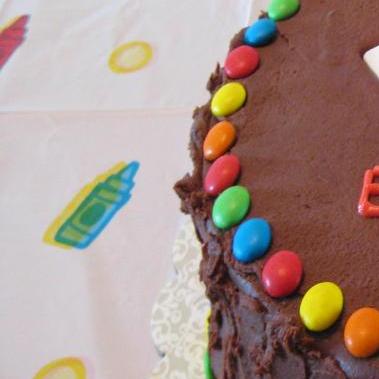}&
            \includegraphics[width=0.14\textwidth,height=0.09\textheight]{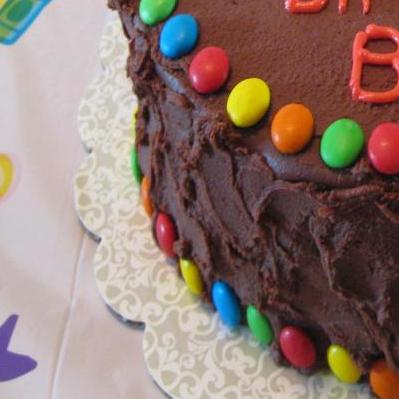}\\
            (a) &(b) &(c)\\
            \includegraphics[width=0.14\textwidth,height=0.09\textheight]{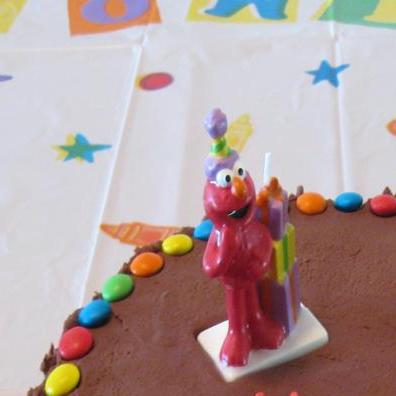}&
            \includegraphics[width=0.14\textwidth,height=0.09\textheight]{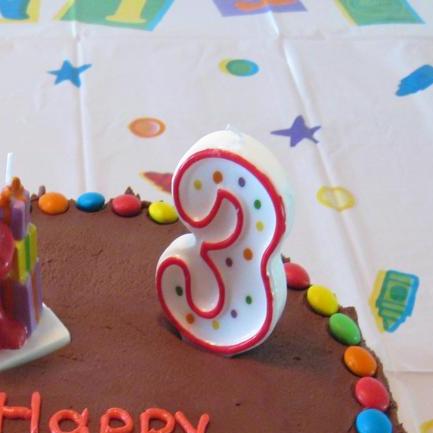}&
            \includegraphics[width=0.14\textwidth,height=0.09\textheight]{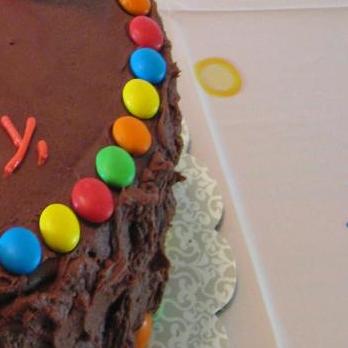}\\
            (d) &(e) &(f)\\
        \end{tabular}
        \vspace{0.8mm}
    \caption{Some sub-images produced by APS algorithm. (a) Original input image. (b)-(f) Typical sub-images produced by APS.}
 \vspace{-2mm}
    \label{fig:patch}
    \end{figure}
Compared with previous patch-based methods, our LRN has a notable difference. Traditional patch-based methods usually infer every patch in the image by sliding window or superpixels, which is extremely time-consuming% and also risky of generating false alarms
. We note that GSN has already succeeded to assign most pixels with right labels. Therefore, LRN only needs to focus on harder regions. Such a hierarchical prediction manner (GSN for easy regions and LRN for harder regions) makes our method more efficient and accurate. An Attended Patch Sampling (APS) scheme is proposed for this task. %and has not been explored by previous methods.
Guided by the results of GSN, it can generate sub-images attended to uncertain regions. Algorithm \ref{alg:crop} presents a rough procedure of APS (More details can be seen in supplementary material.).
%In Algorithm \ref{alg:crop}, w
We use the attention map $\bm{A}_i$ to indicate all uncertain pixels and it can be formulated as:
\begin{equation}\label{Ai}
%\label{equ:equ2}
\bm{A}_{i}(x,y)=
\left\{
\begin{aligned}
&1    &  T_1<\bm{F}_i(x,y)<T_2\\
&0    &otherwise
\end{aligned}
\right.
\end{equation}
In Algorithm \ref{alg:crop}, $w$ denotes the width of non-zero area in $\bm{A}_i$. $X_L$ and $X_R$ are the $x$ coordinates of the leftmost and rightmost non-zero pixels in $\bm{A}_i$. $n$ is a constant, which controls the overlapping between different patches. $r$ is a random numbers for generating sub-images with varied sizes.
%
%We find that our APS is insensitive to above parameters on different datasets.
We have performed grid search for setting these hyper-parameters and found that the results were not sensitive to their specific choices.
Therefore, we set them empirically in this work.
We set $D=384, n = 5$, $T_1$ = 50, $T_2$ = 200, and $r\in [-\frac{D}{6}, \frac{D}{6}]$ in all our experiments. Some image patches produced by APS are shown in Figure \ref{fig:patch}.

\subsection{Global-Local Fusion Network}\label{44}

As illustrated in above sections, GSN and LRN are inherently complementary with each other. Our method leverages GSN to classify easy regions and LRN to refine harder ones.
\begin{algorithm}[t]
\caption{Attended Patch Sampling.}
\begin{algorithmic}[1]
\label{alg:crop}
\REQUIRE RGB image $\bm{I}_i$, ground truth label $\bm{L}_i$, base cropping size $D$.
\ENSURE RGB patch set $\{\bm{P}_m^{\bm{I}_i}\}_{m=1}^M$, ground truth patch set $\{\bm{P}_m^{\bm{L}_i}\}_{m=1}^M$.
\STATE Generate attention map $\bm{A}_i$ from $\bm{F}_i$, as in Equ. \ref{Ai}.%Find out boundary set $\bm{B}\subset\bm{S}$.
\STATE $N_x=\lceil w/D\rceil+n$%Calculate\{$\bm{X^{B_0}}, \bm{X^{B_1}},...,\bm{X^{B_N-1}}$\} and diagonal length $\bm{L}$ of the bounding box.
%\STATE Select $\{\bm{B_0}, \bm{B_1},..., \bm{B_{N-1}}\}\subset\bm{B}$.
%\STATE Calculate cropping sizes $\{\bm{CS_0}, \bm{CS_1},..., \bm{CS_{N-1}}\}$  based on the diagonal length of the bounding box.
\FOR {$t = 1 , \ldots, N_x+1$}
\STATE $C = D + r$
\STATE $X_t=min\{X_L+(t-1)\times \lceil w/N_x\rceil, X_R\}$%Randomly select a pixel $\bm{c_i}(\bm{x_i},\bm{y_i})\in\bm{B_i}$.
%\FOR{$j = 1 , \ldots, N_x+1$}
\STATE $\bm{Y} = \{y\mid \bm{A}_i (X_t, y) = 1\}$
%\STATE D = $max\{Y\}$ - $min\{Y\}$
\STATE Pick out $J$ pixels $(X_t,y(j))_{j=1}^J$ from $(X_t,\bm{Y})$.
\STATE Taking $C$ as cropping size, $(X_t,y(j))_{j=1}^J$ as center pixels, crop $\{\bm{P}_j^{\bm{I}_i}\}_{j=1}^J$ and $\{\bm{P}_j^{\bm{L}_i}\}_{j=1}^J$ from $\bm{I}_i$ and $\bm{L}_i$, respectively.
\ENDFOR
\end{algorithmic}
\end{algorithm}
Then the final predictions can be obtained by fusing their results. A simple way to do this is to replace the saliency values of uncertain regions in $\bm{F}_i$ (the result of GSN) by $\{\bm{R}_m^{\bm{I}_i}\}_{m=1}^M$ (the result of LRN). Overlapped areas will be averaged. However, this kind of fusion lacks spatial consistency and does not leverage rich details in original high-resolution images.   %It will be much better if we can incorporate high-resolution information to help the fusion of GSN and LRN.

\begin{table*}
\begin{center}
\small
\setlength{\tabcolsep}{4.6pt}
\begin{tabular}{c||c|c|c|c|c|c|c|c|c|c|c|c|c|c|c|}
\toprule[2.5pt]
\multirow{2}{*}{Method}      &\multicolumn{3}{c|}{HRSOD-Test}&\multicolumn{3}{c|}{DAVIS-S}  & \multicolumn{3}{c|}{DUTS-Test}&  \multicolumn{3}{c|}{HKU-IS} & \multicolumn{3}{c|}{THUR}\\
\cline{2-16}
& $F_\beta$&S-m&MAE&$F_\beta$&S-m&MAE&$F_\beta$&S-m&MAE& $F_\beta$&S-m&MAE&$F_\beta$&S-m&MAE\\
\midrule[1pt]
%LEGS & & &&0.563&0.708&0.113 &0.585&0.694&0.138&0.723&0.557&0.119&0.607&0.722&0.125\\
%MDF & & &&0.702&0.807&0.061 &0.673& 0.732&0.094&-&-&-&0.636&0.733&0.109\\
RFCN~\cite{wang2016saliency}&0.530&0.608&0.121&0.728&0.842&0.062 &0.712 &0.792&0.091&0.835&0.746&0.079&0.627&0.793&0.100\\
%DCL & & &&0.743&0.765&0.081 &0.714&0.735&0.088&0.853&0.729&0.072&0.676&0.736&0.181\\
DHS~\cite{liu2016dhsnet} &0.746&0.848&0.059&0.774&0.865&0.034&{0.724}&0.817&{0.067}&0.855&0.746&0.053&0.673&0.803&0.082\\
UCF~\cite{zhang2017learning} &0.700&0.819&0.095&0.648&0.827&0.080&0.629&0.778&0.117&0.808&0.747&0.074&0.645&0.785&0.112\\
Amulet~\cite{zhang2017amulet}&0.717&0.830&0.075&0.755&0.848&0.042&0.676&0.803&0.085&0.839&0.772&0.052&0.670&0.797&0.094\\
NLDF~\cite{luo2017non}&0.763&0.853&0.055&0.718&0.858&0.042&0.743&0.815&0.066&0.874&0.770&0.048&0.697&0.801&0.080\\
DSS~\cite{hou2017deeply}&0.756&0.840&0.060&0.728&0.865&0.041&0.791&0.822&{0.057}&\textbf{0.895}&0.779&0.041&0.731&0.801&0.073\\
%BMP &0.779&0.883&0.046&0.775&\textbf{0.901}&0.030&{0.751}&0.861&\textbf{0.049}&0.871&0.907&0.038&0.690&0.819&0.079\\
RAS\cite{chen2018reverse}&0.773&0.842&0.058&0.763&0.867&0.038&0.755&0.839&0.060&0.871&\textbf{0.887}&0.045&0.696&0.787&0.082\\
DGRL~\cite{wang2018detect}&0.789&0.848&0.053&0.772&0.859&0.038&0.768&\textbf{0.841}&\textbf{0.051}&0.882&0.802&\textbf{0.037}&0.716&0.816&0.077\\
%PAGR& 0.816&0.882&0.046&0.816&0.899&0.037&0.788&0.837&0.056&0.886&0.791&0.048&0.729&\textbf{0.830}&0.070\\
DGF~\cite{wu2018fast}&0.795&0.824&0.058&0.785&0.847&0.037&0.776&0.803&0.062&0.893&0.869&0.043&0.734&0.799&0.070\\
\midrule[1pt]
Ours-D%\footnote{This version is trained on DUTS.}
&0.857&0.876&0.040&0.850&0.875&0.029&\textbf{0.796}&0.827&0.052&0.891&0.882&0.042&0.740&0.820&0.067\\
%Ours\footnote{This version is trained on DUTS and finetune on HRSOD-training.}&\textbf{0.886}&\textbf{0.897}&\textbf{0.030}&\textbf{0.879}&0.879&\textbf{0.026}&0.786&0.829&0.050&0.880&\textbf{0.882}&0.041&0.742&\textbf{0.827}&\textbf{0.065}\\
Ours-DH%\footnote{This version is trained on DUTS and finetune on HRSOD-training.}
&\textbf{0.888}&\textbf{0.897}&\textbf{0.030}&\textbf{0.888}&\textbf{0.876}&\textbf{0.026}&0.791&0.822&\textbf{0.051}&0.886&0.877&0.042&\textbf{0.749}&\textbf{0.826}&\textbf{0.064}
\\
\bottomrule[1.5pt]
\end{tabular}
\end{center}
\caption{Quantitative comparisons with other state-of-the-arts in term of F-measure (larger is better) and MAE (smaller is better) on five dataset. The best results are shown in bold.}%\label{tab:fm}
%in terms of maximum F-measure and MAE.}
\label{tab-state}
\vspace{-4mm}
\end{table*}
We propose to directly train a network to incorporate high-resolution information to help the fusion of GSN and LRN. To maintain all the high-resolution details from images, this network should not include any pooling layers or convolutional layers with large strides. %for this fusion task.
With limited GPU memory, popular backbones (\emph{e.g.}, VGG and ResNet) can not be trained with such a high-resolution input size (more than $1000\times1000$ pixels). Therefore, We propose a light-weighted network, name as Global-Local Fusion Network (GLFN). As shown in Figure \ref{glfn}, high-resolution RGB images and combined maps from GSN and LRN are concatenated together to be the inputs of GLFN. GLFN consists of some convolution layers with dense connectivity as in \cite{huang2017densely}. We set the growth rate $g$ to be 2 for saving memory. Similar to \cite{huang2017densely}, we let the bottleneck layers ($1\times1$ convolution) produce $4g$ feature maps. On the top of these densely connected layers, we add four dilated convolutional layers to enlarge receptive field. All the dilated convolutional layers have the same kernel size and output channels, \emph{i.e.}, $k=3$ and $c=2$. The rates of the four dilated convolutional layers are set with $dilation=1, 6, 12, 18$ respectively. At last, a $3\times3$ convolution is appended for final prediction. What is worth mentioning is that our proposed GLFN %does not include any pooling layers or convolutional layers with large strides, so that  What's more, it
has an extremely small model size (\emph{i.e.}, 11.9 kB). %thanks to very few channels.
%-------------------------------------------------------------------------

\section{Experiment}
\subsection{Experimental Setup}
\subsubsection{Datasets}\label{411}
\noindent\textbf{High-Resolution Saliency Detection Datasets.} %a $\bm{h}$igh-$\bm{r}$esolution $\bm{s}$aliency $\bm{d}$etection $\bm{s}$ets, named HRSDS. To our best knowledge, it is the first high-resolution
We mainly use our proposed HRSOD-Test to evaluate the performance of our method along with other state-of-the-art methods. To enrich the diversity, we also collect 92 images which are suitable for saliency detection from DAVIS~\cite{perazzi2016benchmark}, a densely annotated high-resolution video segmentation dataset. Images in this dataset are precisely annotated and have very high resolutions (\emph{i.e.,}$1920\times1080$). We ignore the categories of the objects and generate saliency ground truth masks for this dataset. For convenience, the collected dataset is named as DAVIS-S.

\noindent\textbf{Low-Resolution Saliency Detection Datasets.} In addition, we evaluate our method on three widely used benchmark datasets:
THUR~\cite{cheng2014salientshape}, HKU-IS~\cite{li2015visual} and DUTS~\cite{wang2017}. THUR and HKU-IS are large-scale datasets, with 6232 and 4447 images, respectively. DUTS is a large saliency detection benchmark, contains 5019 test images.
\subsubsection{Evaluation Metrics}
We use four metrics to evaluate all methods: Precision-Recall (PR) curves, $F_\beta$ measure, Mean Absolute Error (MAE) and structure-measure \cite{fan2017structure}. PR curves are generated by binarizing the saliency map with a varied threshold from 0 to 255, then comparing the binary maps with the ground truth. $F_\beta$ measure is defined as ${F_\beta } = \frac{{\left( {1 + {\beta ^2}} \right) \cdot \text{precision} \cdot \text{recall}}}{{{\beta ^2} \cdot \text{precision} + \text{recall}}}$. The precision and recall are computed under the threshold of twice the mean saliency value. $\beta^2$ is set to 0.3 as suggested in~\cite{achanta2009frequency} to emphasize precision. MAE measures the average error of saliency maps. Structure-measure simultaneously evaluates region-aware and object-aware structural similarity between a saliency map and a ground truth mask. For detailed implementations, we refer readers to \cite{fan2017structure}.
\begin{figure*}[!t]
    \centering
    \tabcolsep0.3mm \renewcommand{\arraystretch}{0.5}
        \begin{tabular}{cccccccccccc}
            \includegraphics[width=0.080\textwidth,height=0.08\textheight]{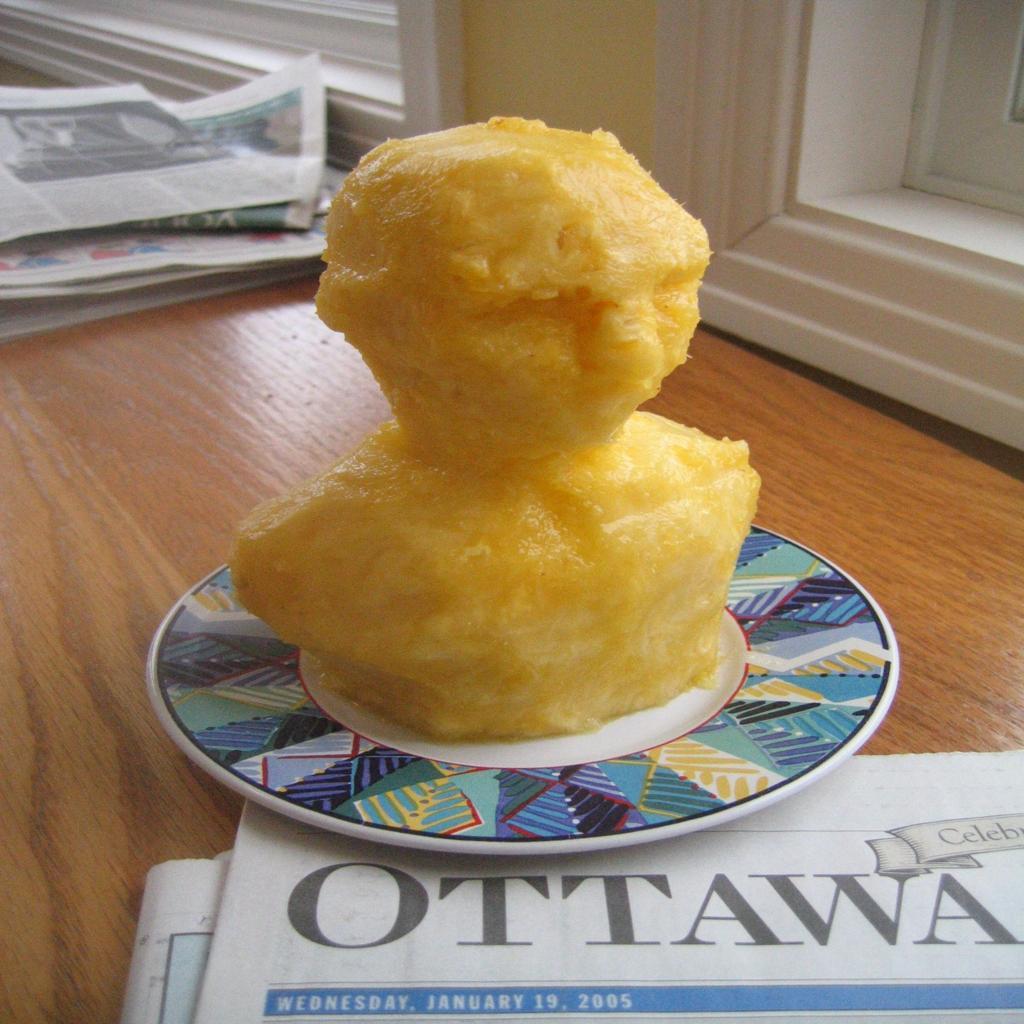}&
            \includegraphics[width=0.080\textwidth,height=0.08\textheight]{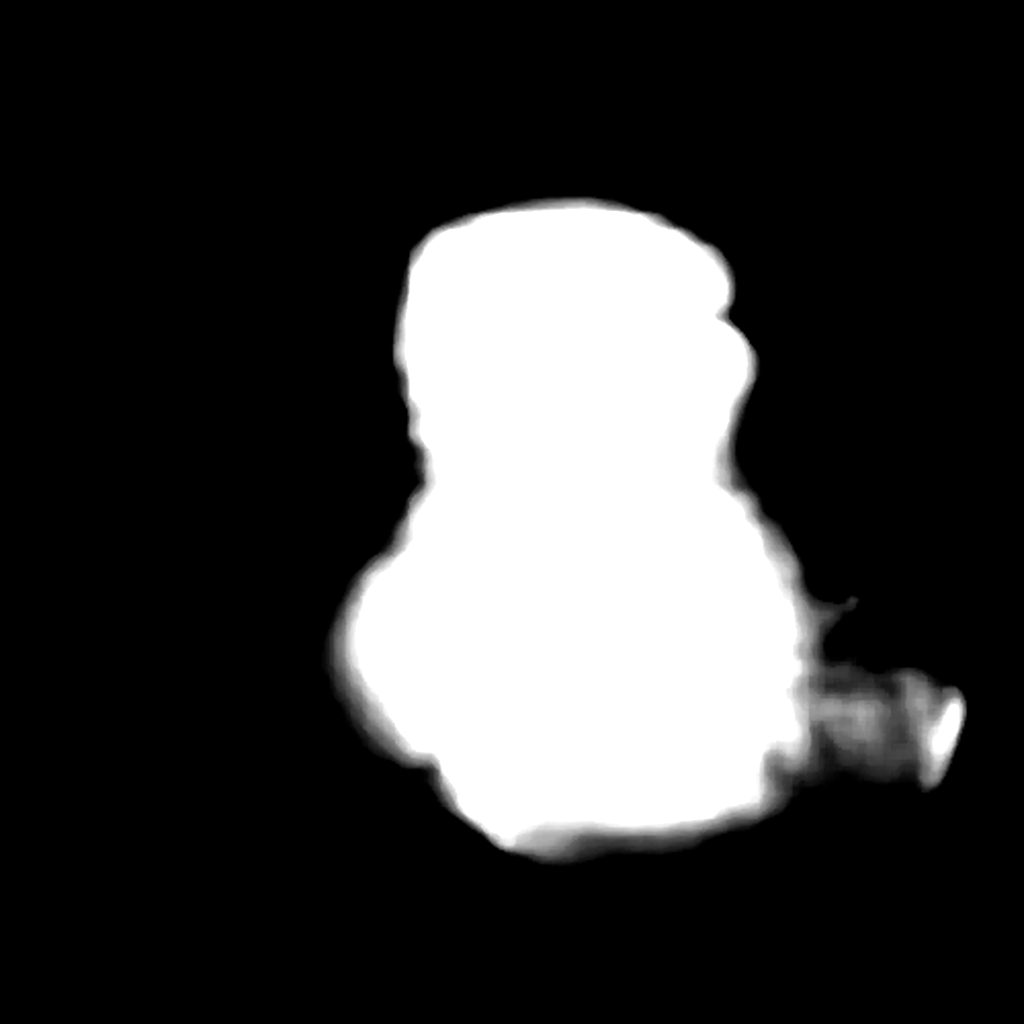}&
            \includegraphics[width=0.080\textwidth,height=0.08\textheight]{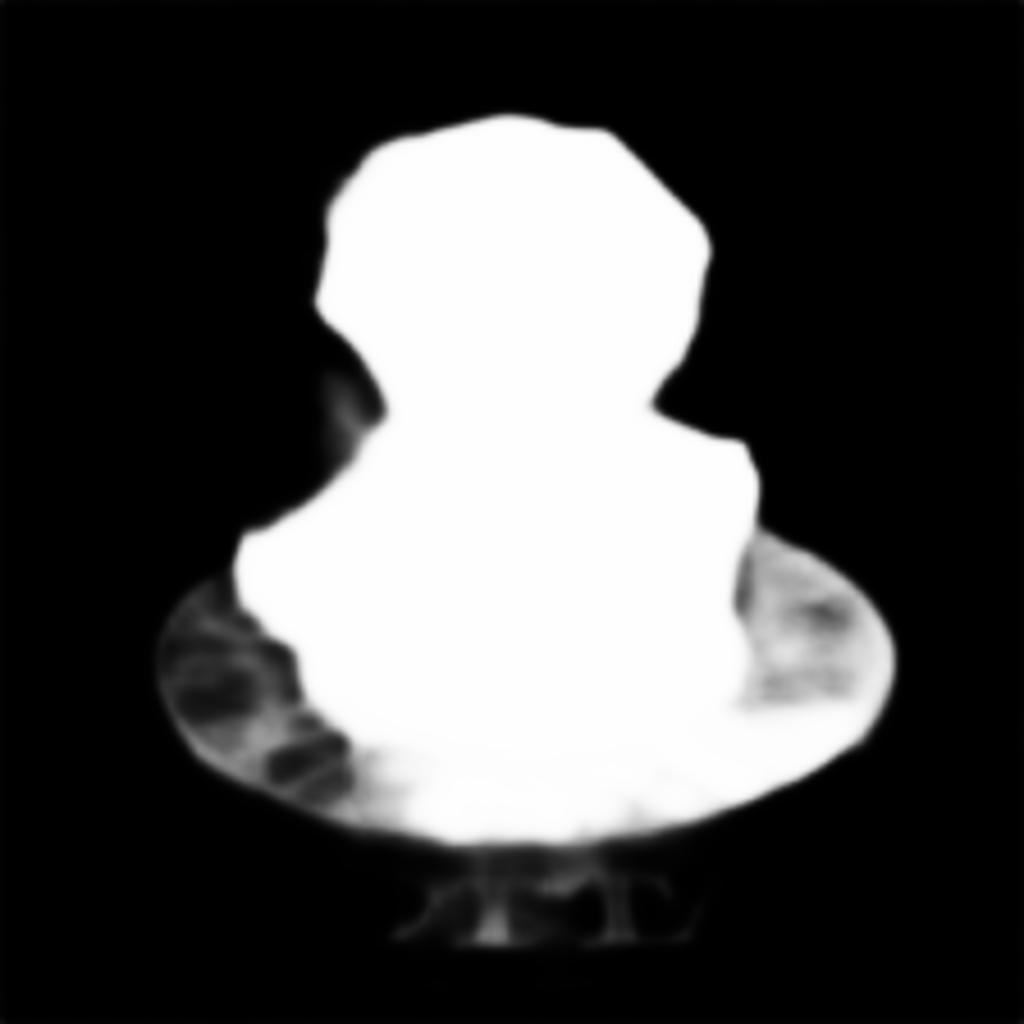}&
            \includegraphics[width=0.080\textwidth,height=0.08\textheight]{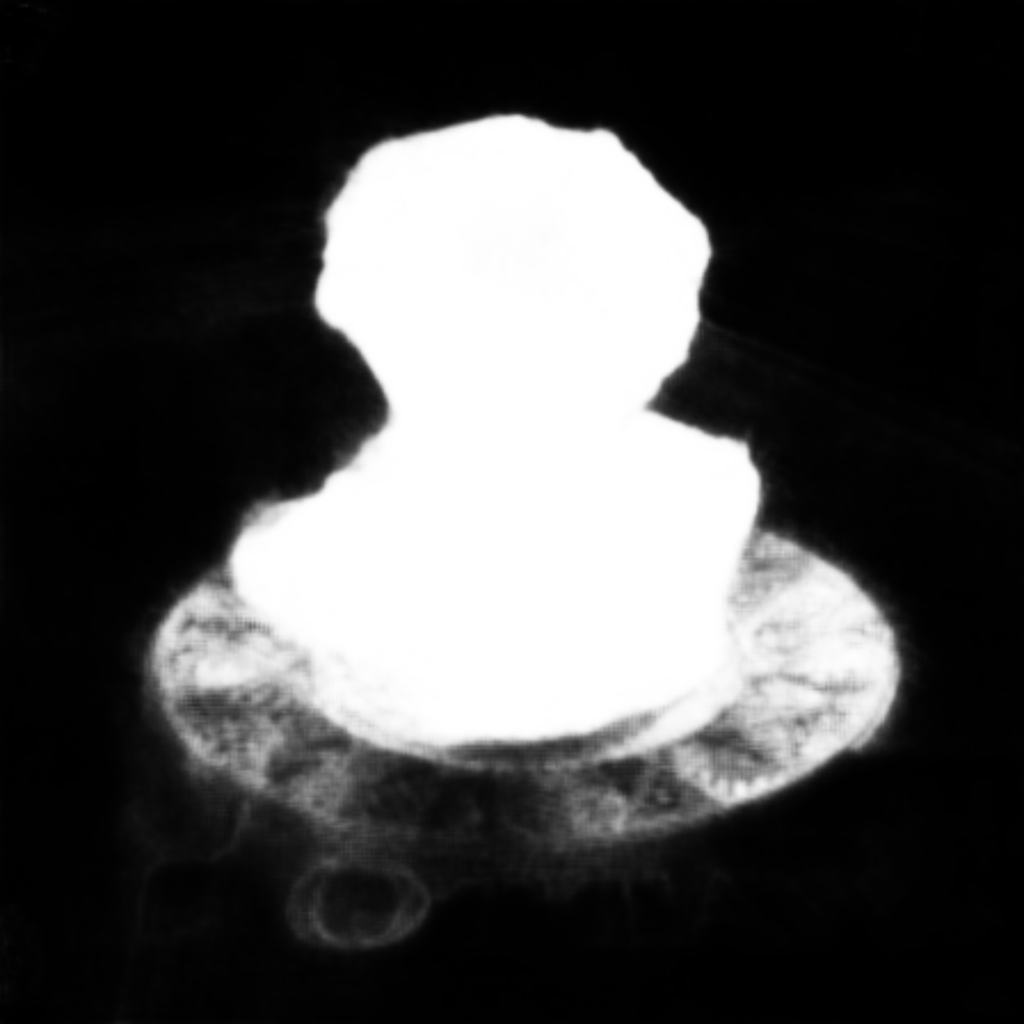}&
            \includegraphics[width=0.080\textwidth,height=0.08\textheight]{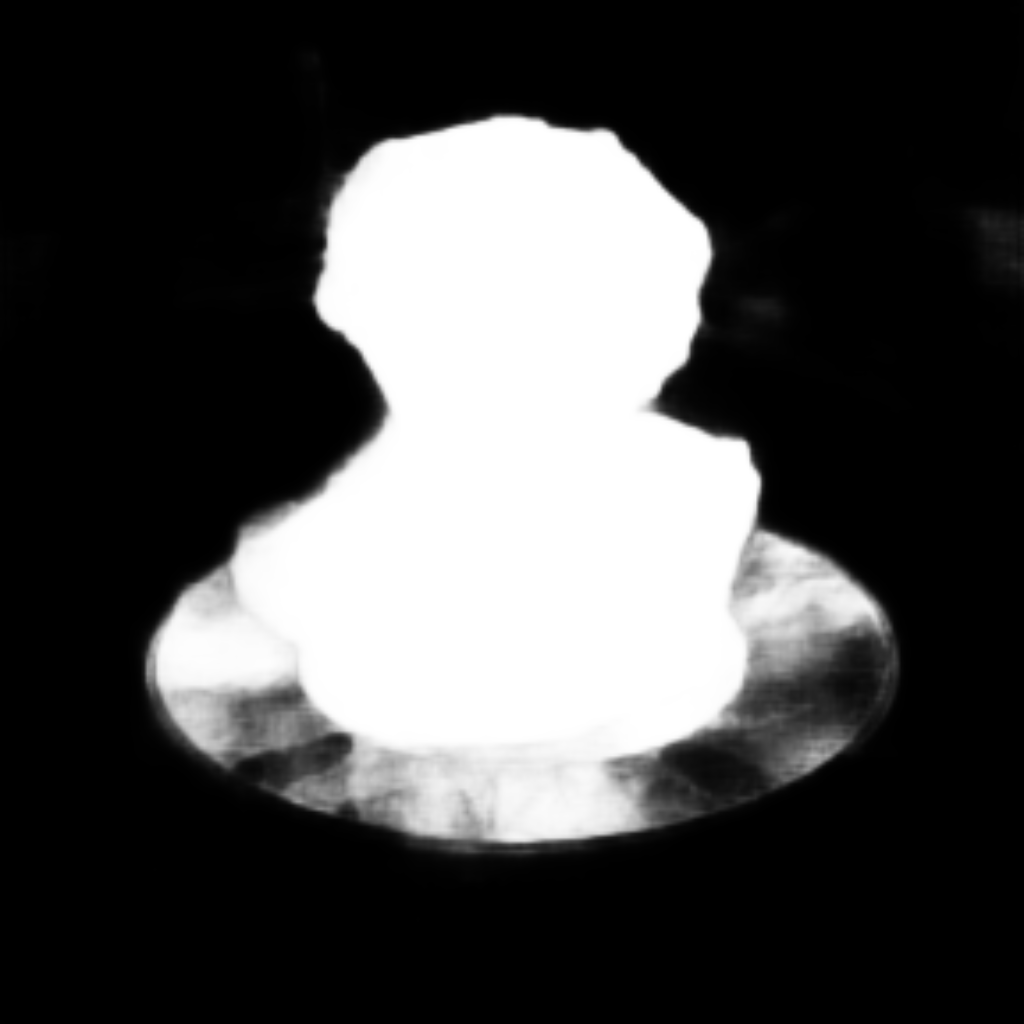}&
            \includegraphics[width=0.080\textwidth,height=0.08\textheight]{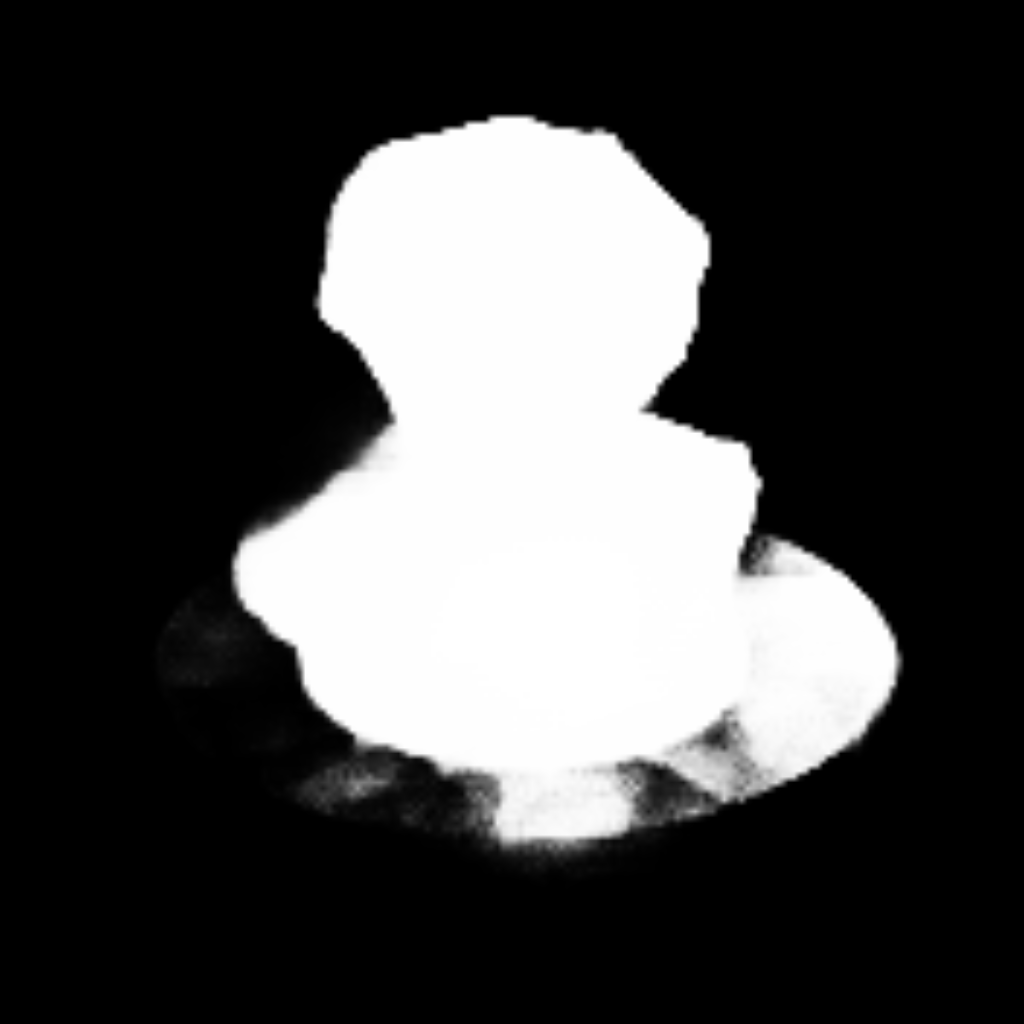}&
            \includegraphics[width=0.080\textwidth,height=0.08\textheight]{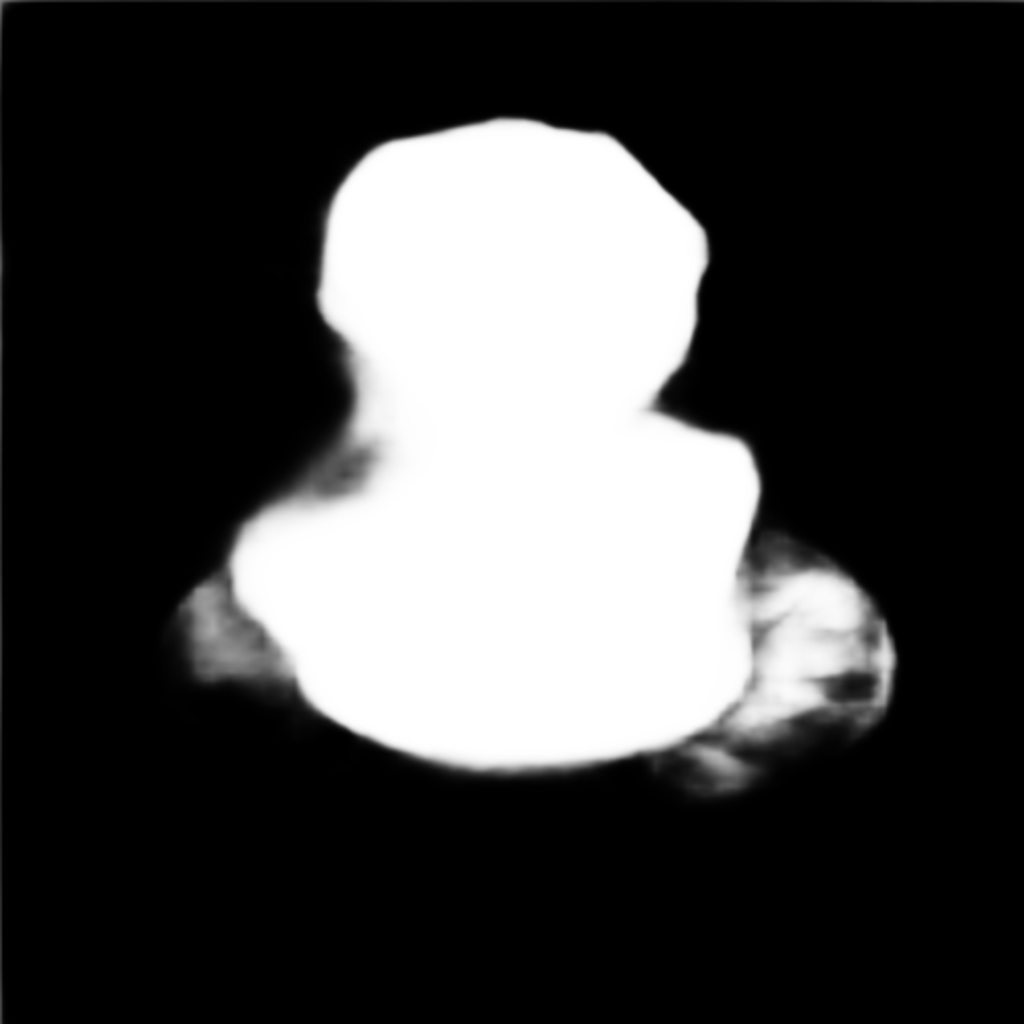}&
            \includegraphics[width=0.080\textwidth,height=0.08\textheight]{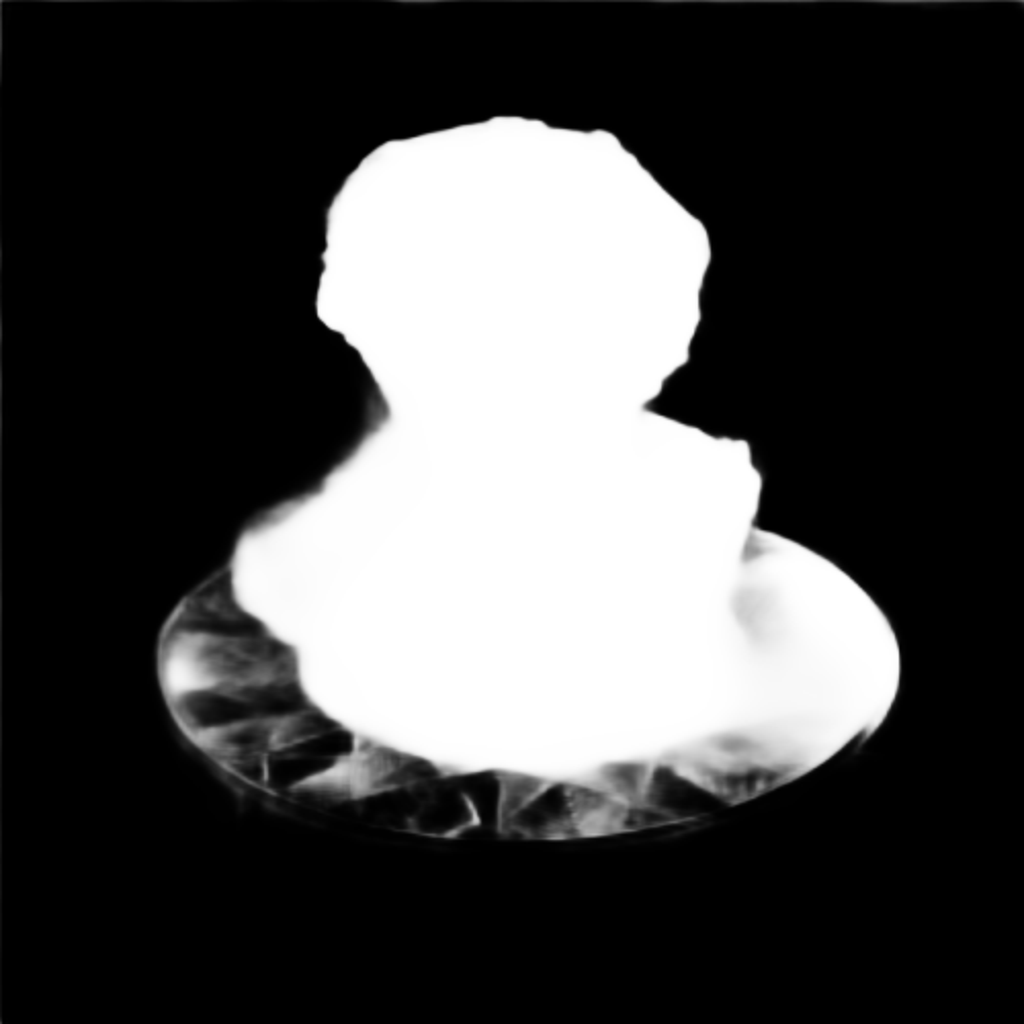}&
            \includegraphics[width=0.080\textwidth,height=0.08\textheight]{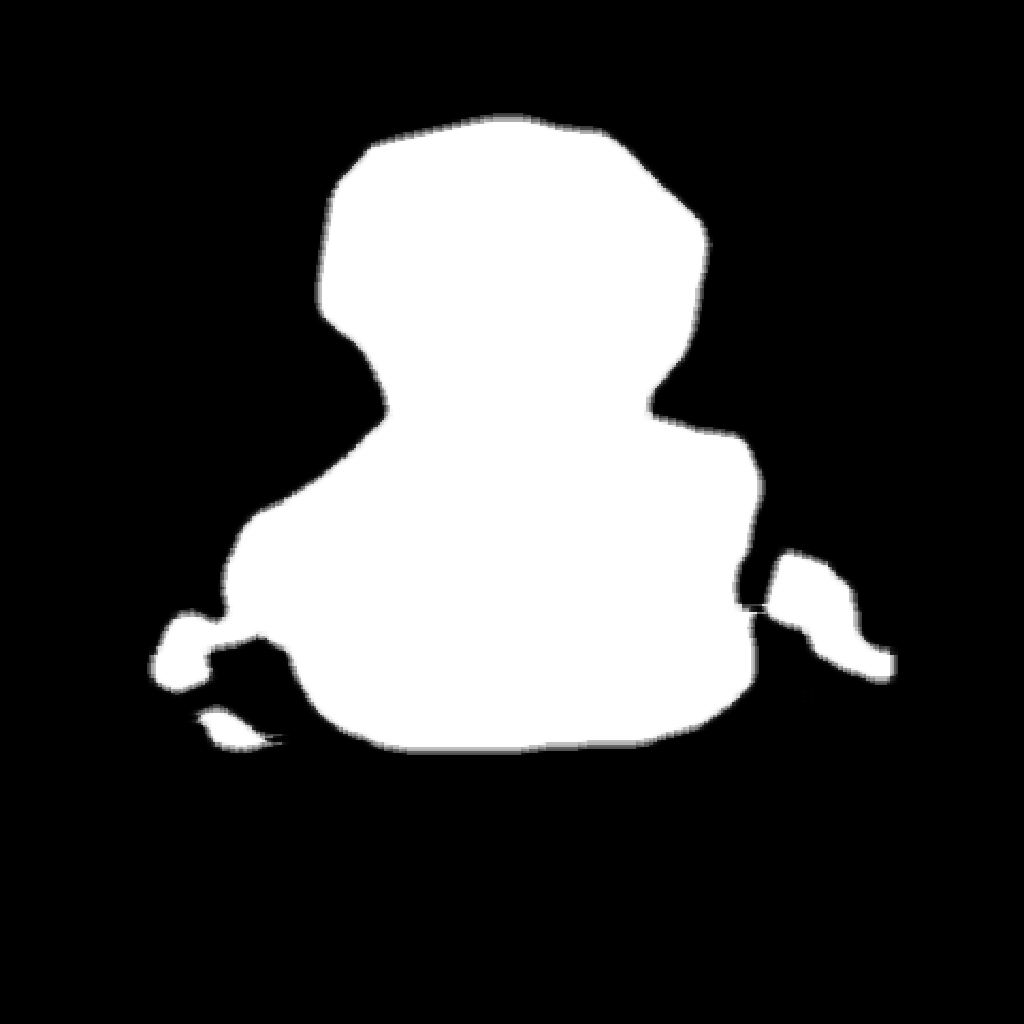}&
            \includegraphics[width=0.080\textwidth,height=0.08\textheight]{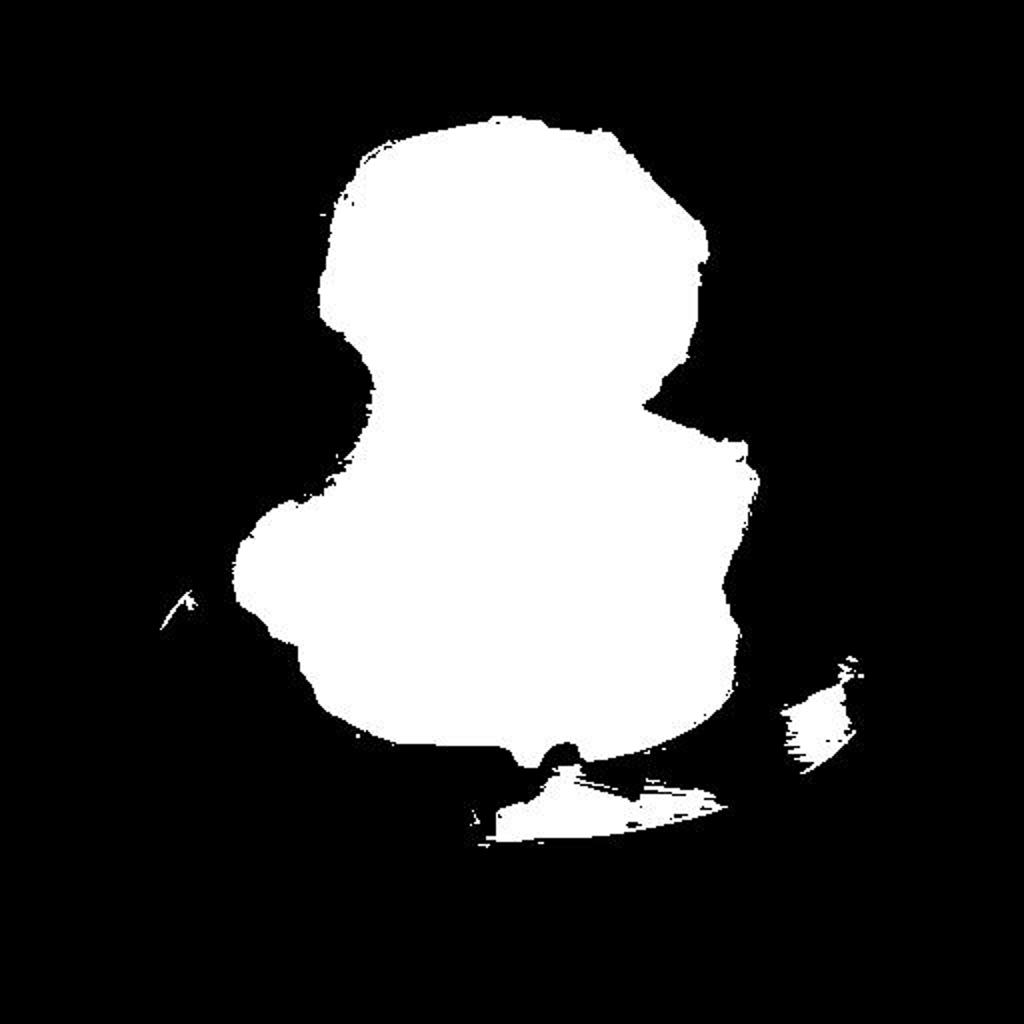}&
            \includegraphics[width=0.080\textwidth,height=0.08\textheight]{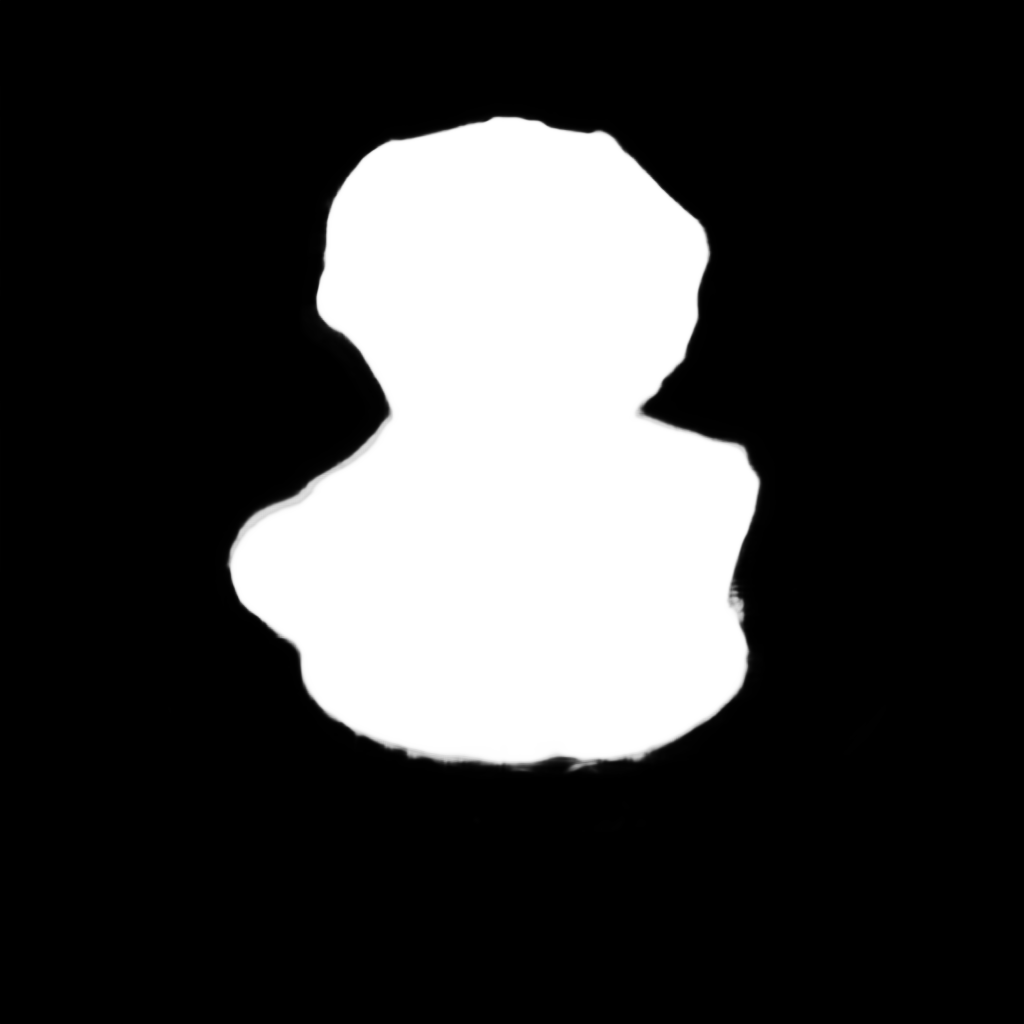}&
            \includegraphics[width=0.080\textwidth,height=0.08\textheight]{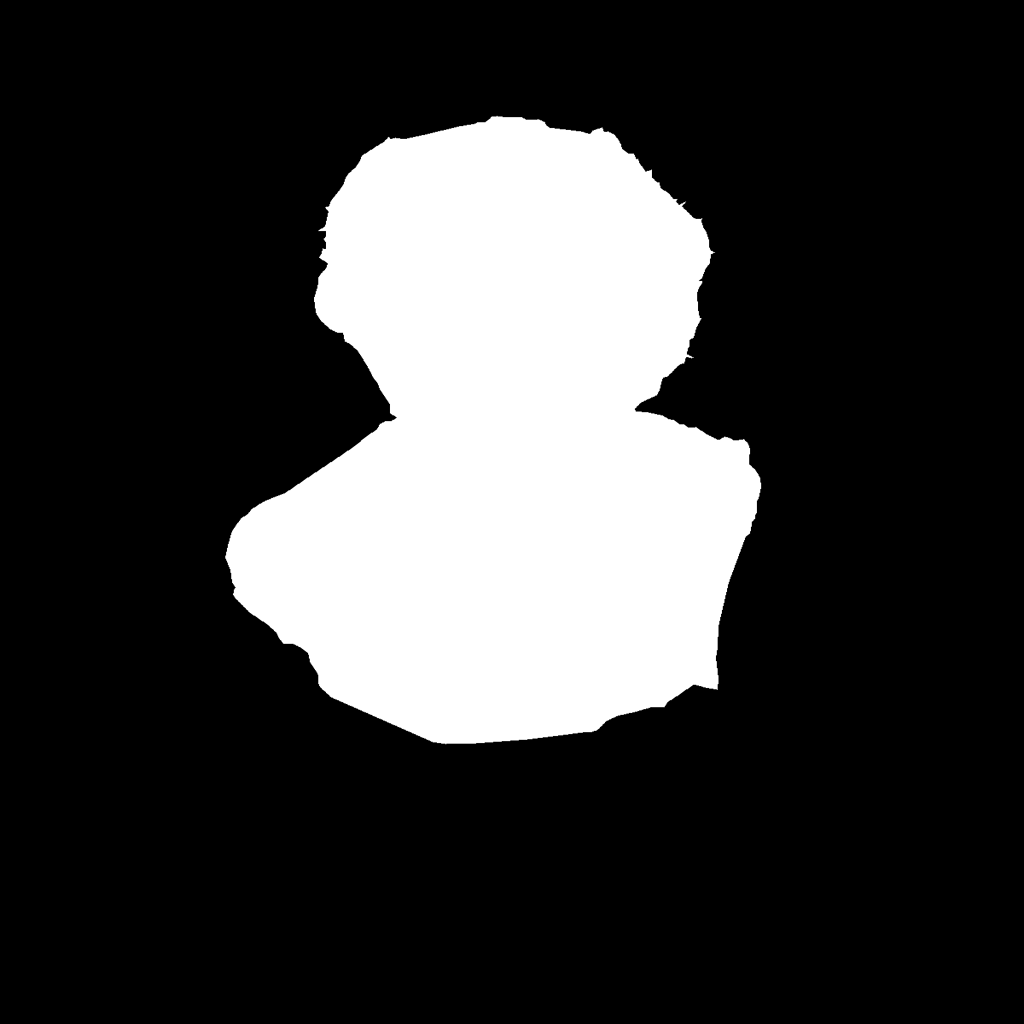}
            \\                       \includegraphics[width=0.080\textwidth,height=0.08\textheight]{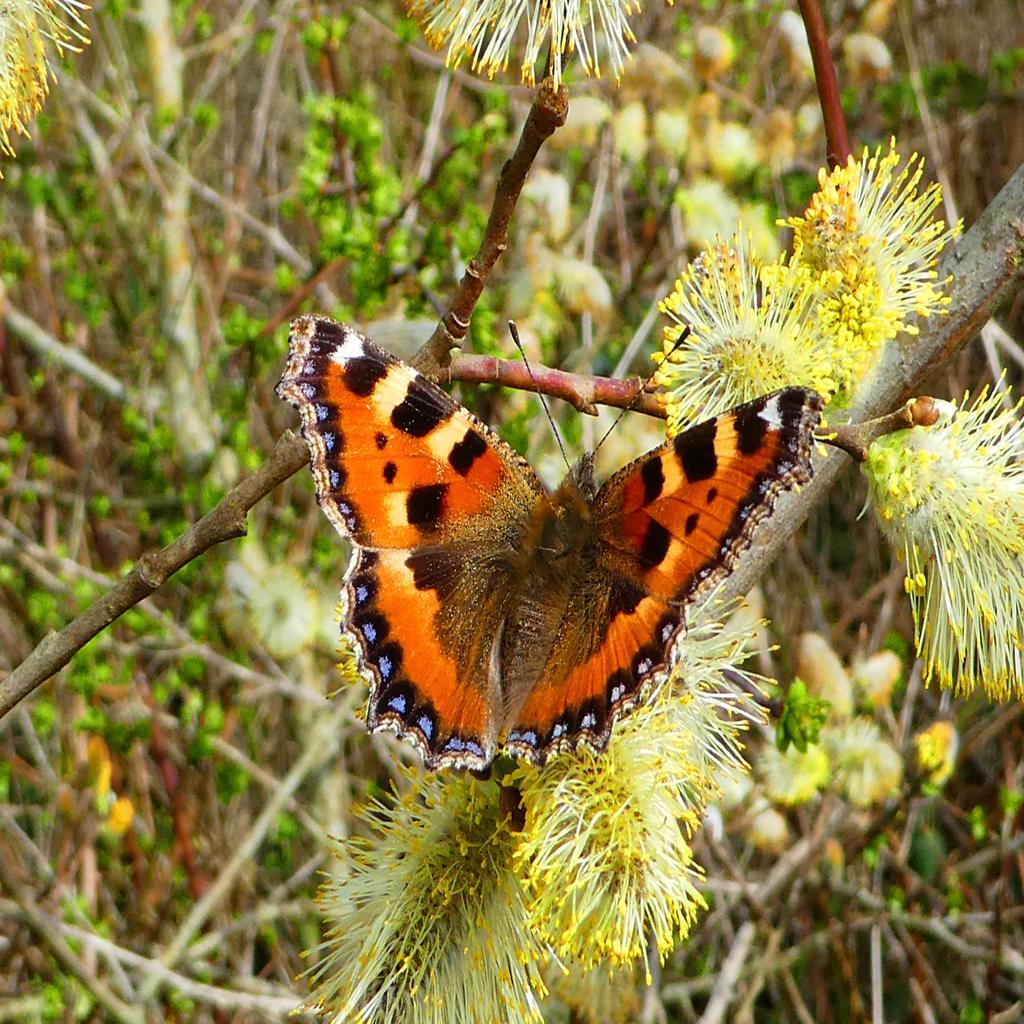}&
            \includegraphics[width=0.080\textwidth,height=0.08\textheight]{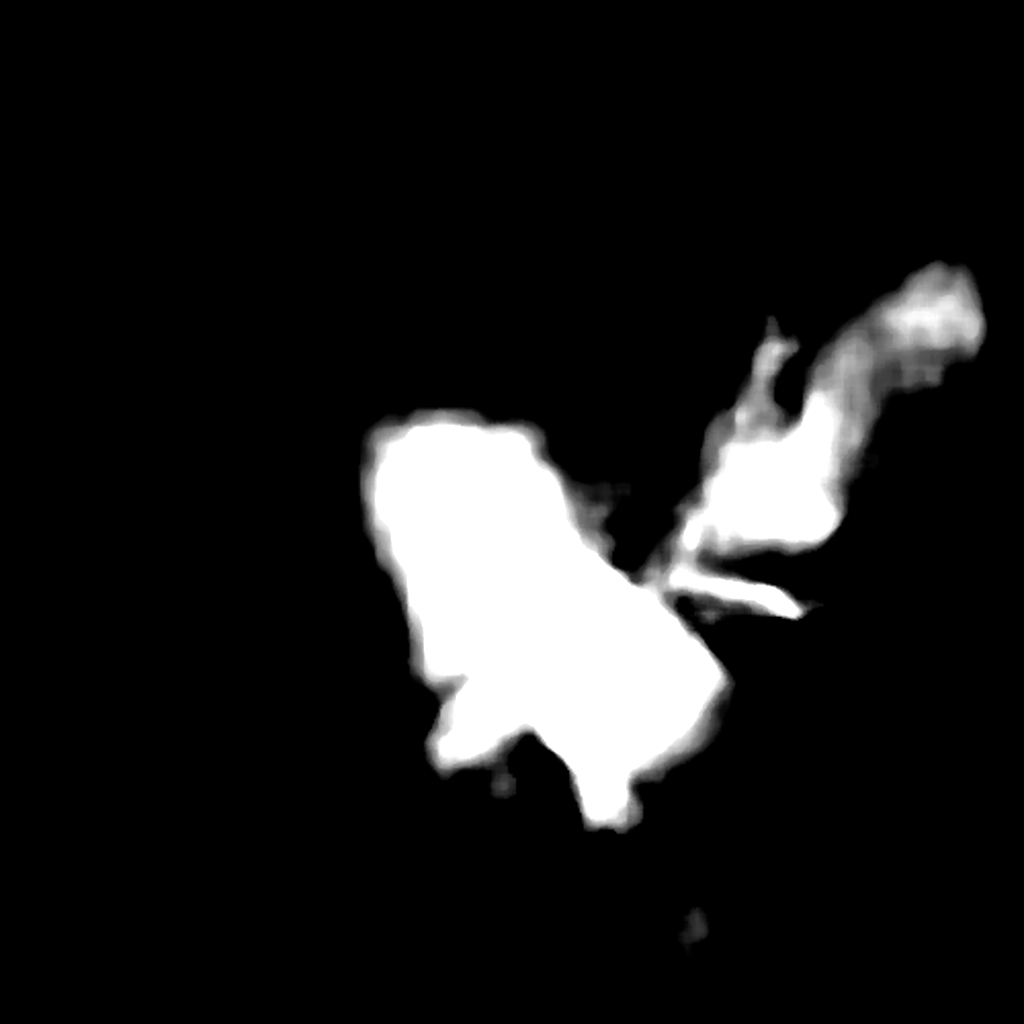}&
            \includegraphics[width=0.080\textwidth,height=0.08\textheight]{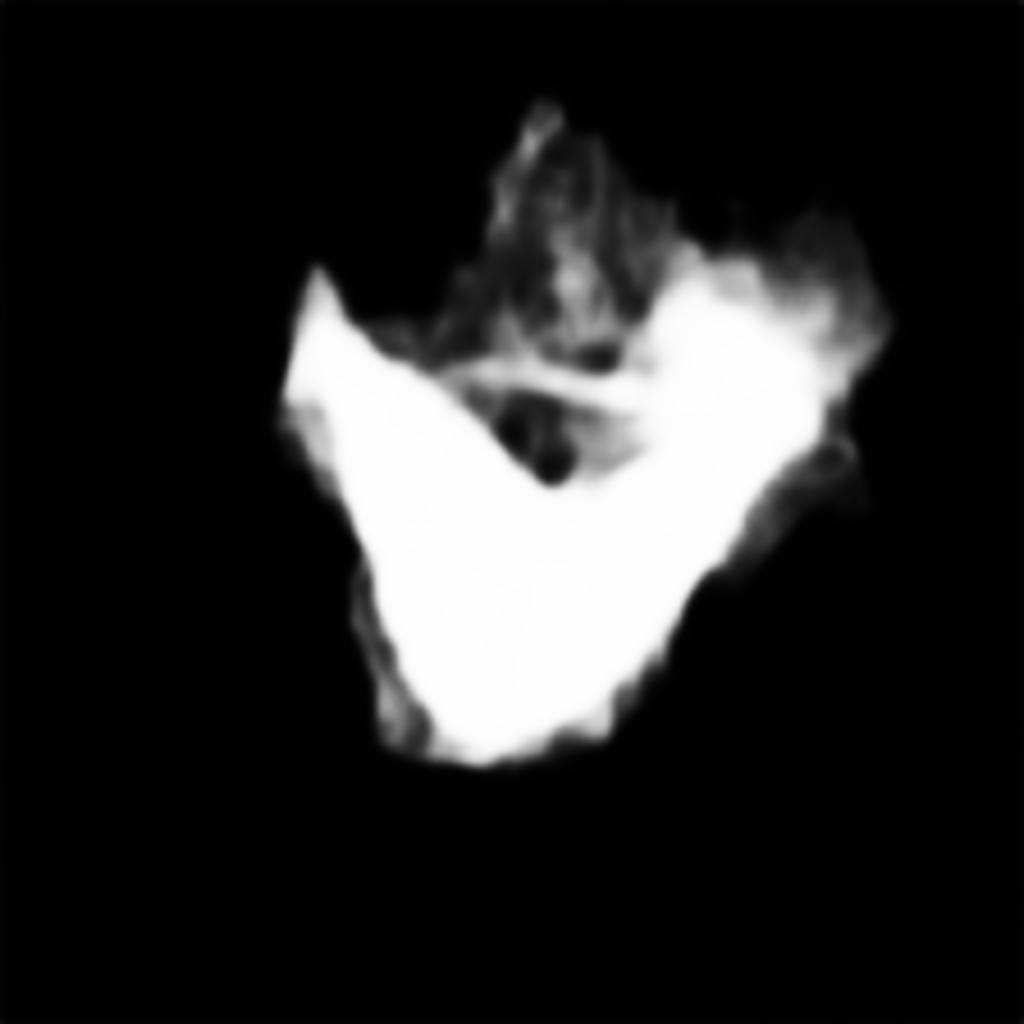}&
            \includegraphics[width=0.080\textwidth,height=0.08\textheight]{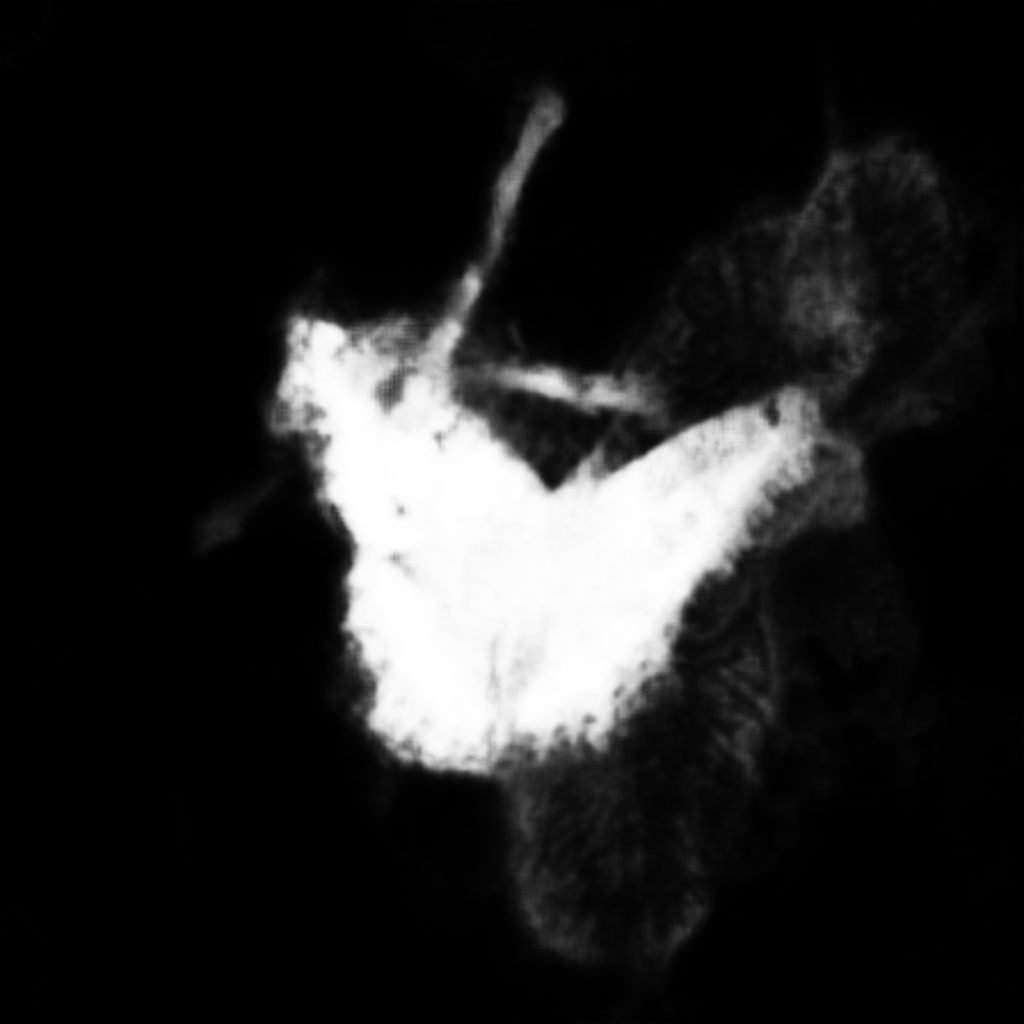}&
            \includegraphics[width=0.080\textwidth,height=0.08\textheight]{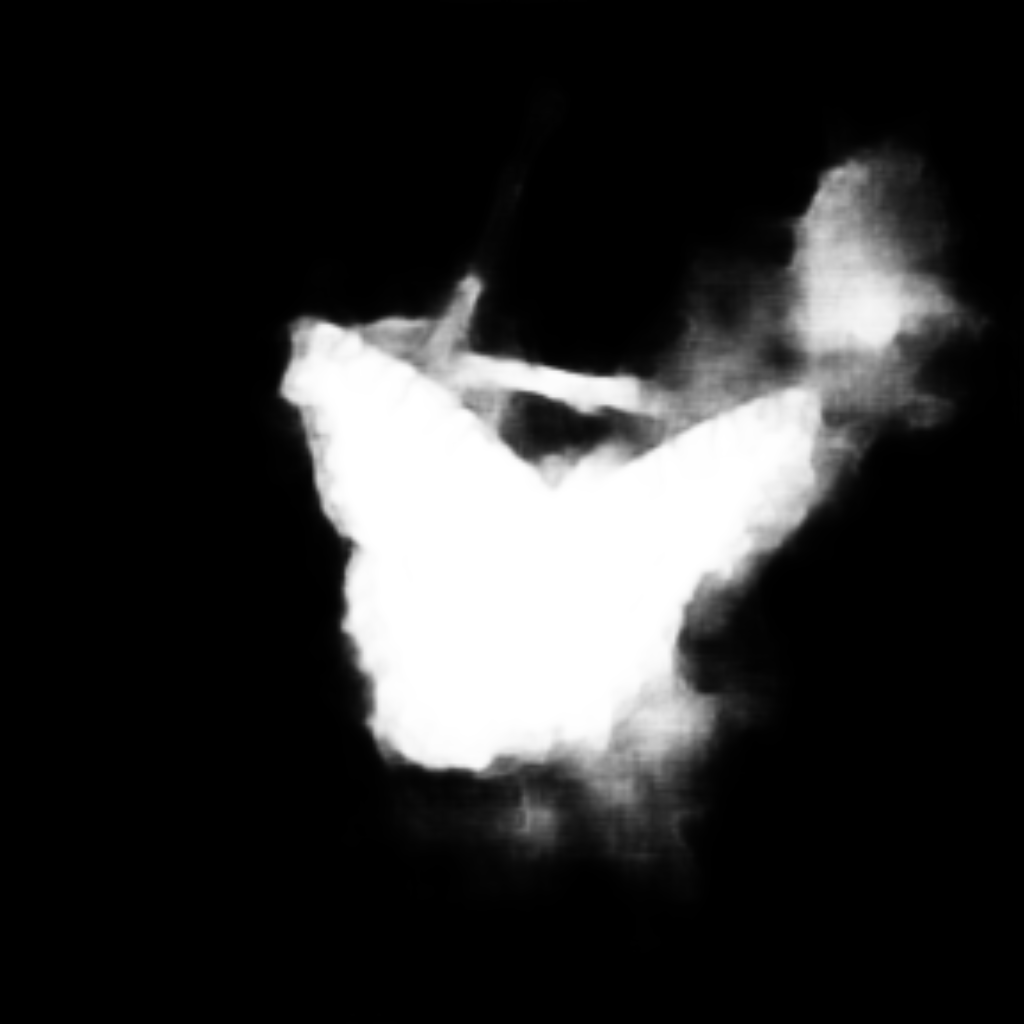}&
            \includegraphics[width=0.080\textwidth,height=0.08\textheight]{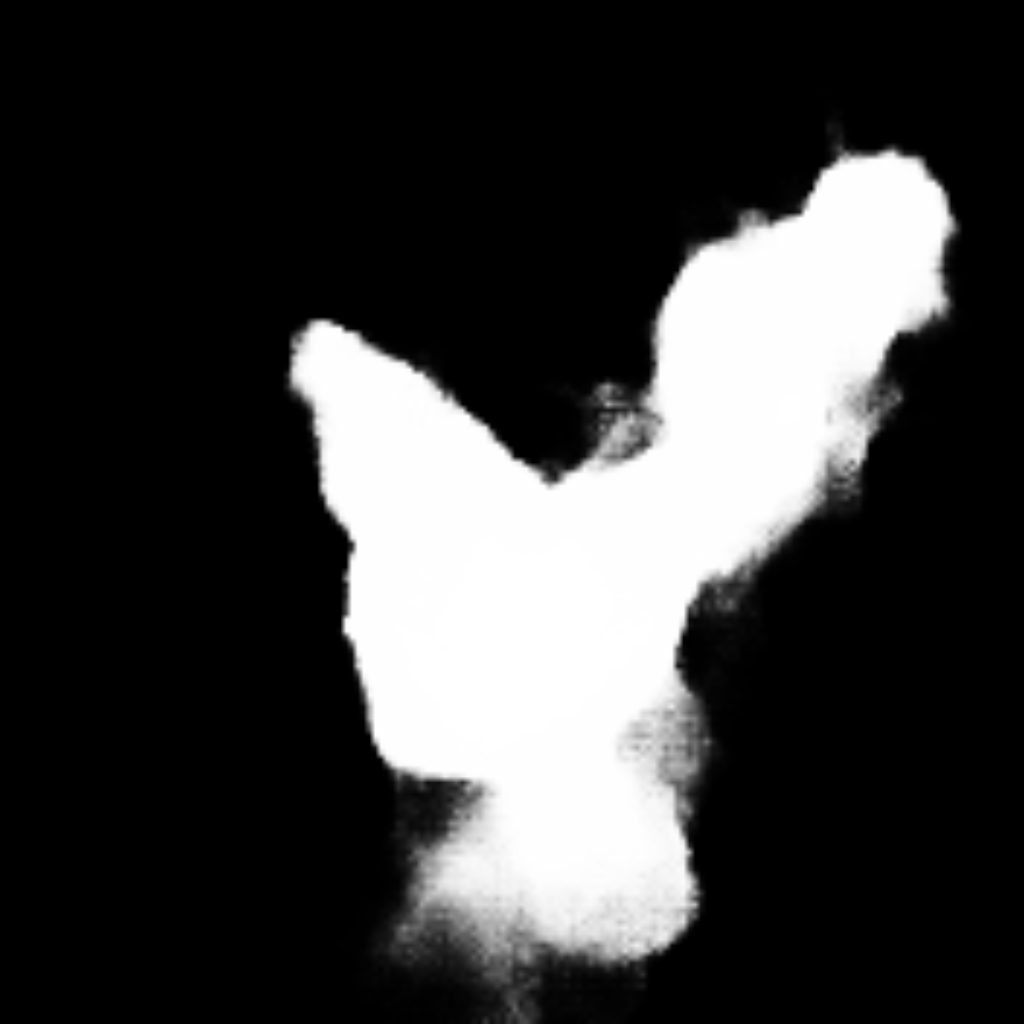}&
            \includegraphics[width=0.080\textwidth,height=0.08\textheight]{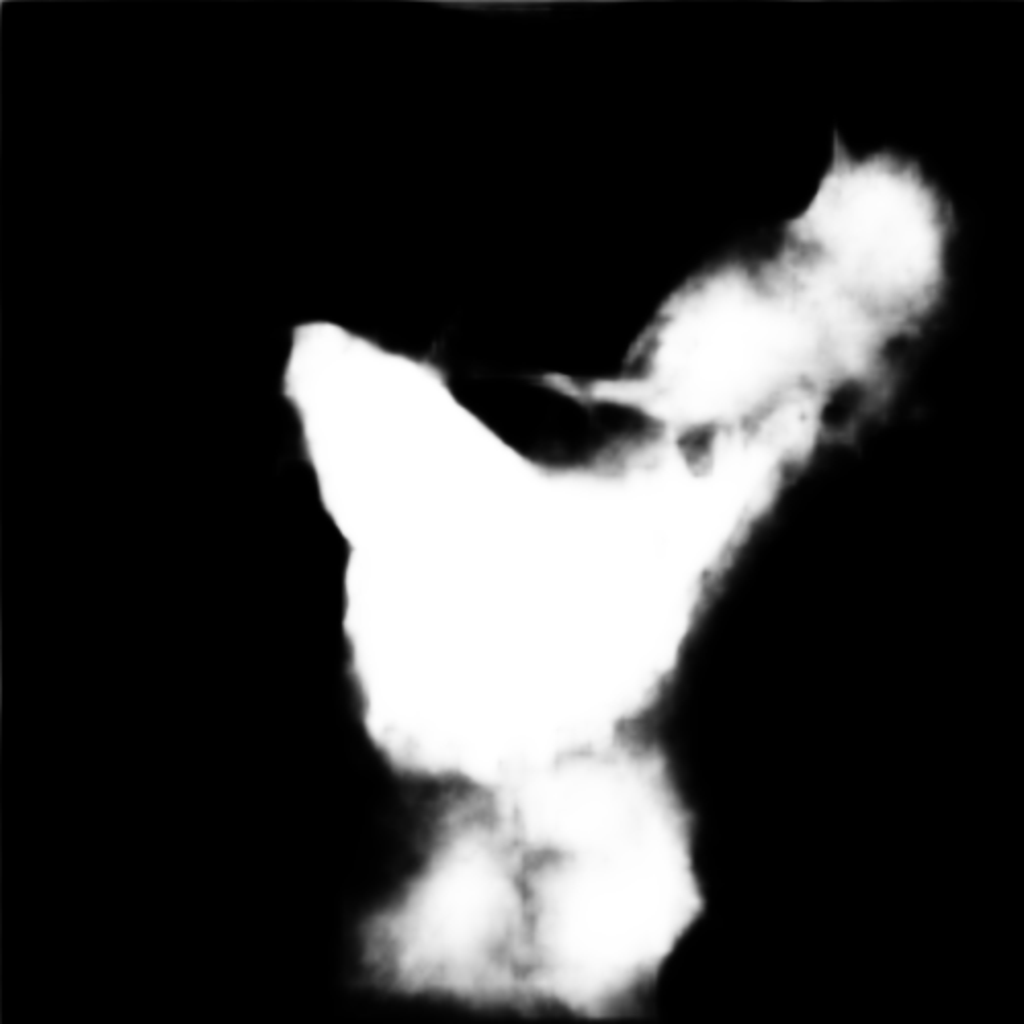}&
            \includegraphics[width=0.080\textwidth,height=0.08\textheight]{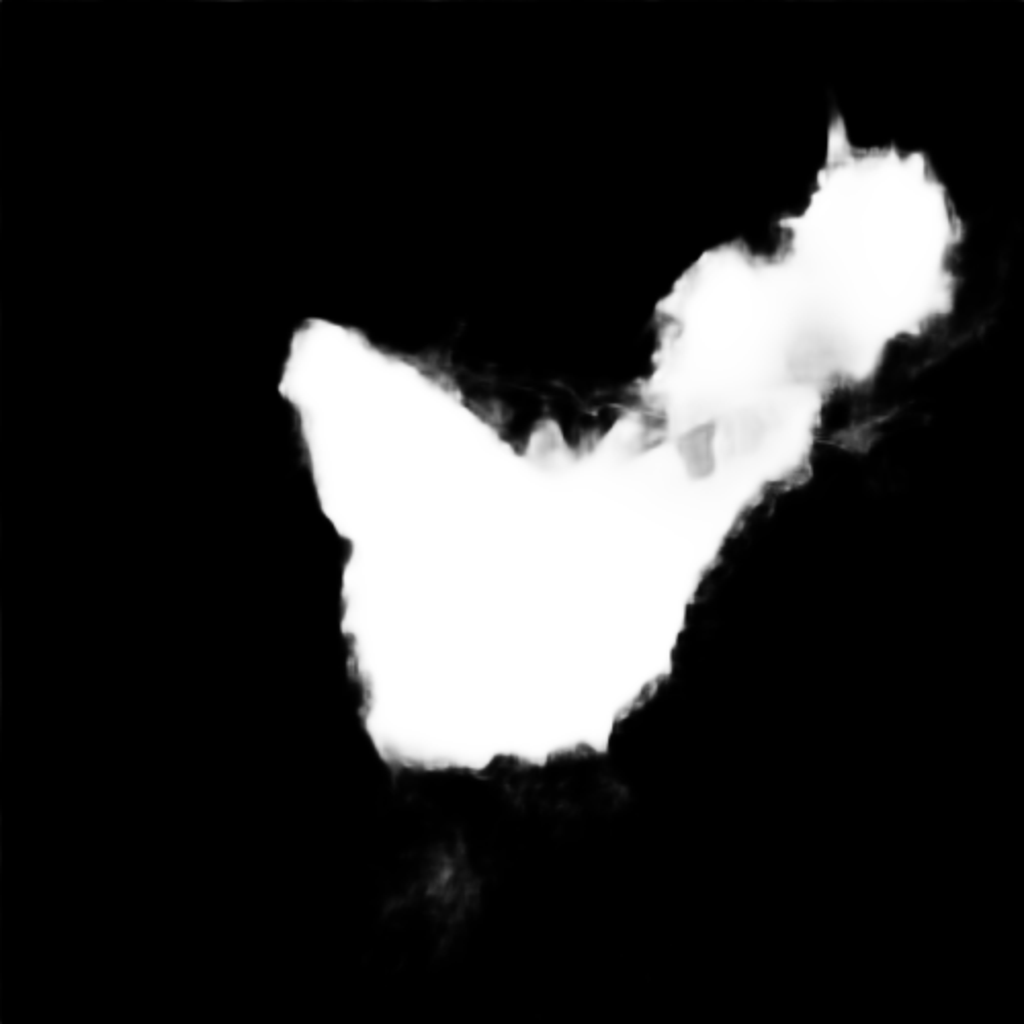}&
            \includegraphics[width=0.080\textwidth,height=0.08\textheight]{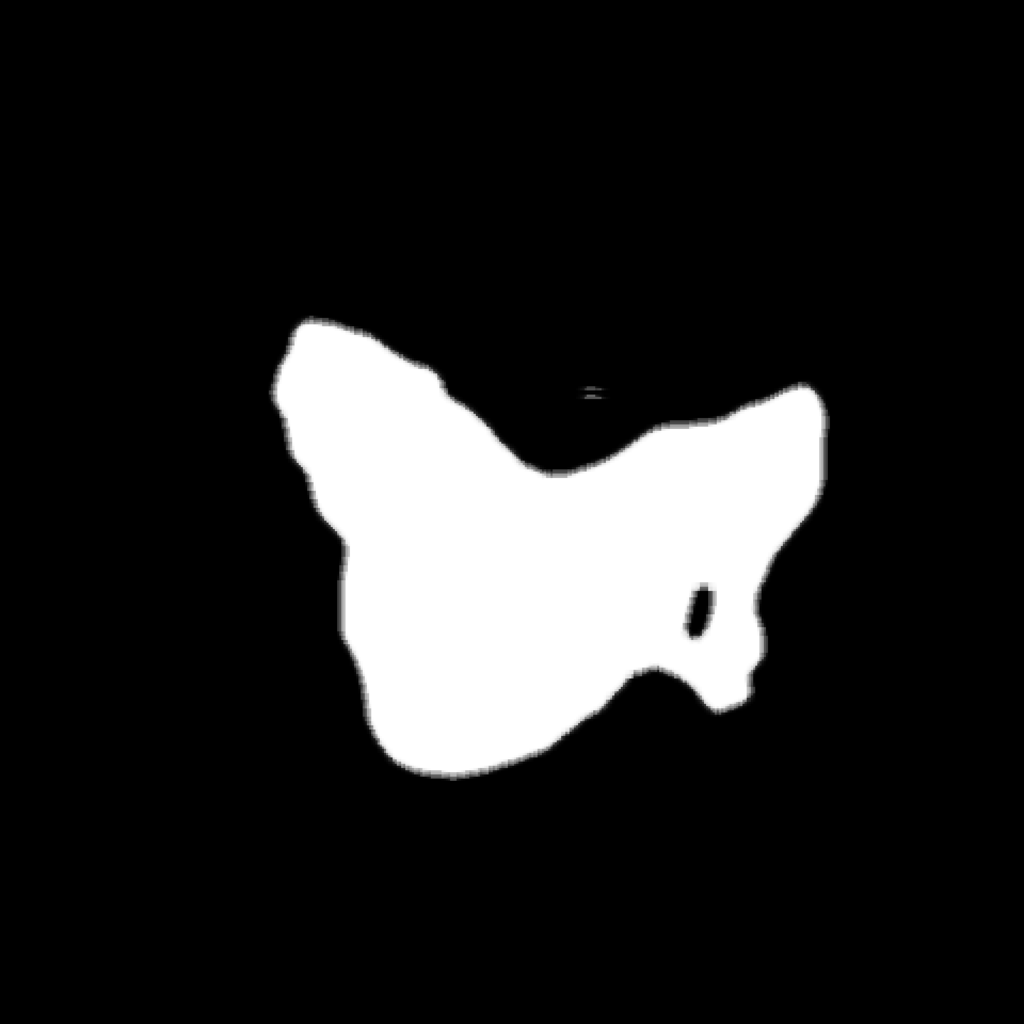}&
            \includegraphics[width=0.080\textwidth,height=0.08\textheight]{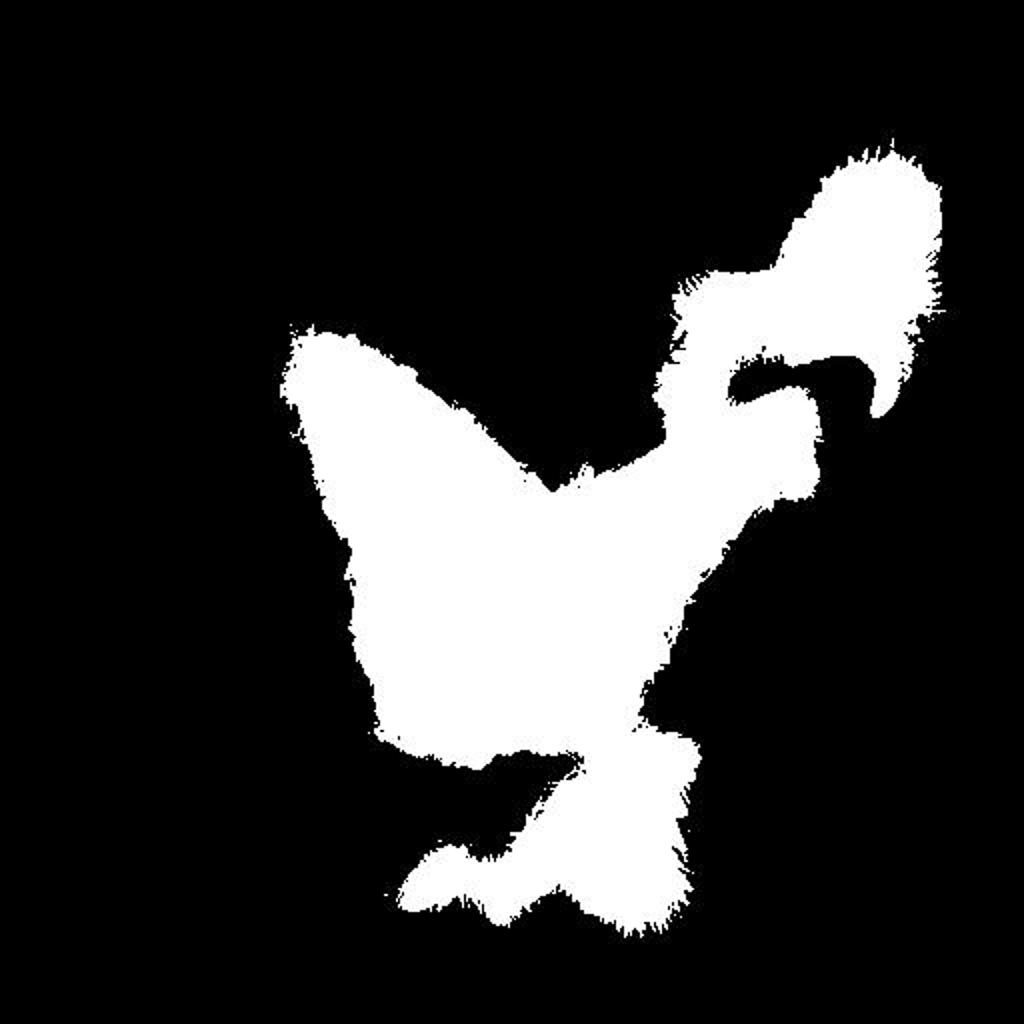}&
            \includegraphics[width=0.080\textwidth,height=0.08\textheight]{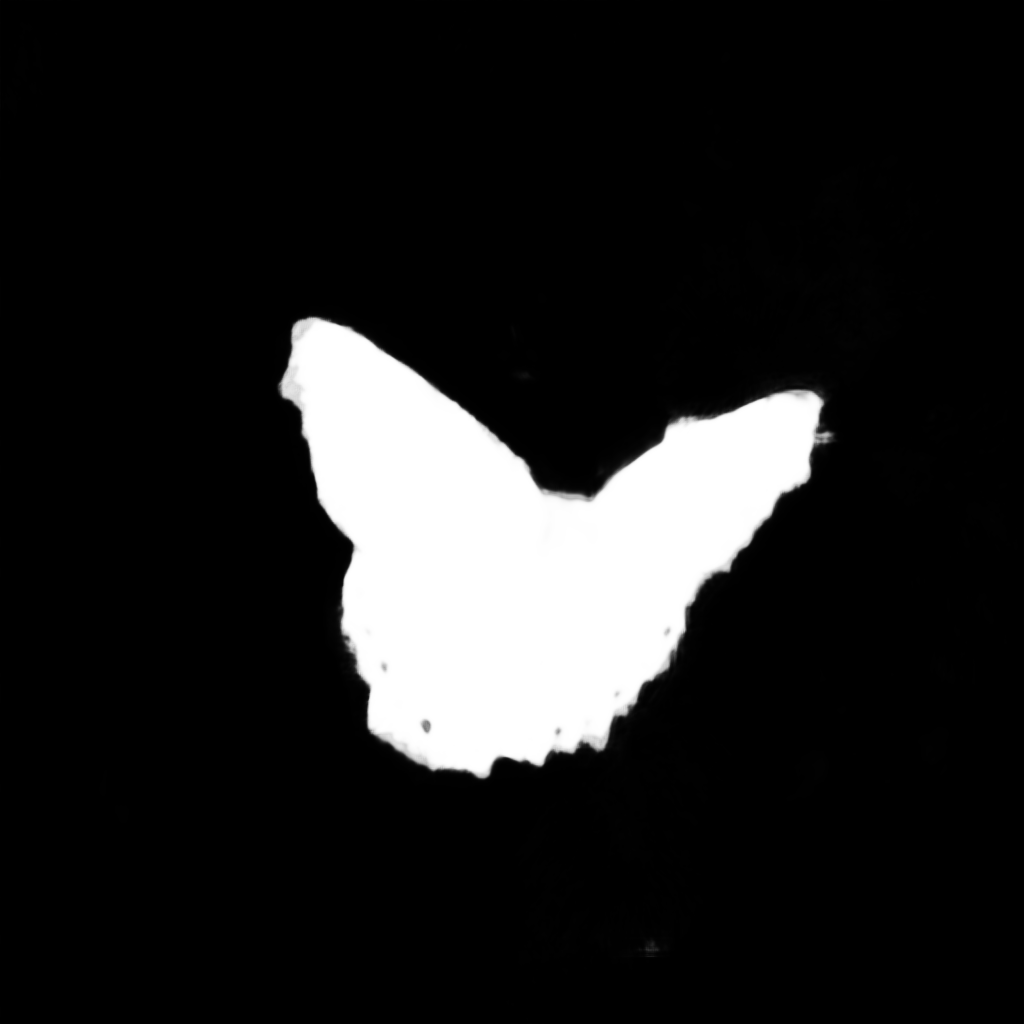}&
            \includegraphics[width=0.080\textwidth,height=0.08\textheight]{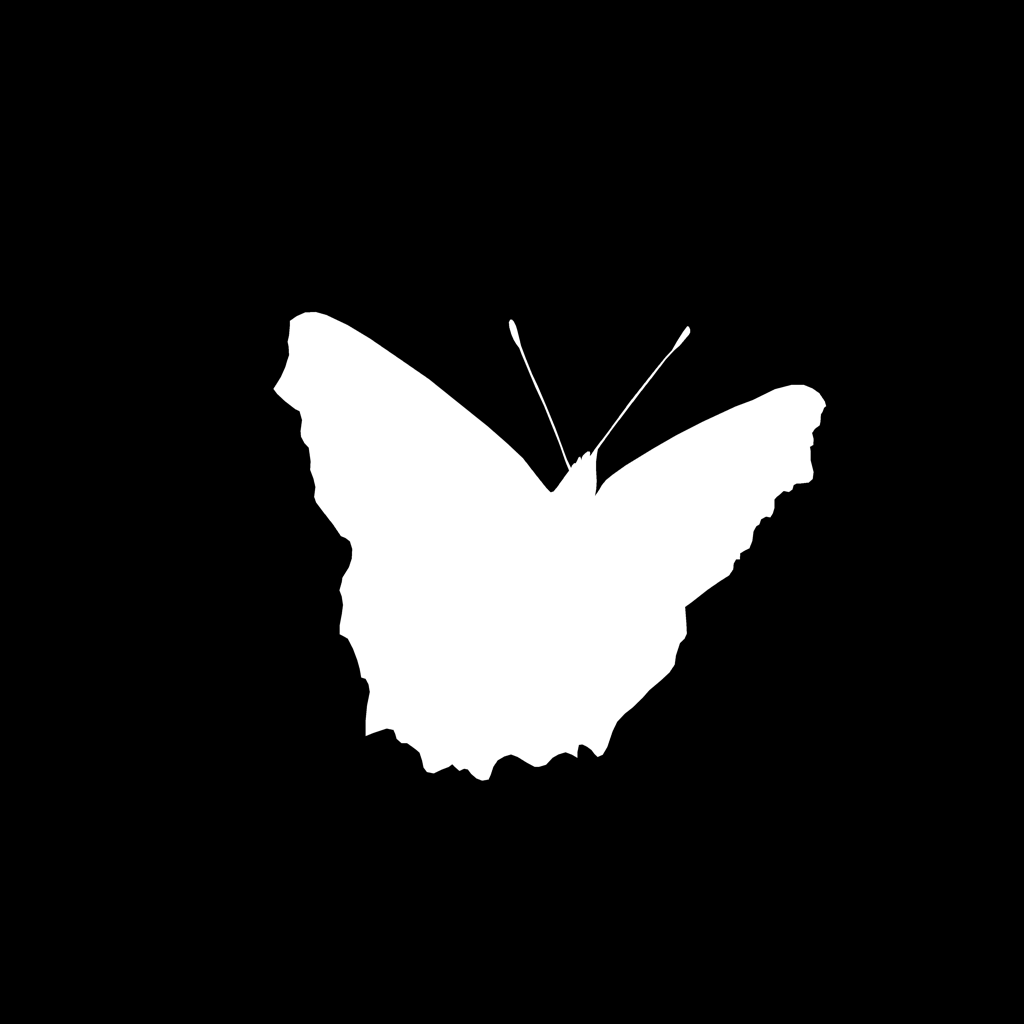}
            \\
            \includegraphics[width=0.080\textwidth,height=0.08\textheight]{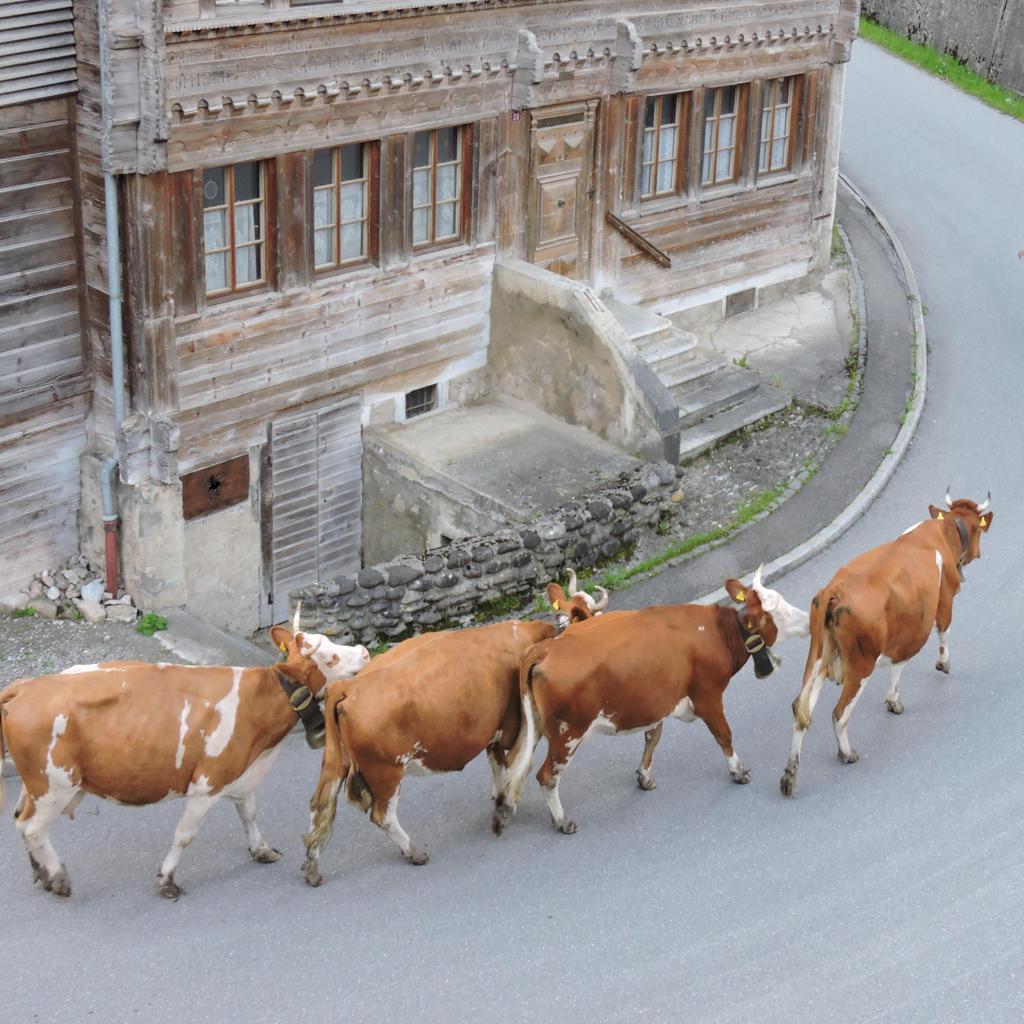}&
            \includegraphics[width=0.080\textwidth,height=0.08\textheight]{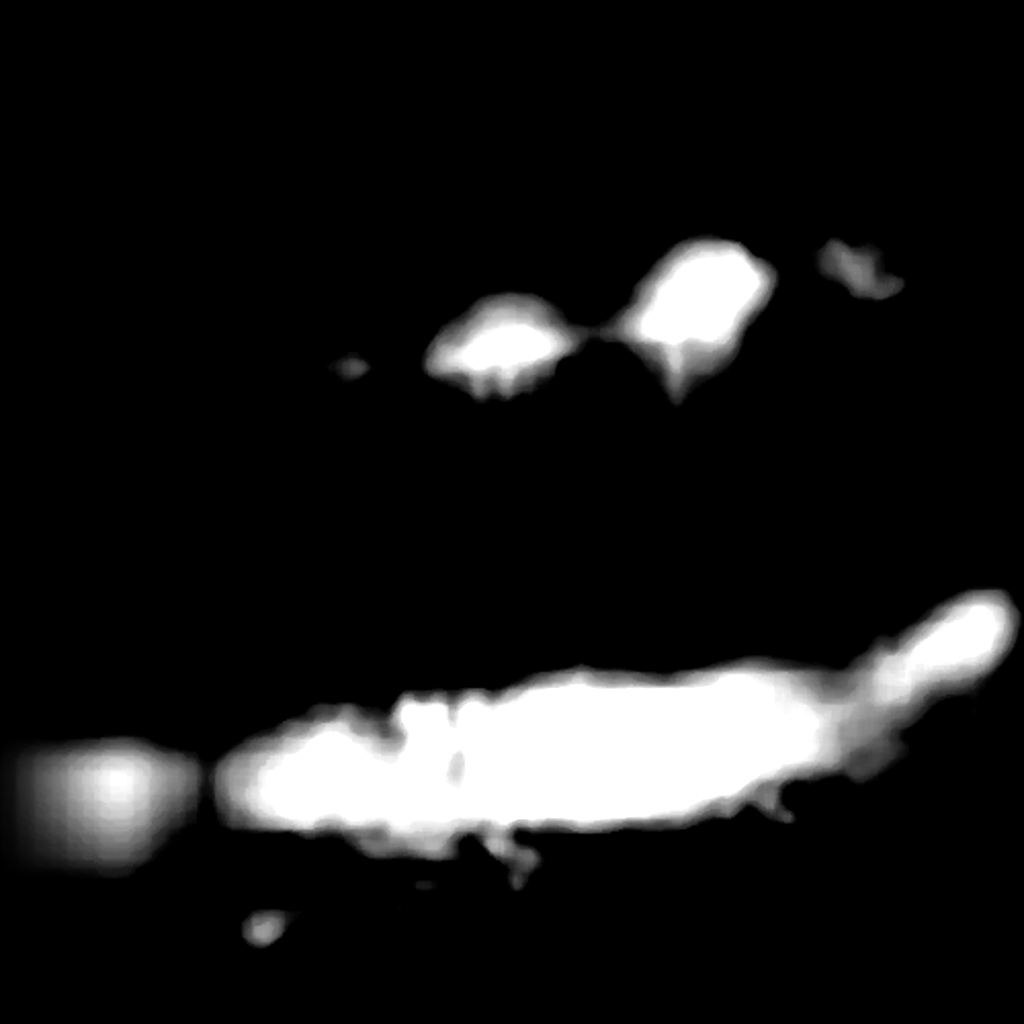}&
            \includegraphics[width=0.080\textwidth,height=0.08\textheight]{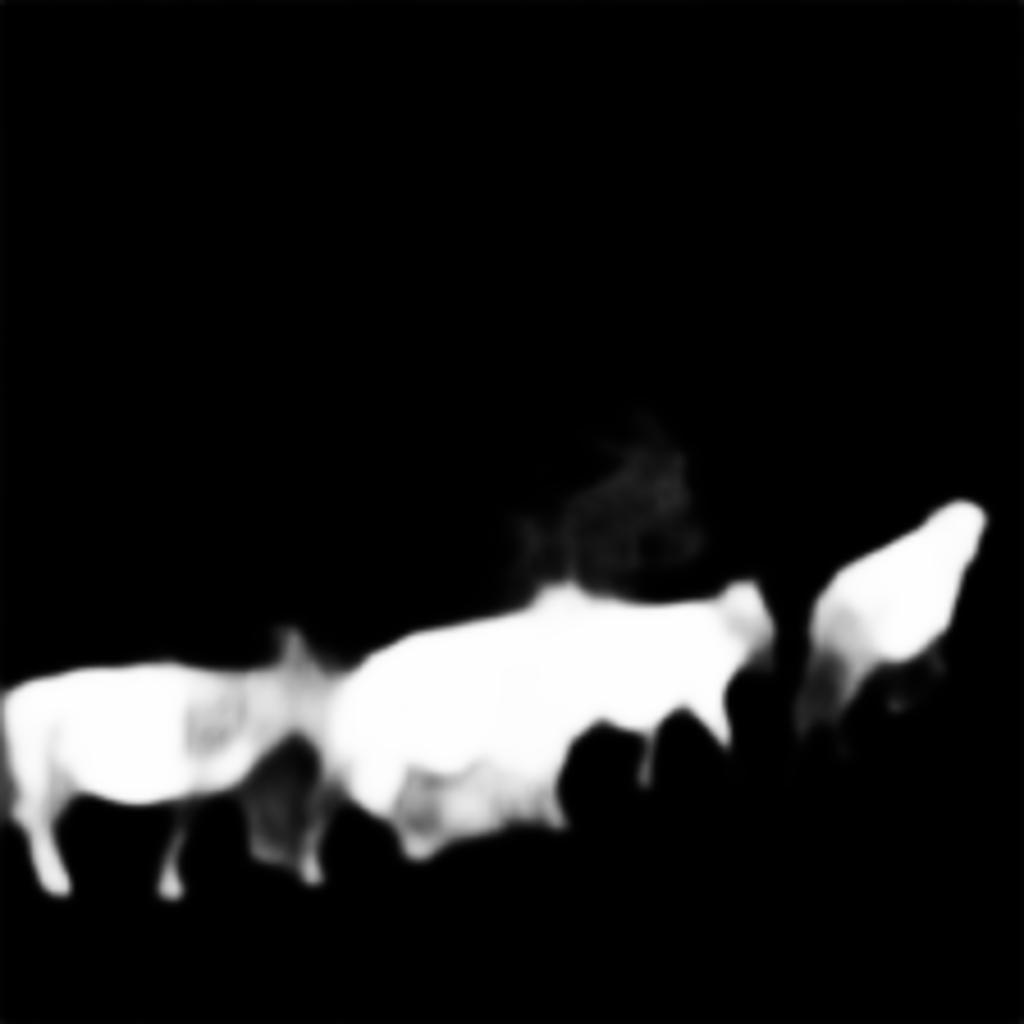}&
            \includegraphics[width=0.080\textwidth,height=0.08\textheight]{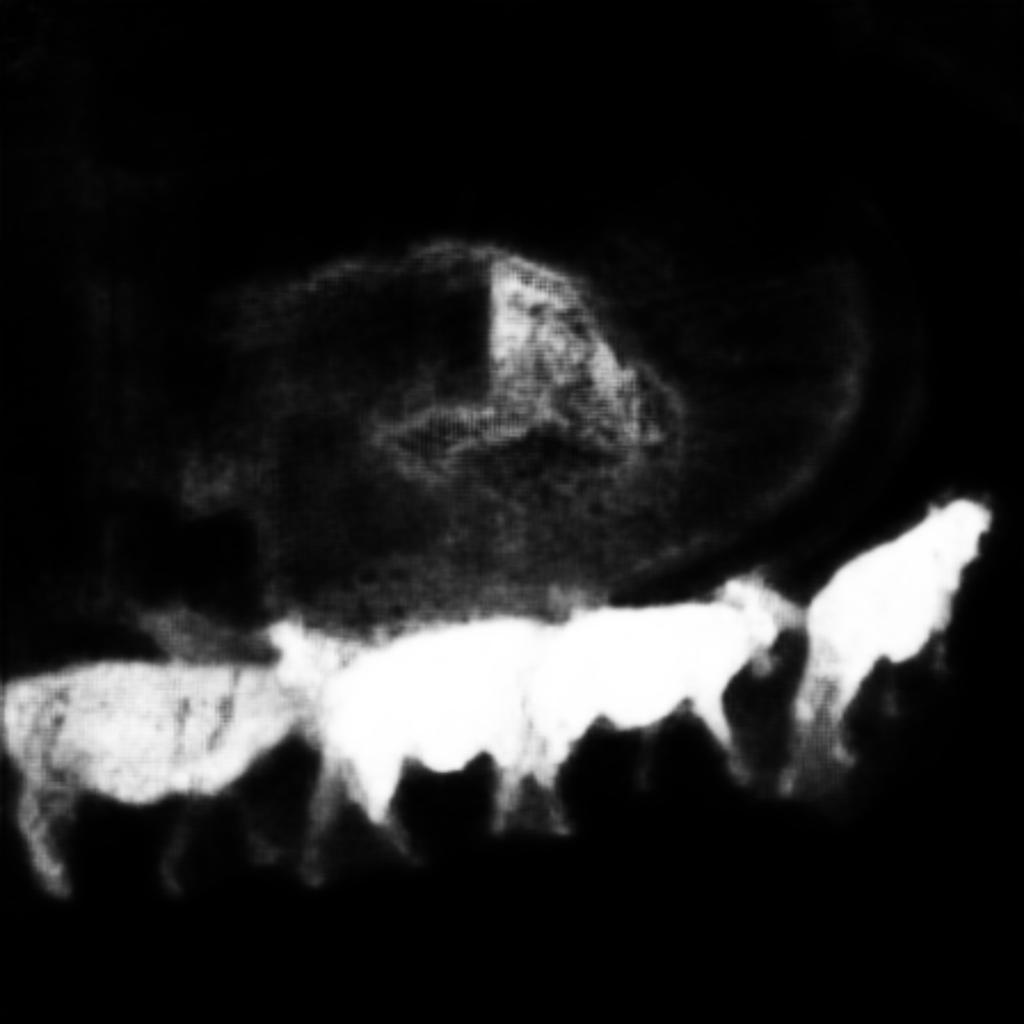}&
            \includegraphics[width=0.080\textwidth,height=0.08\textheight]{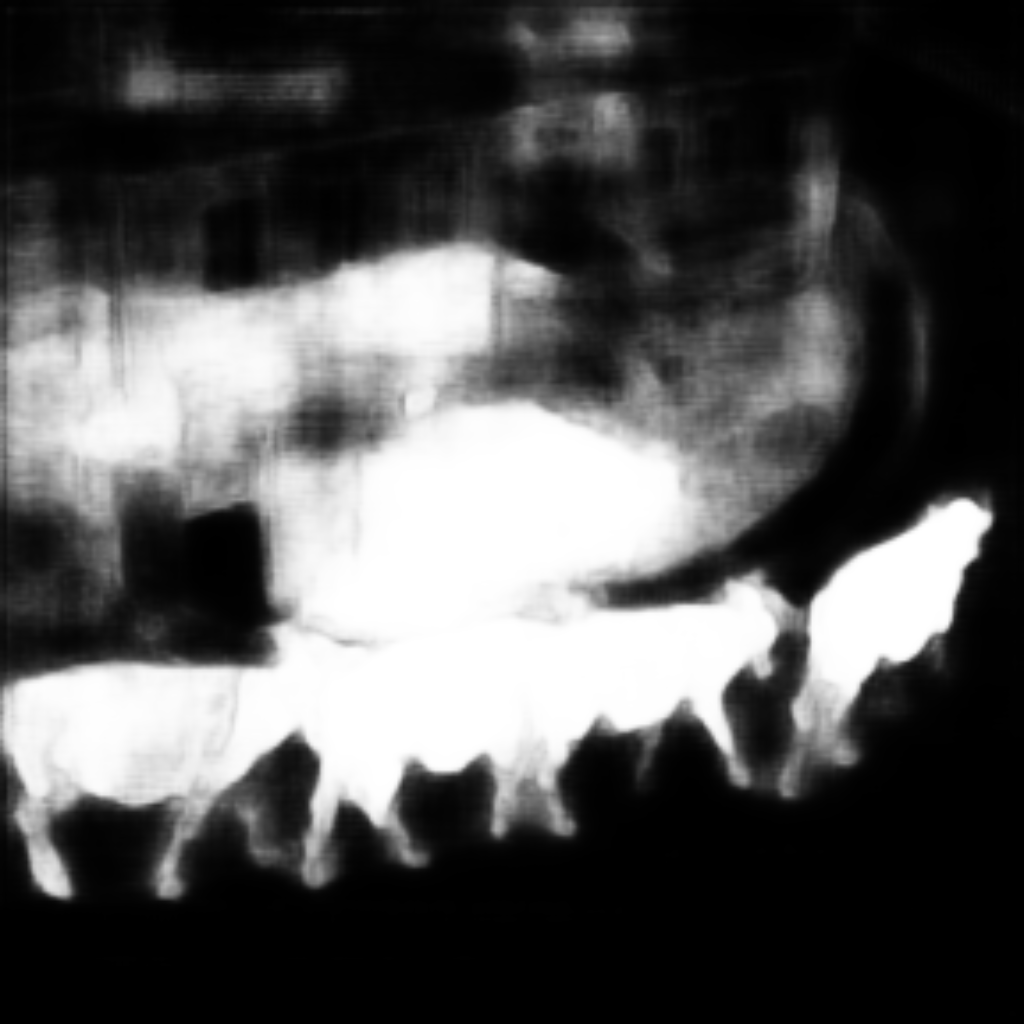}&
            \includegraphics[width=0.080\textwidth,height=0.08\textheight]{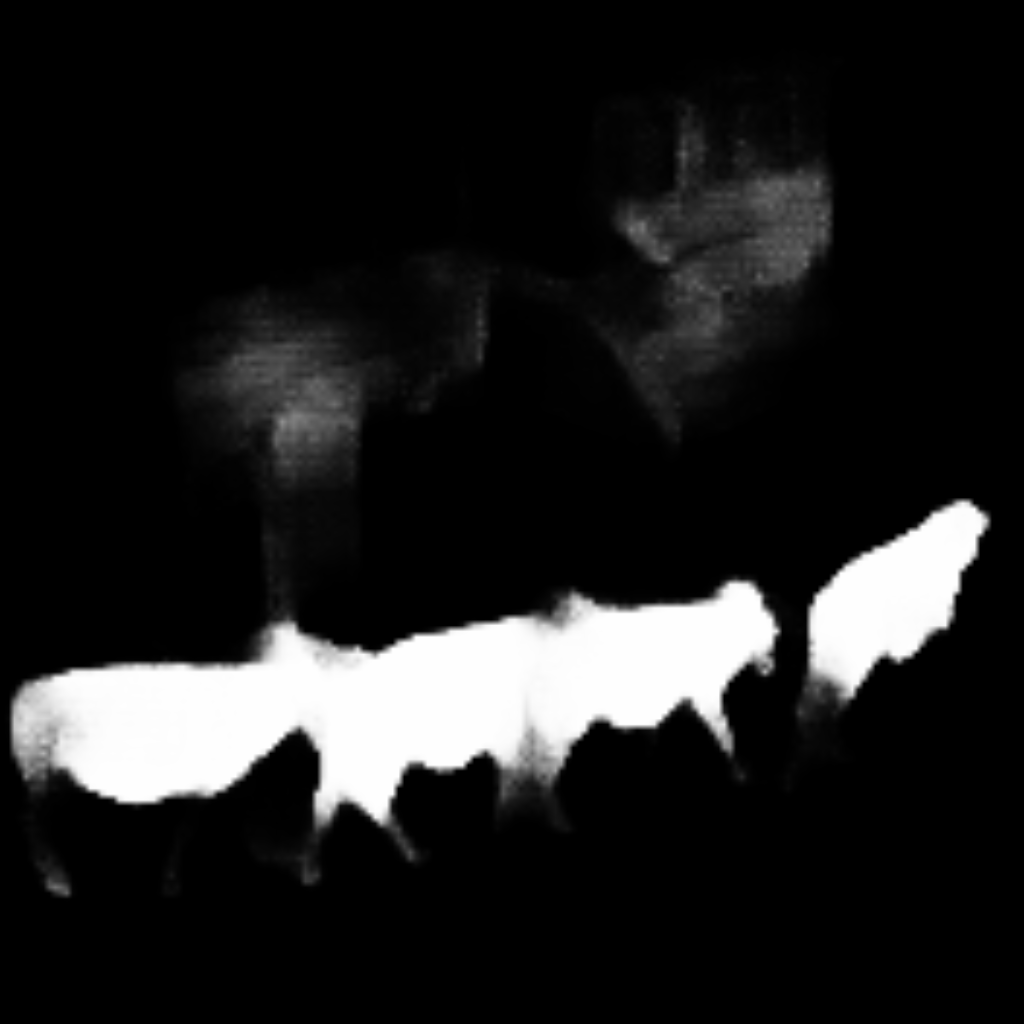}&
            \includegraphics[width=0.080\textwidth,height=0.08\textheight]{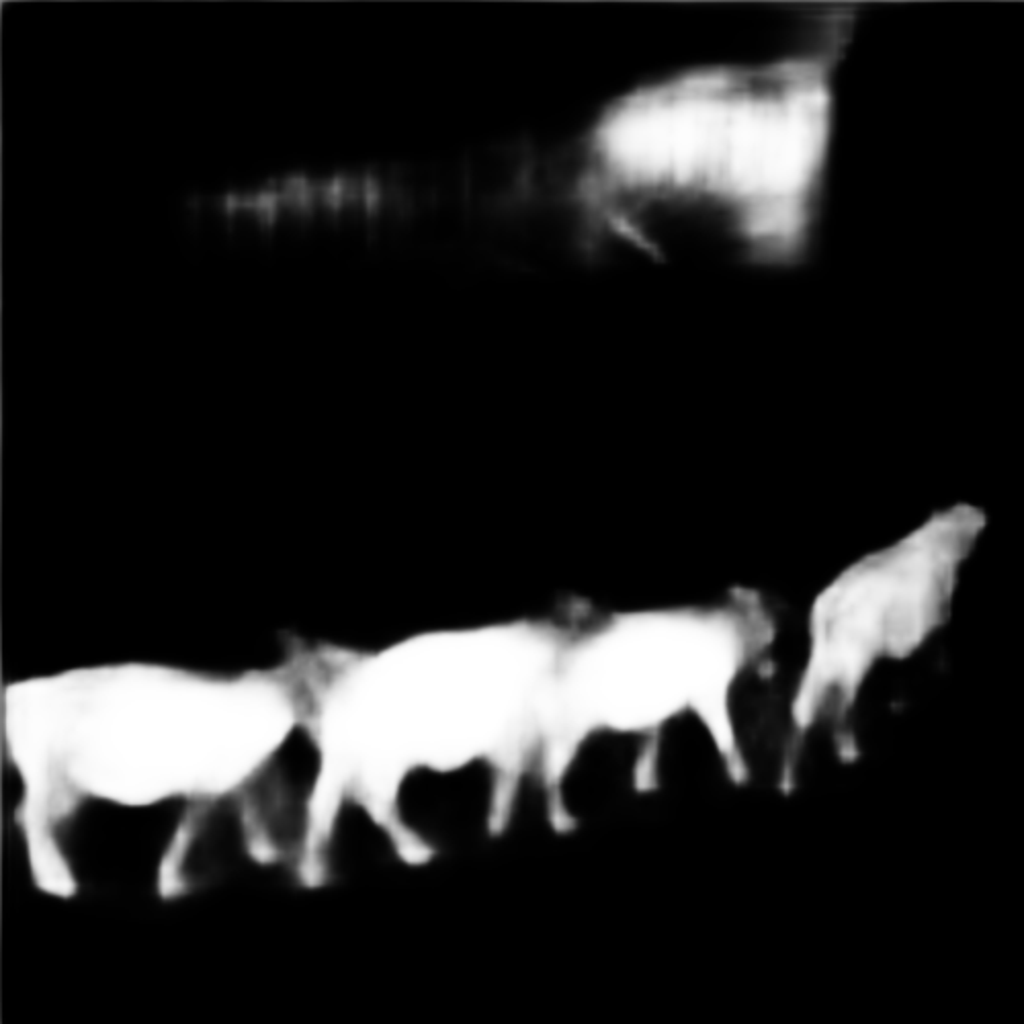}&
            \includegraphics[width=0.080\textwidth,height=0.08\textheight]{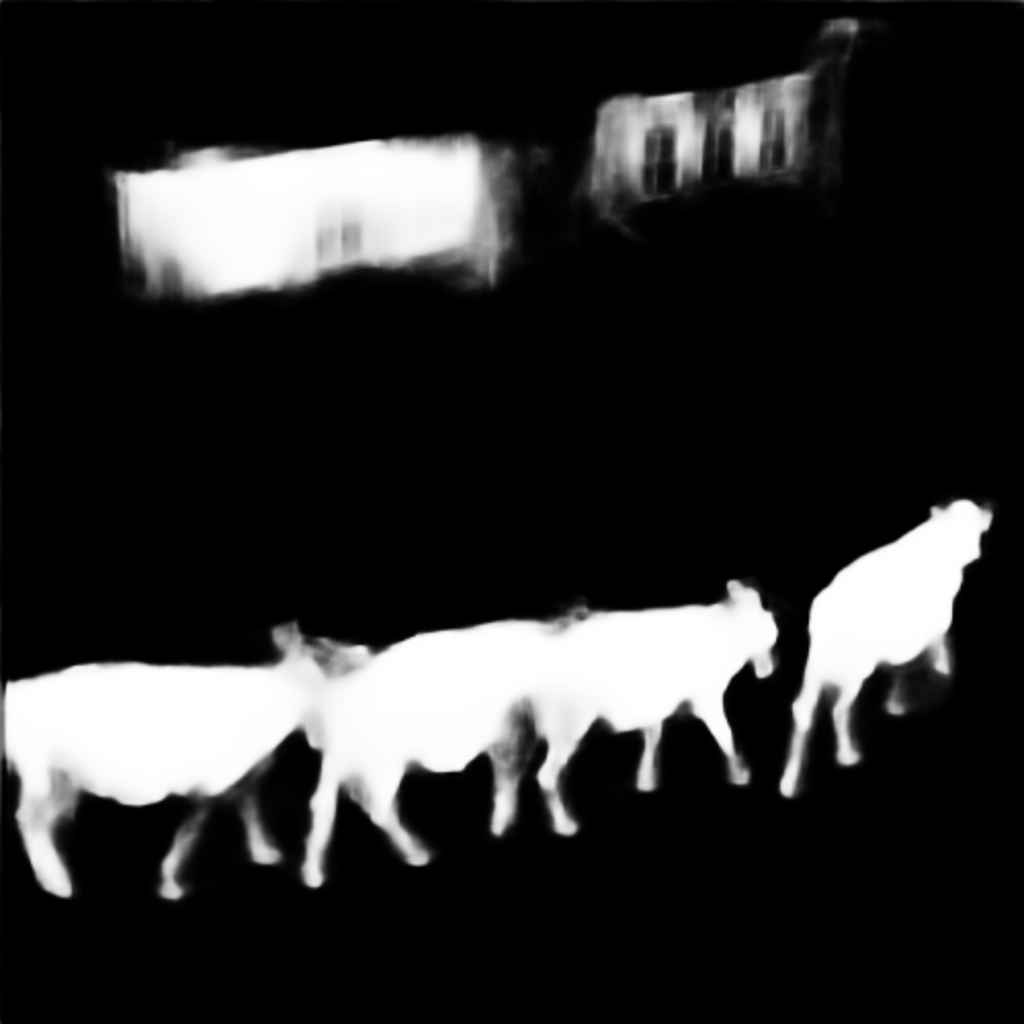}&
            \includegraphics[width=0.080\textwidth,height=0.08\textheight]{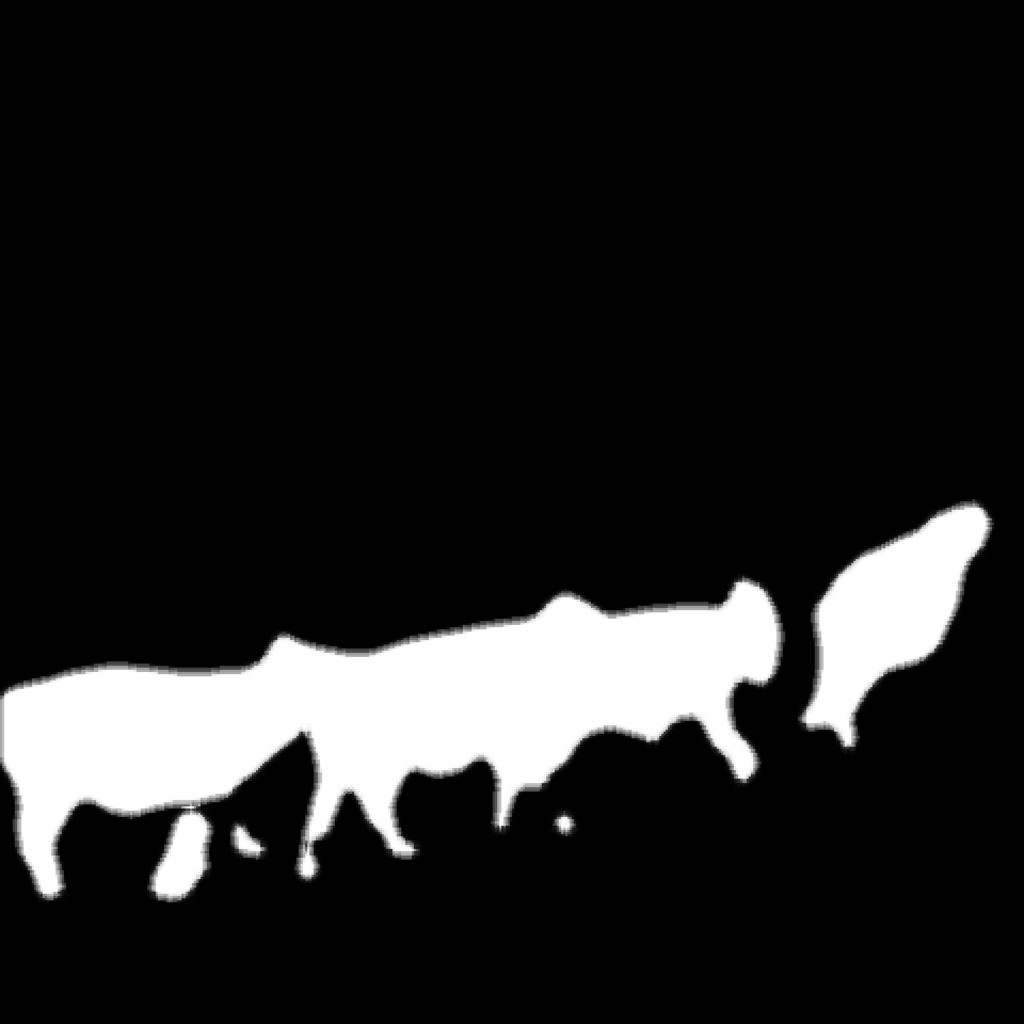}&
            \includegraphics[width=0.080\textwidth,height=0.08\textheight]{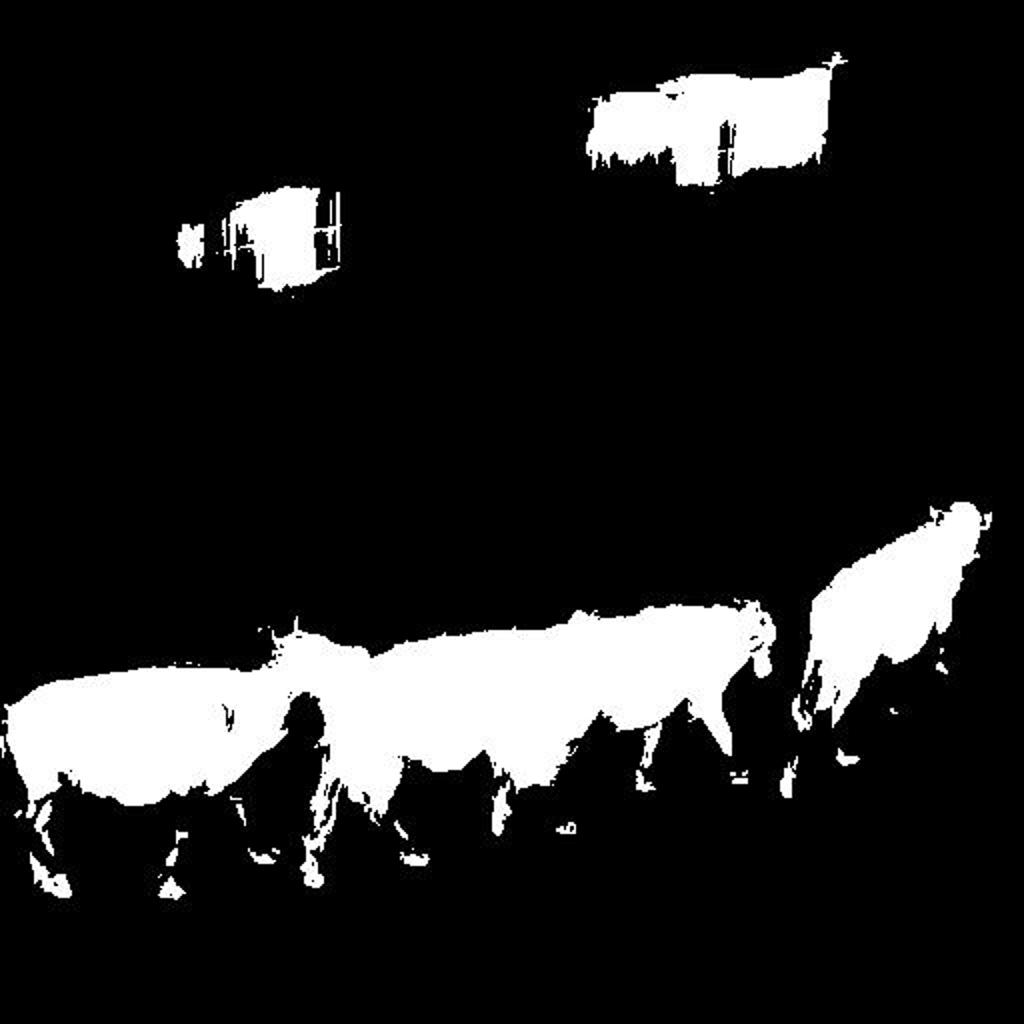}&
            \includegraphics[width=0.080\textwidth,height=0.08\textheight]{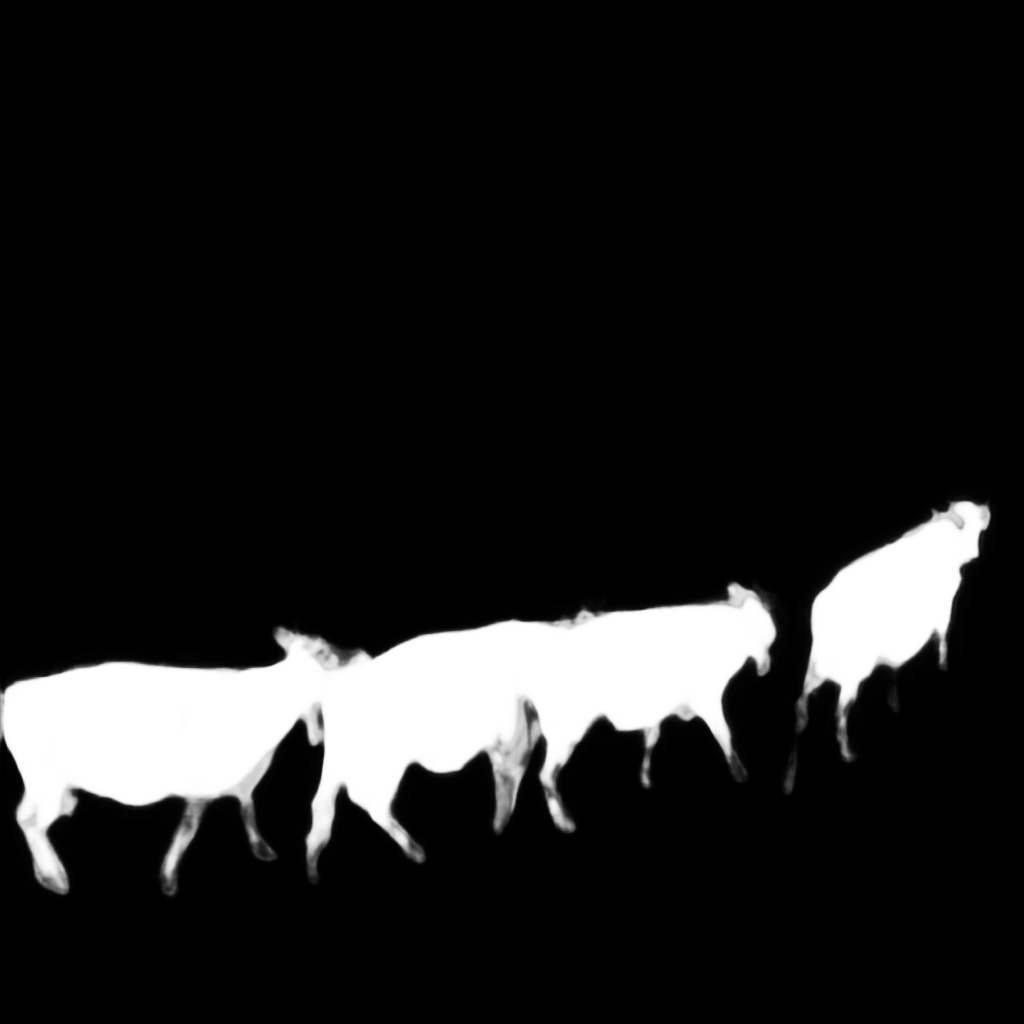}&
            \includegraphics[width=0.080\textwidth,height=0.08\textheight]{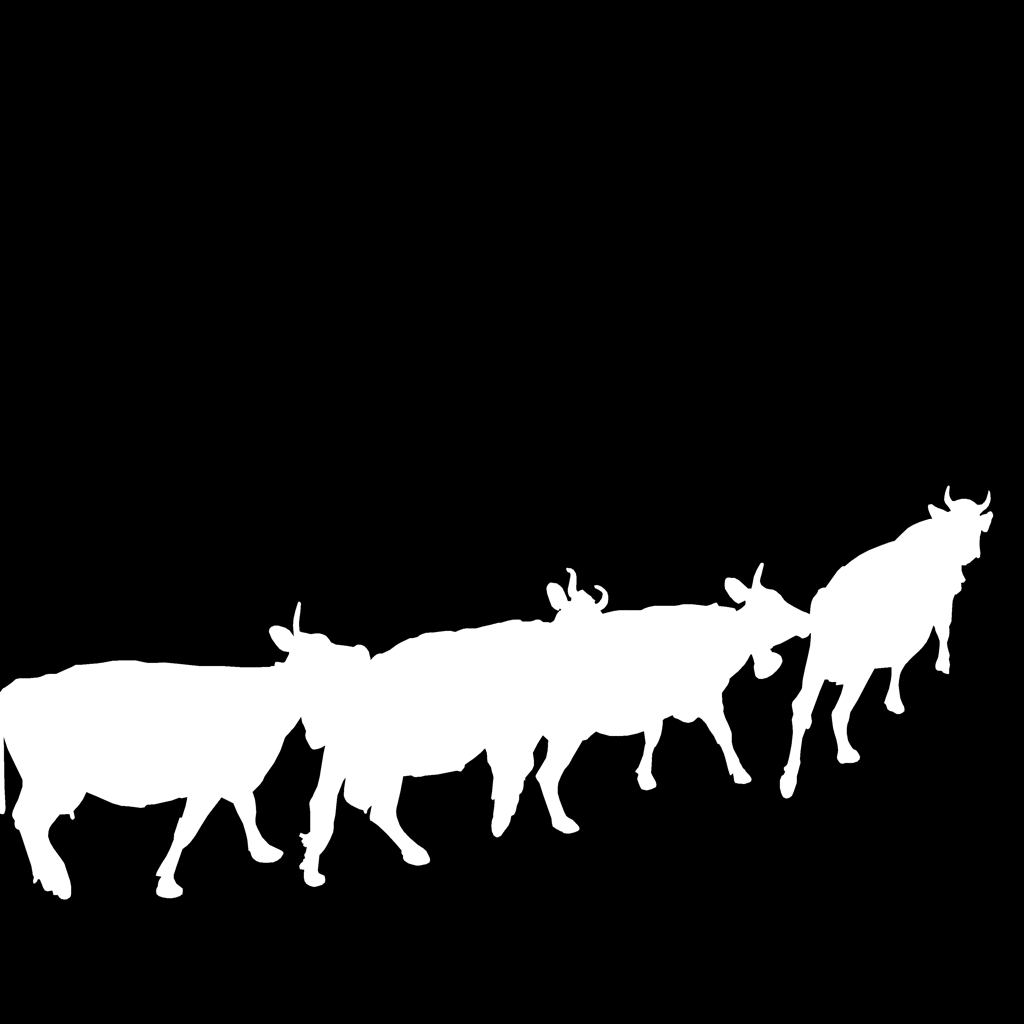}
            \\
            Image &RFCN &DHS &UCF &Amulet &NLDF &DSS &RAS &DGRL &DGF &Ours &GT\\
        \end{tabular}
        \vspace{0.8mm}
    \caption{Visual comparison. All images are from HRSOD-Test dataset. Best viewed by zooming in.}
    \vspace{-1mm}
    \label{fig:compare}
    \end{figure*}
    \begin{figure*}[!t]
    \centering
    \tabcolsep0.3mm \renewcommand{\arraystretch}{0.5}
        \begin{tabular}{ccccccc}
         \includegraphics[width=0.16\textwidth,height=0.04\textheight]{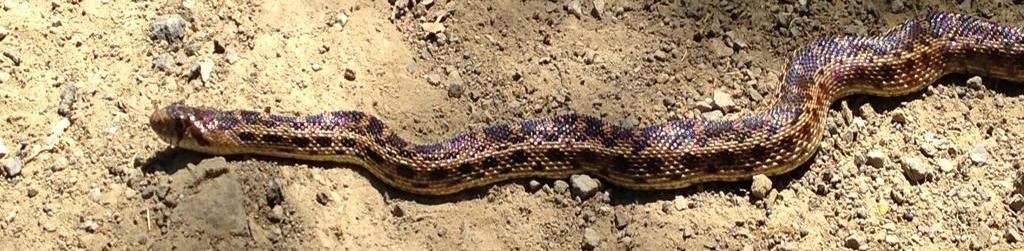}&
        \includegraphics[width=0.16\textwidth,height=0.04\textheight]{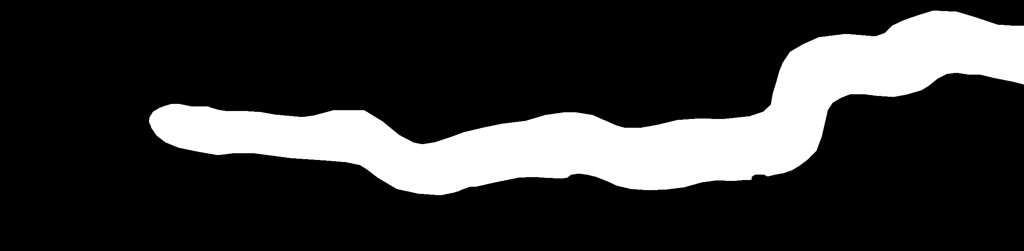}&
        \includegraphics[width=0.16\textwidth,height=0.04\textheight]{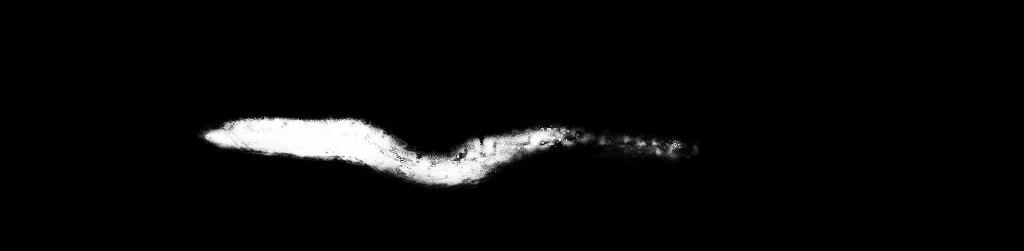}&  \includegraphics[width=0.16\textwidth,height=0.04\textheight]{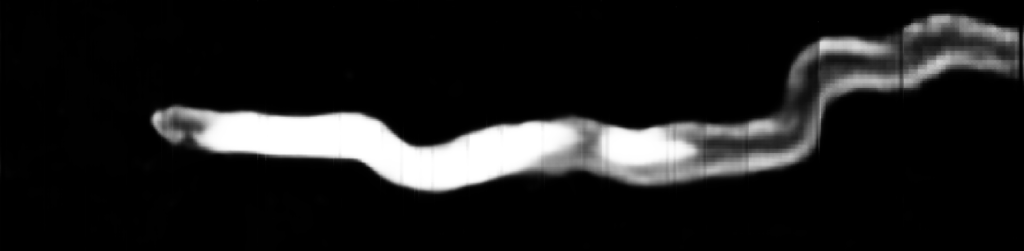}&   \includegraphics[width=0.16\textwidth,height=0.04\textheight]{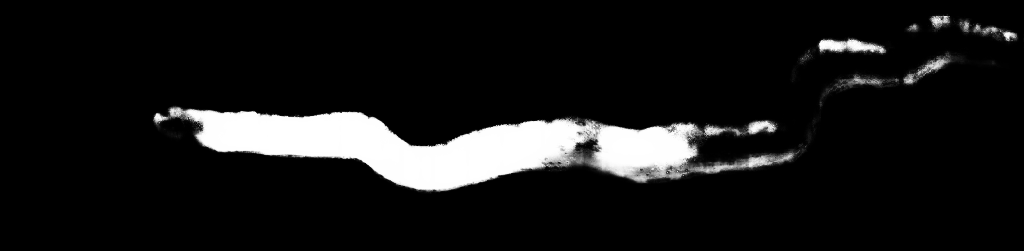}&
        \includegraphics[width=0.16\textwidth,height=0.04\textheight]{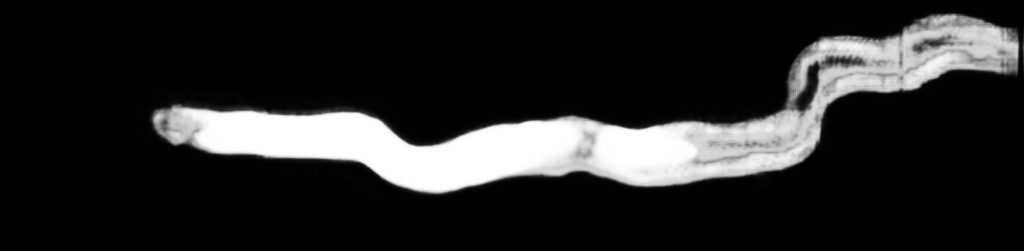}\\
        \includegraphics[width=0.16\textwidth,height=0.07\textheight]{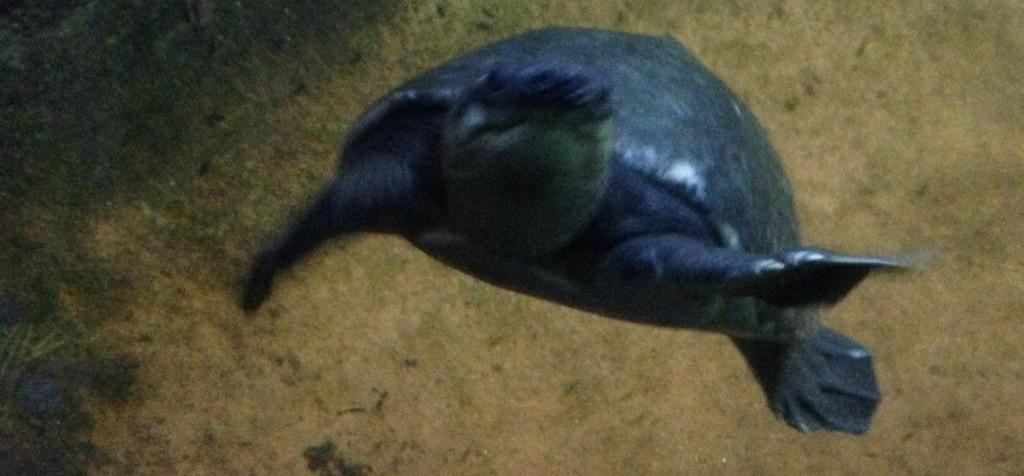}&
        \includegraphics[width=0.16\textwidth,height=0.07\textheight]{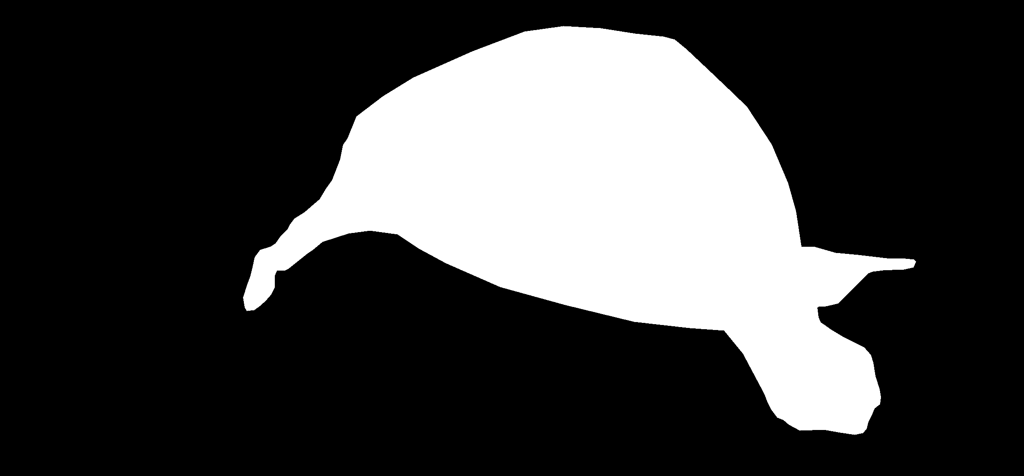}&
        \includegraphics[width=0.16\textwidth,height=0.07\textheight]{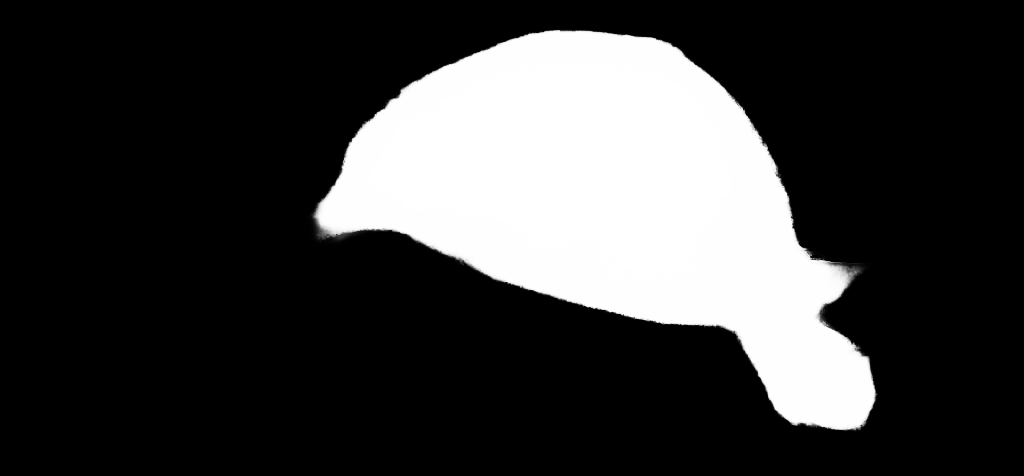}&  \includegraphics[width=0.16\textwidth,height=0.07\textheight]{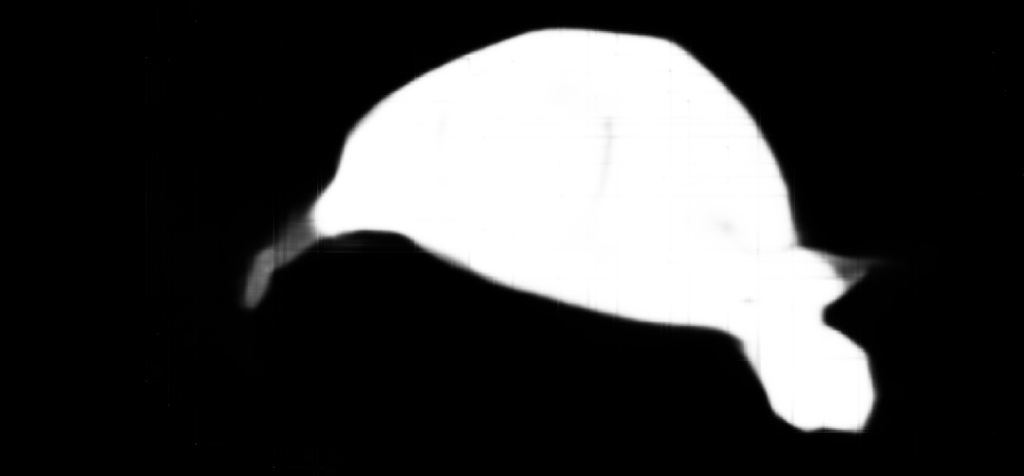}&   \includegraphics[width=0.16\textwidth,height=0.07\textheight]{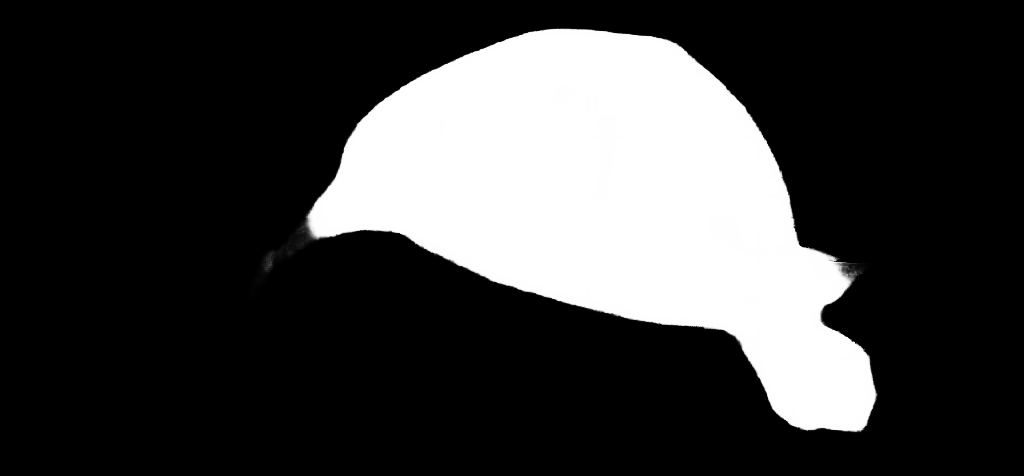}&
        \includegraphics[width=0.16\textwidth,height=0.07\textheight]{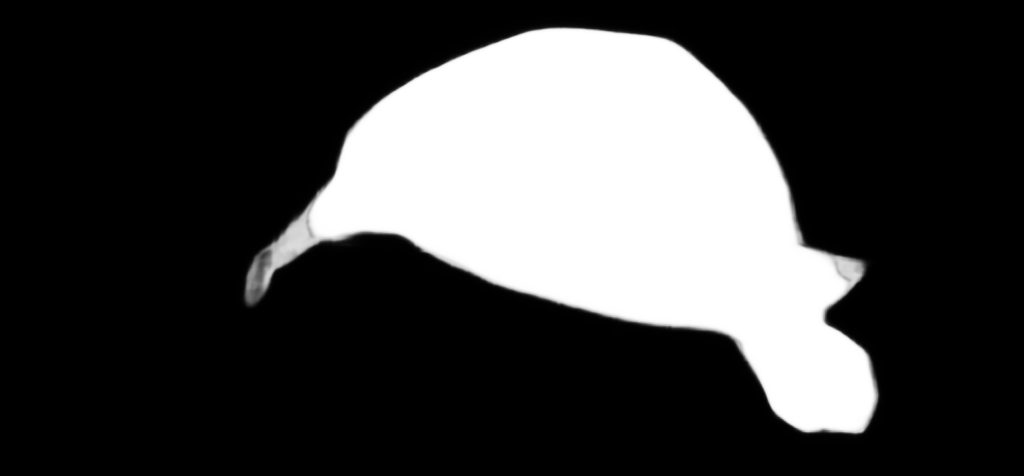}\\
\footnotesize Image &\footnotesize GT &\footnotesize GSN+CRF &\footnotesize GSN+APS+LRN &\footnotesize GSN+APS+LRN+CRF&\footnotesize GSN+APS+LRN+GLFN \\
        \end{tabular}
        \vspace{0.8mm}
    \caption{Visual comparison of our method with variations using Dense CRF~\cite{krahenbuhl2011efficient}.}
    \vspace{-6mm}
    \label{fig:compare}
    \end{figure*}
    \begin{table*}[htp]
\setlength{\tabcolsep}{5pt}
\centering
%\small
%\small
\renewcommand{\arraystretch}{1}
\begin{tabular}{|c|c|c|c|c|c|c|c|c|c|c|}
\hline
Dataset& Ours&NLDF&UCF&DHS&DSS&Amulet&RAS&DGRL&DGF&RFCN\\
\hline
HRSOD-Test&\textbf{17.57}&22.34&22.84&22.85&25.53&25.75&26.26&30.10&32.91&68.98\\
\hline
DAVIS-S&\textbf{8.18}&23.56&15.69&18.34&19.35&21.11&18.49&14.48&19.77&21.00\\
\hline
\end{tabular}
%\end{center}
\vspace{1.1mm}
\caption{The Boundary Displacement Error (smaller is better) of the state-of-the-art methods on high-resolution datasets. The best results are shown in bold.}\label{tab:bde}
%\vspace{-1.1mm}
\end{table*}

\begin{table*}[htp]
\begin{center}
%\tiny
%\caption{Run time analysis of the compared methods.}
\label{tab:runtime}
\vspace{-2mm}
\begin{tabular}{|c|c|c|c|c|c|c|c|c|c|c|c|c|}
\hline
%       &{\wuhao{Ours}}&{\sihao{KSR}
%}&{\sihao{ELD}}&{\sihao{DHS}}&{\sihao{MCDL}}&{\sihao{LEGS}}&{\sihao{DRFI}}&{\sihao{DCL}}&{\sihao{BL}}&{\sihao{RFCN}}&{\sihao{DS}}&{\sihao{MDF}} \\
       &Ours*&Ours&DGF&DGRL&RAS&DSS&NLDF&Amulet&UCF&DHS&RFCN \\
\cline{2-9}
\hline
Time (s)&0.39&0.05&0.41&0.52&0.08&5.12&2.31&0.05&0.14&0.05&4.54\\
\hline
Model Size(MB) &129.6&129.6&248.9&648.0&81&447.3&425.9&132.6&117.9&376.2&1126.4\\

\hline
\end{tabular}
\vspace{-2mm}
\end{center}
\caption{Running time and model size of the state-of-the-art methods.}\label{tab:time}
\vspace{-3mm}
\end{table*}
%\vspace{20mm}
\subsubsection{Implementation Details}

All experiments are conducted on a PC with an i7-8700 CPU and a 1080 Ti GPU, with the Caffe toolbox~\cite{jia2014caffe}.
In our method, every stage is trained to minimize a pixelwise softmax loss function, by using the stochastic gradient descent (SGD). Empirically, the momentum parameter is set to 0.9 and the weight decay is set to 0.0005.
For GSN and LRN, the inputs are first warped into $384\times384$ and the batch size is set to 32. The weights in block 1 to block 5 are initialized with the pre-trained VGG model~\cite{simonyan2014very}, while weight parameters of newly-added convolutional layers are randomly initialized by using the ``msra'' method~\cite{he2015delving}. The learning rates of the pre-trained and newly-added layers are set to 1e-3 and 1e-2, respectively. GLFN is trained from scratch, and its weight parameters of convolutional layers are also randomly initialized by using the ``msra'' method. Its inputs are warped into $1024\times1024$ and the batch size is set to 2. Source code will be released.

\subsection{Comparison with the State-of-the-arts}
We compare our algorithm with 9 state-of-the-art methods, including
RFCN~\cite{wang2016saliency}, DHS~\cite{liu2016dhsnet}, UCF~\cite{zhang2017learning}, Amulet~\cite{zhang2017amulet}, NLDF~\cite{luo2017non}, DSS~\cite{hou2017deeply}, RAS~\cite{chen2018reverse}, DGF~\cite{wu2018fast} and DGRL~\cite{wang2018detect}. For a fair comparison, we use either the implementations with recommended parameter settings
or the saliency maps provided by the authors. To demonstrate the effectiveness of our approach, we provide two versions of our results. Ours-D represents for training on DUTS while Ours-DH represents for training on DUTS and HRSOD.

One thing deserves to be mentioned is that in our framework, GSN and LRN can be any saliency detection model.
We just choose simple FCNs to validate the effectiveness of our framework. With our method, even simple FCNs can outperform other complicated models.

\textbf{Quantitative Evaluation.}
$F_\beta$ measure, S-measure and MAE scores are given in Table~\ref{tab-state}. As can be seen, our method outperforms all the existing state-of-the-art methods on our new-built high-resolution datasets with a large margin. It also achieves comparable or even superior performance than them on some widely used saliency detection datasets. We provide the PR curves in the supplementary material due to limited space.

\textbf{Qualitative Evaluation.} Figure~\ref{fig:compare} shows a visual comparison of our method with respect to others. It can be seen that our method is capable of accurately detecting salient objects as well as suppressing the background clutter. Further, our saliency maps have better boundary shape and are much closer to the ground truth maps in challenging cases.
\subsection{ Ablation Analysis and Discussion}
\subsubsection{Ablation Analysis}
In this section, we provide the results about different variants of our method to further verify our main contributions.

\textbf{LRN, GLFN vs CRF.} In our method, LRN learns to refine uncertain regions under the guidance of GSN. To demonstrate its effectiveness, We also compare it with CRF \cite{krahenbuhl2011efficient}, a widely used post-processing for saliency detection. The parameters are set as in \cite{hou2017deeply}. We employ the CRF to refine predictions of GSN and LRN, denoted as GSN+CRF and GSN+APS+LRN+CRF, respectively. The results in Table~\ref{tab:cmw_asm} show that our method outperforms CRF by a large margin.
Figure \ref{fig:compare} shows the qualitative results. We find that our LRN and GLFN progressively improve details
of saliency maps while the CRF fails to recover lost details.
%\vspace{-1mm}
\begin{table}[htp]
\setlength{\tabcolsep}{5pt}
\centering
%\small
%\small
\renewcommand{\arraystretch}{1}
\begin{tabular}{|c|c|c|c|}
\hline
Network Structure&$F_\beta$&S-m& MAE\\
\hline
GSN&0.842&0.866&0.047\\
\hline
GSN+CRF&0.858&0.852&0.038\\
\hline
GSN+RPS+LRN&0.860&0.871&0.037\\
\hline
GSN+APS+LRN&0.877&0.883&0.036\\
\hline
GSN+APS+LRN+CRF&0.880&0.875&0.033\\
\hline
GSN+APS+LRN+GLFN&0.888&0.897&0.030\\
\hline
\end{tabular}
\vspace{0.5mm}
\caption{Comparison of the different variants on HRSOD-Test.}\label{tab:cmw_asm}
\vspace{-3mm}
\end{table}

\textbf{APS vs RPS.} To demonstrate the effectiveness of the proposed APS scheme, we train LRN on patches which are randomly sampled. For fair comparison, we set the number and size of sampled patches to be the same with our proposed APS.
We denote this setting as GSN+RPS+LRN. Various metrics in Table~\ref{tab:cmw_asm}  demonstrate that APS significantly outperforms RPS, which indicates the important role of our proposed APS.

\begin{figure}[htp]
    \centering
    \hspace{-2mm}
        \begin{tabular}{c@{\hspace{0.5mm}} c@{\hspace{0.5mm}}}
            \hspace{-3mm}\includegraphics[width=0.23\textwidth,height=0.15\textheight]{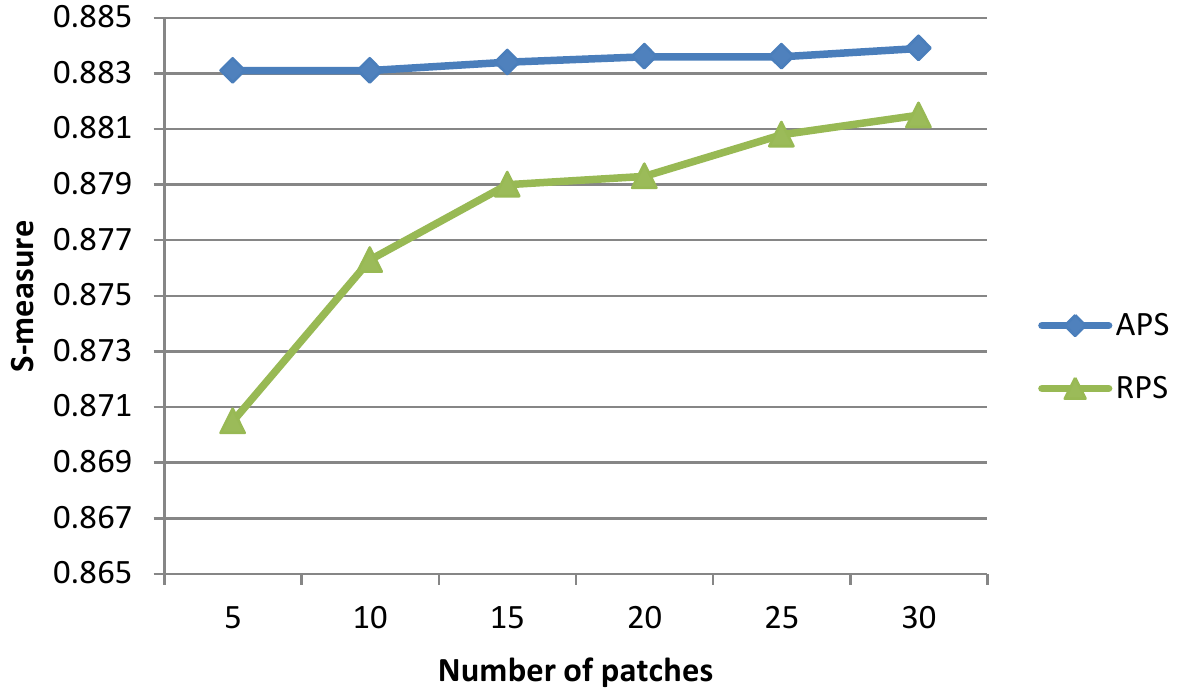}&
            \includegraphics[width=0.23\textwidth,height=0.15\textheight]{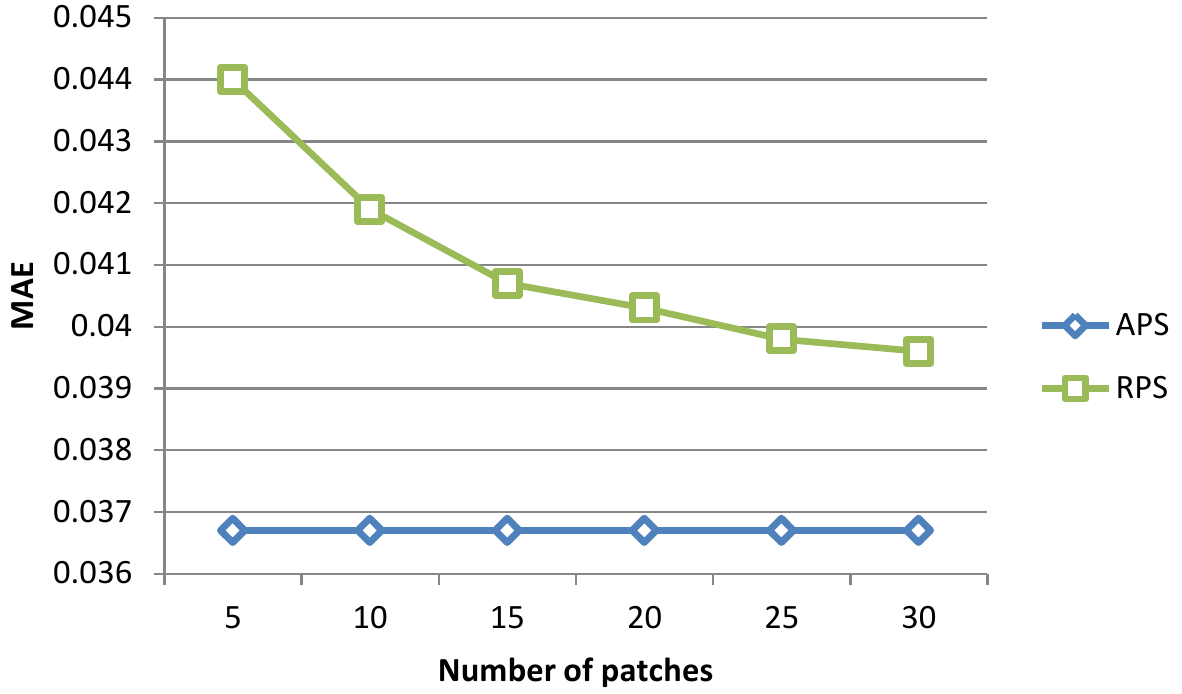}\\
            (a)&(b)\\
        \end{tabular}
    \caption{Refinement quality versus patch of numbers for
different approaches. (a) S-measure vs. patch of numbers. (b) MAE versus patch of numbers. Results are measured on the outputs of LRN.}
    \label{fig:ablation}
    \vspace{-3mm}
\end{figure}

\textbf{Performance vs number of patches.} For traditional patch-based methods, refining more patches brings more performance gain, but results in more computational cost. It seems like a tricky trade-off problem. Our proposed APS can ingeniously relieve this problem thanks to its focusing on  uncertain regions. Figure \ref{fig:ablation} shows that our APS is less sensitive to number of patches.
\vspace{-3mm}
\vspace{-4mm}
\subsubsection{More Discussion}
\textbf{Running time and model size.} Table \ref{tab:time} shows a comparison of running time and model size. Since other methods can not directly handle high-resolution images, the running time analysis of the compared methods is conducted with the same input size ($384\times384$) for fair. Also, we provide our running time for $1024\times1024$ inputs, denoted as Ours*. As it can be seen, our method is the fasted among all the compared methods and is quite efficient when directly handling high-resolution images.

\textbf{Boundary quality.} To further evaluate the precision of boundaries, we compare different methods by the Boundary Displacement Error (BDE) metric~\cite{freixenet2002yet}. This metric measures the average displacement error of boundary pixels between two predictions, which can be formulated as: %Particularly, we compute the error of a boundary pixel in a saliency map by the distance between the pixel and the closest pixel in the corresponding ground truth mask. The original BDE
\begin{equation*}
\begin{split}
 BDE(\bm{X},\bm{Y})=&\frac{1}{2}\bigg[\frac{1}{N_X}\sum_{x}inf_{y\in \bm{Y}}d(x,y)\\
 &+\frac{1}{N_Y}\sum_{y} inf_{x\in \bm{X}}d(x,y)\bigg]
\end{split}
\end{equation*}
where $\bm{X}$ and $\bm{Y}$ are two boundary pixel sets, and $x$, $y$ are pixels in them, respectively. $N_X$ and $N_Y$ denote the number of pixels in $\bm{X}$ and $\bm{Y}$. $inf$ represents for the infimum and $d(\cdot)$ denotes Euclidean distance. %Considering that BDE is sensitive to the threshold for binarizing salieny maps, we slide the threshold from 0 to 255 and take the minimum BDE. %For this type of evaluation, it is more reasonable to compute BDE on high-resolution datasets. %(\emph{i.e.}, HRSOD-test and DAVIS ).
We only compute the BDE on high-resolution datasets because other benchmarks are not qualified enough on boundaries in pixel-level due to relatively poor annotations. %suitable for evaluating the quality of saliency prediction in a global view ( \emph{e.g.}, F-measure ), but
The BDE for the state-of-the-art methods on HRSOD-Test and DAVIS-S are listed in Table \ref{tab:bde}. The results indicate that our predictions have better boundary shape and are closer to the ground truth maps.
%-------------------------------------------------------------------------
\section{Conclusion}
In this paper, we push forward high-resolution saliency detection task and provide a high-resolution saliency detection dataset (HRSOD) for facilitating studies in high-resolution saliency prediction. A novel approach is proposed to address this challenging task. It leverages both global semantic information and local high-resolution details to accurately detect salient objects in high-resolution images. %A GLFN is further proposed to enforce spatial consistency and boost performance. %Without increasing extra computational cost and memory burden, we succeed to optimize boundary details under global guidance.
Extensive evaluations on high-resolution datasets and popular benchmark datasets verify the effectiveness of our method. We will explore to develop weakly supervised high-resolution salient object detection in the future.
%-----------------------------------------------------------------------

\noindent\textbf{Acknowledgements.} This paper is supported in part by National Natural Science Foundation of China No. 61725202, 61829102, 61751212, in part by the Fundamental Research Funds for the Central Universities under Grant No. DUT19GJ201 and gifts from Adobe.

{\small
\bibliographystyle{ieee_fullname}
\bibliography{egbib}

\begin{thebibliography}{10}\itemsep=-1pt

\bibitem{achanta2009frequency}
Ravi Achanta, Sheila Hemami, Francisco Estrada, and Sabine Susstrunk.
\newblock Frequency-tuned salient region detection.
\newblock In {\em CVPR}, pages 1597--1604, 2009.

\bibitem{borji2015salient}
Ali Borji, Ming-Ming Cheng, Huaizu Jiang, and Jia Li.
\newblock Salient object detection: A benchmark.
\newblock {\em IEEE TIP}, 24(12):5706--5722, 2015.

\bibitem{chen2018reverse}
Shuhan Chen, Xiuli Tan, Ben Wang, and Xuelong Hu.
\newblock Reverse attention for salient object detection.
\newblock In {\em ECCV}, pages 234--250, 2018.

\bibitem{chen2019learning}
Xu Chen, Bryan~M Williams, Srinivasa~R Vallabhaneni, Gabriela Czanner, Rachel
  Williams, and Yalin Zheng.
\newblock Learning active contour models for medical image segmentation.
\newblock In {\em CVPR}, pages 11632--11640, 2019.

\bibitem{cheng2014salientshape}
Ming-Ming Cheng, Niloy~J Mitra, Xiaolei Huang, and Shi-Min Hu.
\newblock Salientshape: Group saliency in image collections.
\newblock {\em The Visual Computer}, 30(4):443--453, 2014.

\bibitem{cheng2015global}
Ming-Ming Cheng, Niloy~J Mitra, Xiaolei Huang, Philip~HS Torr, and Shi-Min Hu.
\newblock Global contrast based salient region detection.
\newblock {\em IEEE TPAMI}, 37(3):569--582, 2015.

\bibitem{das2017human}
Abhishek Das, Harsh Agrawal, Larry Zitnick, Devi Parikh, and Dhruv Batra.
\newblock Human attention in visual question answering: Do humans and deep
  networks look at the same regions?
\newblock {\em CVIU}, 163:90--100, 2017.

\bibitem{fan2018salient}
Deng-Ping Fan, Ming-Ming Cheng, Jiang-Jiang Liu, Shang-Hua Gao, Qibin Hou, and
  Ali Borji.
\newblock Salient objects in clutter: Bringing salient object detection to the
  foreground.
\newblock In {\em ECCV}, pages 186--202, 2018.

\bibitem{fan2017structure}
Deng-Ping Fan, Ming-Ming Cheng, Yun Liu, Tao Li, and Ali Borji.
\newblock {Structure-measure: A New Way to Evaluate Foreground Maps}.
\newblock In {\em ICCV}, pages 4548--4557, 2017.

\bibitem{fang2015captions}
Hao Fang, Saurabh Gupta, Forrest Iandola, Rupesh~K Srivastava, Li Deng, Piotr
  Doll{\'a}r, Jianfeng Gao, Xiaodong He, Margaret Mitchell, John~C Platt,
  et~al.
\newblock From captions to visual concepts and back.
\newblock In {\em CVPR}, pages 1473--1482, 2015.

\bibitem{freixenet2002yet}
Jordi Freixenet, Xavier Mu{\~n}oz, David Raba, Joan Mart{\'\i}, and Xavier
  Cuf{\'\i}.
\newblock Yet another survey on image segmentation: Region and boundary
  information integration.
\newblock In {\em ECCV}, pages 408--422, 2002.

\bibitem{he2015delving}
Kaiming He, Xiangyu Zhang, Shaoqing Ren, and Jian Sun.
\newblock Delving deep into rectifiers: Surpassing human-level performance on
  imagenet classification.
\newblock In {\em CVPR}, pages 1026--1034, 2015.

\bibitem{he2016deep}
Kaiming He, Xiangyu Zhang, Shaoqing Ren, and Jian Sun.
\newblock Deep residual learning for image recognition.
\newblock In {\em CVPR}, pages 770--778, 2016.

\bibitem{hou2017deeply}
Qibin Hou, Ming-Ming Cheng, Xiaowei Hu, Ali Borji, Zhuowen Tu, and Philip Torr.
\newblock Deeply supervised salient object detection with short connections.
\newblock In {\em CVPR}, pages 5300--5309, 2017.

\bibitem{huang2017densely}
Gao Huang, Zhuang Liu, Laurens Van Der~Maaten, and Kilian~Q Weinberger.
\newblock Densely connected convolutional networks.
\newblock In {\em CVPR}, pages 4700--4708, 2017.

\bibitem{itti1998model}
Laurent Itti, Christof Koch, and Ernst Niebur.
\newblock A model of saliency-based visual attention for rapid scene analysis.
\newblock {\em IEEE TPAMI}, 20(11):1254--1259, 1998.

\bibitem{jia2014caffe}
Yangqing Jia, Evan Shelhamer, Jeff Donahue, Sergey Karayev, Jonathan Long, Ross
  Girshick, Sergio Guadarrama, and Trevor Darrell.
\newblock Caffe: Convolutional architecture for fast feature embedding.
\newblock In {\em ACM Multimedia}, pages 675--678, 2014.

\bibitem{joon2017exploiting}
Seong Joon~Oh, Rodrigo Benenson, Anna Khoreva, Zeynep Akata, Mario Fritz, and
  Bernt Schiele.
\newblock Exploiting saliency for object segmentation from image level labels.
\newblock In {\em CVPR}, pages 4410--4419, 2017.

\bibitem{krahenbuhl2011efficient}
Philipp Kr{\"a}henb{\"u}hl and Vladlen Koltun.
\newblock Efficient inference in fully connected crfs with gaussian edge
  potentials.
\newblock In {\em NIPS}, pages 109--117, 2011.

\bibitem{lai2016saliency}
Baisheng Lai and Xiaojin Gong.
\newblock Saliency guided dictionary learning for weakly-supervised image
  parsing.
\newblock In {\em CVPR}, pages 3630--3639, 2016.

\bibitem{li2015visual}
Guanbin Li and Yizhou Yu.
\newblock Visual saliency based on multiscale deep features.
\newblock In {\em CVPR}, pages 5455--5463, 2015.

\bibitem{li2013saliency}
Xiaohui Li, Huchuan Lu, Lihe Zhang, Xiang Ruan, and Ming-Hsuan Yang.
\newblock Saliency detection via dense and sparse reconstruction.
\newblock In {\em ICCV}, pages 2976--2983, 2013.

\bibitem{lischinski2006interactive}
Dani Lischinski, Zeev Farbman, Matt Uyttendaele, and Richard Szeliski.
\newblock Interactive local adjustment of tonal values.
\newblock In {\em ACM TOG}, volume~25, pages 646--653, 2006.

\bibitem{liu2016dhsnet}
Nian Liu and Junwei Han.
\newblock Dhsnet: Deep hierarchical saliency network for salient object
  detection.
\newblock In {\em CVPR}, pages 678--686, 2016.

\bibitem{long2015fully}
Jonathan Long, Evan Shelhamer, and Trevor Darrell.
\newblock Fully convolutional networks for semantic segmentation.
\newblock In {\em CVPR}, pages 3431--3440, 2015.

\bibitem{luo2017non}
Zhiming Luo, Akshaya~Kumar Mishra, Andrew Achkar, Justin~A Eichel, Shaozi Li,
  and Pierre-Marc Jodoin.
\newblock Non-local deep features for salient object detection.
\newblock In {\em CVPR}, pages 6609--6617, 2017.

\bibitem{perazzi2016benchmark}
Federico Perazzi, Jordi Pont-Tuset, Brian McWilliams, Luc Van~Gool, Markus
  Gross, and Alexander Sorkine-Hornung.
\newblock A benchmark dataset and evaluation methodology for video object
  segmentation.
\newblock In {\em CVPR}, pages 724--732, 2016.

\bibitem{rother2004grabcut}
Carsten Rother, Vladimir Kolmogorov, and Andrew Blake.
\newblock Grabcut: Interactive foreground extraction using iterated graph cuts.
\newblock In {\em ACM TOG}, volume~23, pages 309--314, 2004.

\bibitem{shen2016automatic}
Xiaoyong Shen, Aaron Hertzmann, Jiaya Jia, Sylvain Paris, Brian Price, Eli
  Shechtman, and Ian Sachs.
\newblock Automatic portrait segmentation for image stylization.
\newblock In {\em Computer Graphics Forum}, volume~35, pages 93--102, 2016.

\bibitem{simonyan2014very}
Karen Simonyan and Andrew Zisserman.
\newblock Very deep convolutional networks for large-scale image recognition.
\newblock {\em ICLR}, 2015.

\bibitem{tsai2016sky}
Yi-Hsuan Tsai, Xiaohui Shen, Zhe Lin, Kalyan Sunkavalli, and Ming-Hsuan Yang.
\newblock Sky is not the limit: semantic-aware sky replacement.
\newblock {\em ACM TOG}, 35(4):149--162, 2016.

\bibitem{wang2015deep}
Lijun Wang, Huchuan Lu, Xiang Ruan, and Ming-Hsuan Yang.
\newblock Deep networks for saliency detection via local estimation and global
  search.
\newblock In {\em CVPR}, pages 3183--3192, 2015.

\bibitem{wang2017}
Lijun Wang, Huchuan Lu, Yifan Wang, Mengyang Feng, Dong Wang, Baocai Yin, and
  Xiang Ruan.
\newblock Learning to detect salient objects with image-level supervision.
\newblock In {\em CVPR}, pages 136--145, 2017.

\bibitem{wang2016saliency}
Linzhao Wang, Lijun Wang, Huchuan Lu, Pingping Zhang, and Xiang Ruan.
\newblock Saliency detection with recurrent fully convolutional networks.
\newblock In {\em ECCV}, pages 825--841, 2016.

\bibitem{wang2018salience}
Shizheng Wang, Wenjuan Liao, Phil Surman, Zhigang Tu, Yuanjin Zheng, and
  Junsong Yuan.
\newblock Salience guided depth calibration for perceptually optimized
  compressive light field 3d display.
\newblock In {\em CVPR}, pages 2031--2040, 2018.

\bibitem{wang2018detect}
Tiantian Wang, Lihe Zhang, Shuo Wang, Huchuan Lu, Gang Yang, Xiang Ruan, and
  Ali Borji.
\newblock Detect globally, refine locally: A novel approach to saliency
  detection.
\newblock In {\em CVPR}, pages 3127--3135, 2018.

\bibitem{wei2012geodesic}
Yichen Wei, Fang Wen, Wangjiang Zhu, and Jian Sun.
\newblock Geodesic saliency using background priors.
\newblock In {\em ECCV}, pages 29--42, 2012.

\bibitem{wu2018fast}
Huikai Wu, Shuai Zheng, Junge Zhang, and Kaiqi Huang.
\newblock Fast end-to-end trainable guided filter.
\newblock In {\em CVPR}, pages 1838--1847, 2018.

\bibitem{xiao2019auto}
Yunxuan Xiao, Yikai Li, Yuwei Wu, and Lizhen Zhu.
\newblock Auto-retoucher (art)-a framework for background replacement and image
  editing.
\newblock {\em arXiv preprint arXiv:1901.03954}, 2019.

\bibitem{xu2015show}
Kelvin Xu, Jimmy Ba, Ryan Kiros, Kyunghyun Cho, Aaron Courville, Ruslan
  Salakhudinov, Rich Zemel, and Yoshua Bengio.
\newblock Show, attend and tell: Neural image caption generation with visual
  attention.
\newblock In {\em ICML}, pages 2048--2057, 2015.

\bibitem{yan2013hierarchical}
Qiong Yan, Li Xu, Jianping Shi, and Jiaya Jia.
\newblock Hierarchical saliency detection.
\newblock In {\em CVPR}, pages 1155--1162, 2013.

\bibitem{yang2013saliency}
Chuan Yang, Lihe Zhang, Huchuan Lu, Xiang Ruan, and Ming-Hsuan Yang.
\newblock Saliency detection via graph-based manifold ranking.
\newblock In {\em CVPR}, pages 3166--3173, 2013.

\bibitem{yu2015multi}
Fisher Yu and Vladlen Koltun.
\newblock Multi-scale context aggregation by dilated convolutions.
\newblock {\em ICLR}, 2015.

\bibitem{zhang2015saliency}
Fan Zhang, Bo Du, and Liangpei Zhang.
\newblock Saliency-guided unsupervised feature learning for scene
  classification.
\newblock {\em IEEE TGRS}, 53(4):2175--2184, 2015.

\bibitem{zhang2018bi}
Lu Zhang, Ju Dai, Huchuan Lu, You He, and Gang Wang.
\newblock A bi-directional message passing model for salient object detection.
\newblock In {\em CVPR}, pages 1741--1750, 2018.

\bibitem{zhang2018hyperfusion}
Pingping Zhang, Wei Liu, Yinjie Lei, and Huchuan Lu.
\newblock Hyperfusion-net: Hyper-densely reflective feature fusion for salient
  object detection.
\newblock {\em PR}, 93:521--533, 2019.

\bibitem{zhang2019salient}
Pingping Zhang, Wei Liu, Huchuan Lu, and Chunhua Shen.
\newblock Salient object detection with lossless feature reflection and
  weighted structural loss.
\newblock {\em IEEE TIP}, 28(6):3048--3060, 2019.

\bibitem{zhang2018non}
Pingping Zhang, Dong Wang, Huchuan Lu, and Hongyu Wang.
\newblock Non-rigid object tracking via deep multi-scale spatial-temporal
  discriminative saliency maps.
\newblock {\em arXiv:1802.07957}, 2018.

\bibitem{zhang2017amulet}
Pingping Zhang, Dong Wang, Huchuan Lu, Hongyu Wang, and Xiang Ruan.
\newblock Amulet: Aggregating multi-level convolutional features for salient
  object detection.
\newblock In {\em ICCV}, pages 202--211, 2017.

\bibitem{zhang2017learning}
Pingping Zhang, Dong Wang, Huchuan Lu, Hongyu Wang, and Baocai Yin.
\newblock Learning uncertain convolutional features for accurate saliency
  detection.
\newblock In {\em ICCV}, pages 212--221, 2017.

\bibitem{zhang2018agile}
Pingping Zhang, Luyao Wang, Dong Wang, Huchuan Lu, and Chunhua Shen.
\newblock Agile amulet: Real-time salient object detection with contextual
  attention.
\newblock {\em arXiv:1802.06960}, 2018.

\bibitem{zhao2015saliency}
Rui Zhao, Wanli Ouyang, Hongsheng Li, and Xiaogang Wang.
\newblock Saliency detection by multi-context deep learning.
\newblock In {\em CVPR}, pages 1265--1274, 2015.

\bibitem{zhu2015unsupervised}
Jun-Yan Zhu, Jiajun Wu, Yan Xu, Eric Chang, and Zhuowen Tu.
\newblock Unsupervised object class discovery via saliency-guided multiple
  class learning.
\newblock {\em IEEE TPAMI}, 37(4):862--875, 2015.

\end{thebibliography}
}

\end{document}